\documentclass[10pt,journal,compsoc]{IEEEtran}
\ifCLASSOPTIONcompsoc
\else
\fi

\usepackage{url}
\usepackage{amsmath}
\usepackage{booktabs} 
\usepackage{graphicx}
\usepackage{color}
\usepackage{float}
\usepackage{array}
\usepackage{multirow}
\usepackage{amssymb}
\usepackage{graphbox}

\newcommand{\etal}{\textit{et al.}}

\begin{document}

\title{Prior Guided Feature Enrichment Network for Few-Shot Segmentation}

\author{Zhuotao Tian,~\IEEEmembership{Student Member,~IEEE,}
	    Hengshuang Zhao,~\IEEEmembership{Member,~IEEE,}
	    Michelle Shu,~\IEEEmembership{Student Member,~IEEE,}
	    Zhicheng Yang,~\IEEEmembership{Member,~IEEE,}
	    Ruiyu Li,~\IEEEmembership{Member,~IEEE,}
	    Jiaya Jia,~\IEEEmembership{Fellow,~IEEE}
        
\IEEEcompsocitemizethanks{
\IEEEcompsocthanksitem Z. Tian (\protect\url{tianzhuotao@link.cuhk.edu.hk}), H. Zhao and J. Jia are with the Department of Computer Science and Engineering, The Chinese University of Hong Kong.
M. Shu is with Johns Hopkins University.
Z. Yang and R. Li are with SmartMore.
\IEEEcompsocthanksitem  Corresponding Author: H. Zhao.
}
}
\definecolor{mypink3}{cmyk}{0, 0.7808, 0.4429, 0.1412}
\newcommand{\zttian}[1]{{\color{mypink3}{\bf\sf [zttian: #1]}}}

\IEEEtitleabstractindextext{%
\begin{abstract}
State-of-the-art semantic segmentation methods require sufficient labeled data to achieve good results and hardly work on unseen classes without fine-tuning. Few-shot segmentation is thus proposed to tackle this problem by learning a model that quickly adapts to new classes with a few labeled support samples. Theses frameworks still face the challenge of generalization ability reduction on unseen classes due to inappropriate use of high-level semantic information of training classes and spatial inconsistency between query and support targets. To alleviate these issues, we propose the Prior Guided Feature Enrichment Network (PFENet). It consists of novel designs of (1) a training-free prior mask generation method that not only retains generalization power but also improves model performance and (2) Feature Enrichment Module (FEM) that overcomes spatial inconsistency by adaptively enriching query features with support features and prior masks. Extensive experiments on PASCAL-5$^i$ and COCO prove that the proposed prior generation method and FEM both improve the baseline method significantly. Our PFENet also outperforms state-of-the-art methods by a large margin without efficiency loss. It is surprising that our model even generalizes to cases without labeled support samples. 
Our code is available at \protect\url{https://github.com/Jia-Research-Lab/PFENet/}.
\end{abstract}

}

\maketitle

\IEEEdisplaynontitleabstractindextext

\IEEEpeerreviewmaketitle

\IEEEraisesectionheading{\section{Introduction}}
\label{sec:intro}
\IEEEPARstart{R}{apid} development of deep learning has brought significant improvement to semantic segmentation. The iconic frameworks \cite{pspnet,deeplab} have profited a wide range of applications of automatic driving, robot vision, medical image, etc. The performance of these frameworks, however, worsens quickly without sufficient fully-labeled data or when working on unseen classes. Even if additional data is provided, fine-tuning is still time- and resource-consuming.

\begin{figure}[t]
\centering
    \begin{minipage}   {1.0\linewidth}
        \centering
        \includegraphics
        [width=1.0\linewidth]{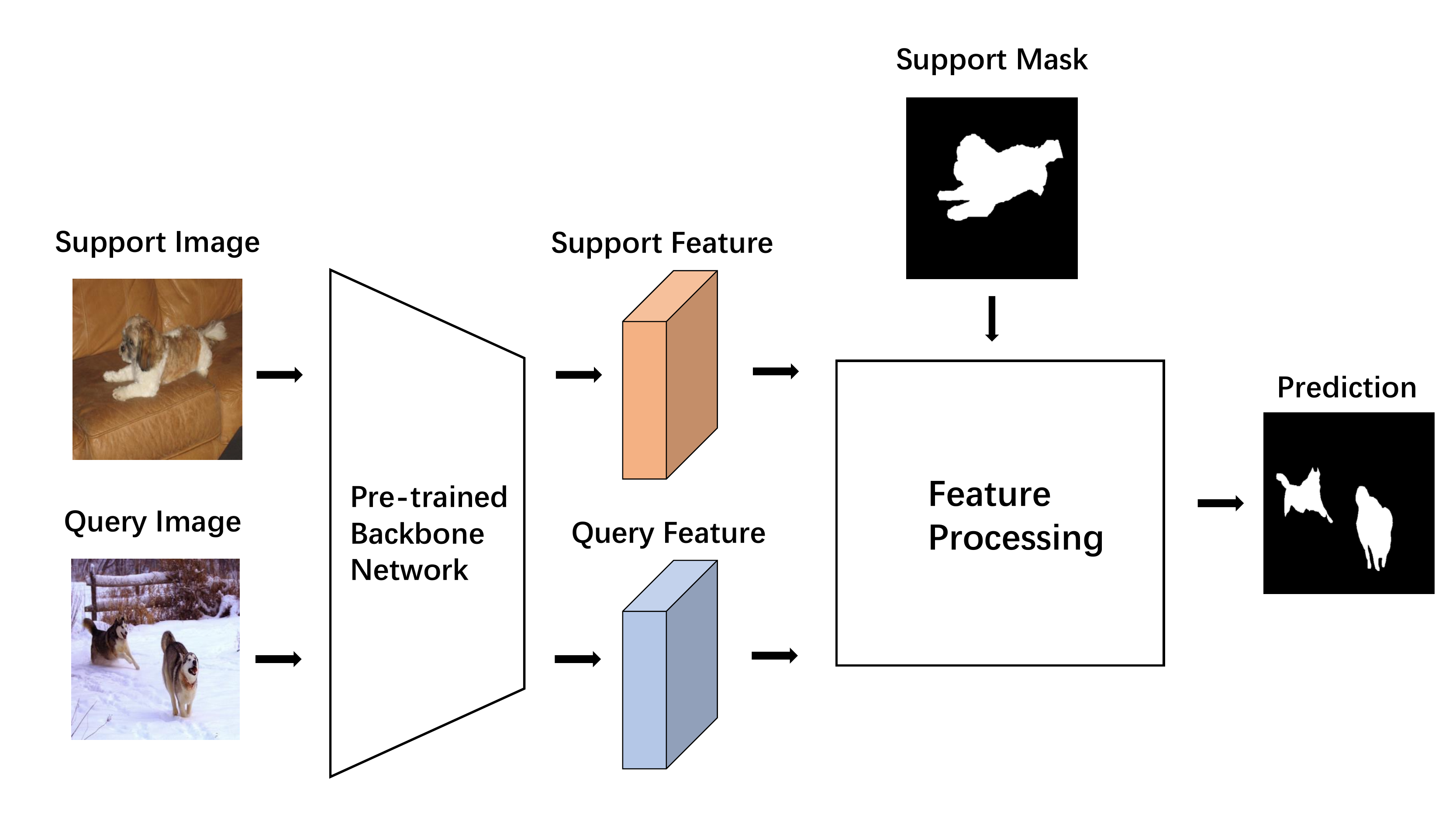}
    \end{minipage}
    \caption{Summary of recent few-shot segmentation frameworks. The backbone method used to extract support and query features can be either a single shared network or two Siamese networks. }
    \label{fig:first_exp}
\end{figure}

To address this issue, few-shot segmentation was proposed \cite{shaban} where data is divided into a support set and a query set. As shown in Figure \ref{fig:first_exp}, images from both support and query sets are first sent to the backbone network to extract features. Feature processing can be accomplished by generating weights for the classifier \cite{shaban,diffentiable}, cosine-similarity calculation \cite{prototype,panet,partaware}, or convolutions \cite{AttantionMCG, canet,mmm,SimPropNet,texturebias} to generate the final prediction. 

The support set provides information about the target class that helps the model to make accurate segmentation prediction on the query images. This process mimics the scenario where a model makes the prediction of unseen classes on testing images (query) with few labeled data (support). Therefore, a few-shot model needs to quickly adapt to the new classes. However, the common problems of existing few-shot segmentation methods include generalization loss due to misuse of high-level features and spatial inconsistency between the query and support samples. In this paper, we mainly tackle these two difficulties.

\vspace{0.1cm}
\noindent \textbf{Generalization Reduction \& High-level Features~~} 
Common semantic segmentation models rely heavily on high-level features with semantic information. Experiments of CANet \cite{canet} show that simply adding high-level features during feature processing in a few-shot model causes performance drop. Thus the way to utilize semantic information in the few-shot setting is not straightforward. Unlike previous methods, we use ImageNet \cite{imagenet} pre-trained high-level features of the query and support images to produce `priors' for the model. These priors help the model to better identify targets in query images. Since the prior generation process is training-free, the resulting model does not lose the generalization ability to unseen classes, despite the frequent use of high-level information of seen classes during training. 

\vspace{0.1cm}
\noindent \textbf{Spatial Inconsistency~~} 
Besides, due to the limited samples, scale and pose of each support object may vary greatly from its query target, which we call spatial inconsistency. To tackle this problem, we propose a new module named Feature Enrichment Module (FEM) to adaptively enrich query features with the support features. Ablation study in Section \ref{sec:ablation_study} shows that merely incorporating the multi-scale scheme to tackle the spatial inconsistency is sub-optimal by showing that FEM provides conditioned feature selection that helps retain essential information passed across different scales. FEM achieves superior performance than other multi-scale structures, such as HRNet \cite{hrnet_pami}, PPM \cite{pspnet}, ASPP \cite{aspp} and GAU \cite{Zhang_2019_ICCV}. 

Finally, based on the proposed prior generation method and Feature Enrichment Module (FEM), we build a new network -- Prior Guided Feature Enrichment Network (PFENet). The ResNet-50 based PFENet only contains 10.8 M learnable parameters, and yet achieves new state-of-the-art results on both PASCAL-5$^i$ \cite{shaban} and COCO \cite{coco} benchmark with 15.9 and 5.1 FPS with 1-shot and 5-shot settings respectively. 
Moreover, we manifest the effectiveness by applying our model to the zero-shot scenario where no labeled data is available. The result is surprising -- PFENet sill achieves decent performance without major structural modification.

Our contribution in this paper is threefold:
\begin{itemize}
    \item We leverage high-level features and propose training-free prior generation to greatly improve prediction accuracy and retain high generalization.
    \item By incorporating the support feature and prior information, our FEM helps adaptively refine the query feature with the conditioned inter-scale information interaction.
    \item PFENet achieves new state-of-the-art results on both PASCAL-5$^i$ and COCO datasets without compromising efficiency. 
\end{itemize}

\section{Related Work}

\subsection{Semantic Segmentation}
Semantic segmentation is a fundamental topic to predict the label for each pixel. The Fully Convolutional Network (FCN) \cite{fcn} is developed for semantic segmentation by replacing the fully-connected layer in a classification framework with convolutional layers. Following approaches, such as DeepLab~\cite{deeplab}, DPN~\cite{liu2015semantic} and CRF-RNN~\cite{zheng2015conditional}, utilize CRF/MRF to help refine coarse prediction. The receptive field is important for semantic segmentation; thus DeepLab~\cite{deeplab} and Dilation~\cite{yu2016multi} introduce the dilated convolution to enlarge the receptive field. Encoder-decoder structures~\cite{unet, ghiasi2016laplacian, lin2017refine} are adopted to help reconstruct and refine segmentation in steps. 

Contextual information is vital for complex scene understanding. ParseNet~\cite{liu2015parsenet} applies global pooling for semantic segmentation. PSPNet~\cite{pspnet} utilizes a Pyramid Pooling Module (PPM) for context information aggregation over different regions, which is very effective. DeepLab~\cite{deeplab} develops atrous spatial pyramid pooling (ASPP) with filters in different dilation rates. Attention models are also introduced. PSANet~\cite{psanet} develops point-wise spatial attention with a bi-directional information propagation paradigm. Channel-wise attention~\cite{encnet} and non-local style attention~\cite{cooccurant,danet,yuan2018ocnet,ccnet} are also effective for segmentation. These methods work well on large-sample classes. They are not designed to deal with rare and unseen classes. They also cannot be easily adapted without fine-tuning.

\subsection{Few-shot Learning}
Few-shot learning aims at image classification when only a few training examples are available. There are meta-learning based methods \cite{memory_match,dynamic_noforget,maml} and metric-learning ones \cite{matchingnet,relationnet,prorotypenet,deepemd}. Data is essential to deep models; therefore, several methods improve performance by synthesizing more training samples \cite{hallucinate_saliency, hallucinating, imaginary}. Different from few-shot learning where prediction is at the image-level, few-shot segmentation makes pixel-level predictions, which is much more challenging.

Our work closely relates to metric-learning based few-shot learning methods. Prototypical network \cite{prorotypenet} is trained to map input data to a metric space where classes are represented as prototypes. During inference, classification is achieved by finding the closest prototype for each input image, because data belonging to the same class should be close to the prototype. Another representative metric-based work is the relation network \cite{relationnet} that projects query and support images to 1$\times$1 vectors and then performs classification based on the cosine similarity between them.

\subsection{Few-shot Segmentation}
Few-shot segmentation places the general semantic segmentation in a few-shot scenario, where models perform dense pixel labeling on new classes with only a few support samples. OSLSM \cite{shaban} first tackles few-shot segmentation by learning to generate weights of the classifier for each class. PL \cite{prototype} applies prototyping \cite{prorotypenet} to the segmentation task. It learns a prototype for each class and calculates the cosine similarity between pixels and prototypes to make the prediction. More recently,  CRNet \cite{crnet} processes query and support images through a Siamese Network followed by a Cross-Reference Module to mine cooccurrent features in two images. PANet \cite{panet} introduces prototype alignment regularization that encourages the model to learn consistent embedding prototypes for better performance, and CANet \cite{canet} uses the iterative optimization module on the merged query and support feature to iteratively refine results. 

Similar to CANet \cite{canet}, we use convolution to replace the cosine similarity that may not well tackle complex pixel-wise classification in the segmentation task. However, different from CANet, our baseline model uses fewer convolution operations and still achieves decent performance.

As discussed before, these few-shot segmentation methods do not sufficiently consider generalization loss and spatial inconsistency. Unlike PGNet \cite{Zhang_2019_ICCV} that uses a graph-based pyramid structure to refine results via Graph Attention Unit (GAU) followed by three residual blocks and an ASPP \cite{aspp}, we instead incorporate a few basic convolution operations with the proposed prior masks and FEM in a multi-scale structure to accomplish decent performance.

\begin{figure*}[t]
\centering
    \begin{minipage}   {0.12\linewidth}
        \centering
        \includegraphics [width=1\linewidth,height=0.8\linewidth] 
        {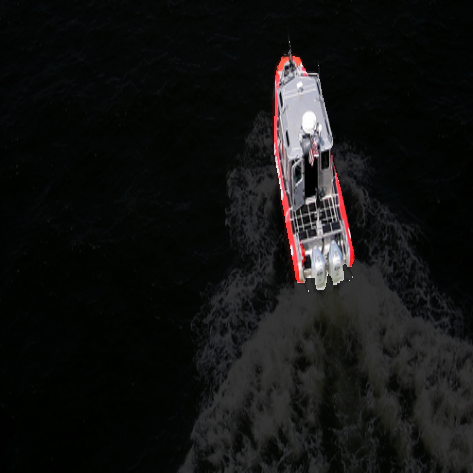}
    \end{minipage}
    \begin{minipage}   {0.12\linewidth}
        \centering
        \includegraphics [width=1\linewidth,height=0.8\linewidth]
        {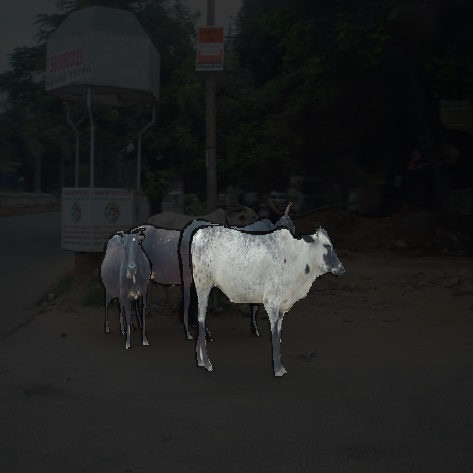}
    \end{minipage}
    \begin{minipage}   {0.12\linewidth}
        \centering
        \includegraphics [width=1\linewidth,height=0.8\linewidth]
        {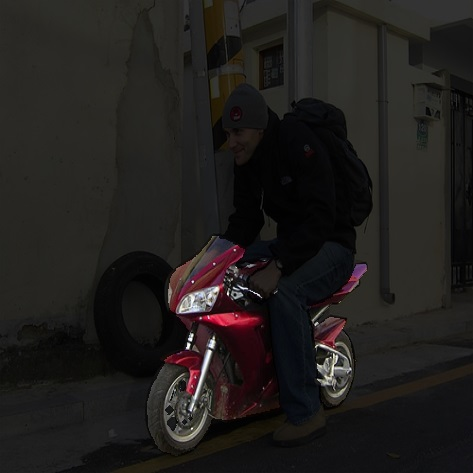}
    \end{minipage}
    \begin{minipage}   {0.12\linewidth}
        \centering
        \includegraphics [width=1\linewidth,height=0.8\linewidth]
        {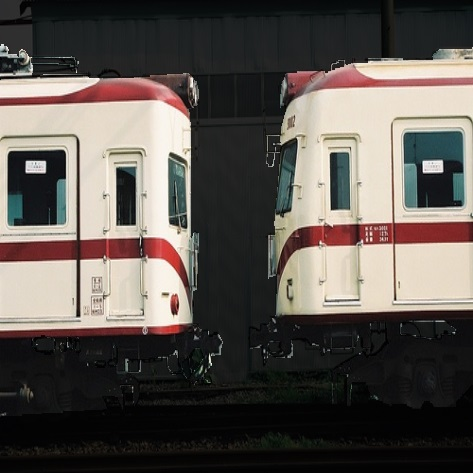}
    \end{minipage}
    \begin{minipage}   {0.12\linewidth}
        \centering
        \includegraphics [width=1\linewidth,height=0.8\linewidth]
        {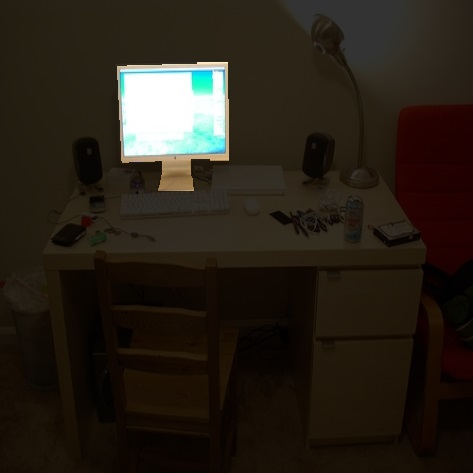}
    \end{minipage}    
    \begin{minipage}   {0.12\linewidth}
        \centering
        \includegraphics [width=1\linewidth,height=0.8\linewidth]
        {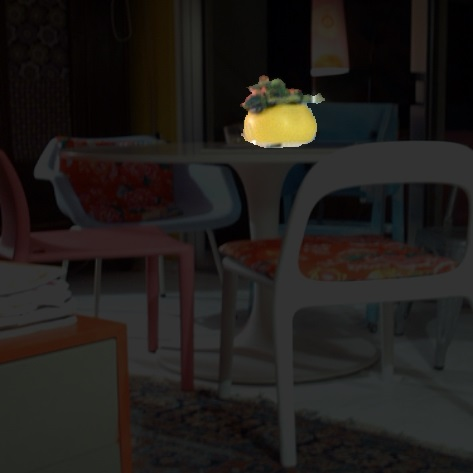}
    \end{minipage}      
    \begin{minipage}   {0.12\linewidth}
        \centering
        \includegraphics [width=1\linewidth,height=0.8\linewidth]
        {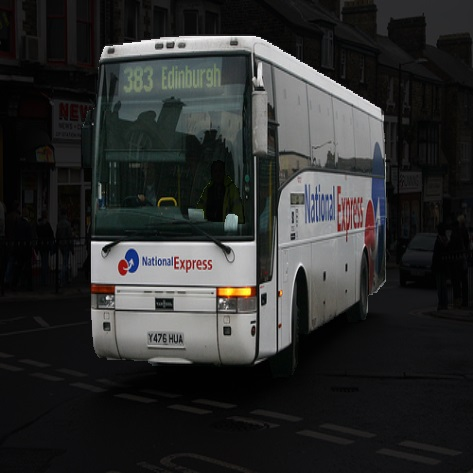}
    \end{minipage}      
    \begin{minipage}   {0.12\linewidth}
        \centering
        \includegraphics [width=1\linewidth,height=0.8\linewidth]
        {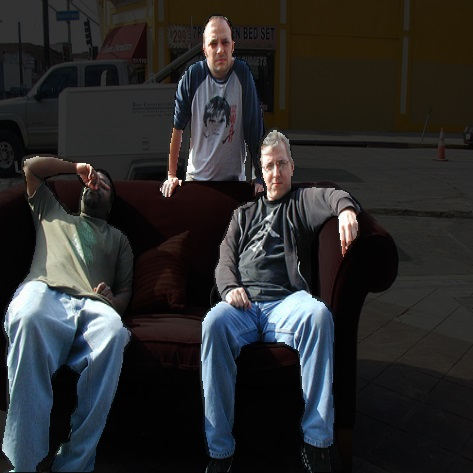}
    \end{minipage}       
    \vspace{0.075cm}
    
    \begin{minipage}   {0.12\linewidth}
        \centering
        \includegraphics [width=1\linewidth,height=0.8\linewidth] 
        {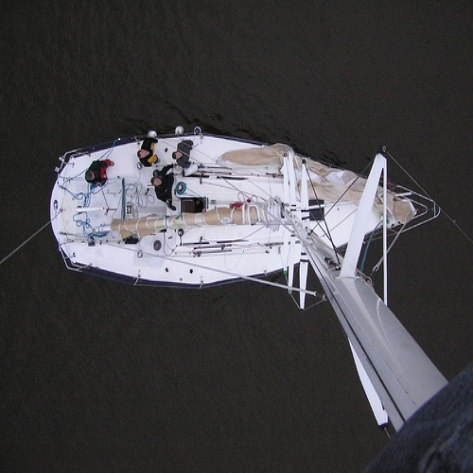}
    \end{minipage}
    \begin{minipage}   {0.12\linewidth}
        \centering
        \includegraphics [width=1\linewidth,height=0.8\linewidth]
        {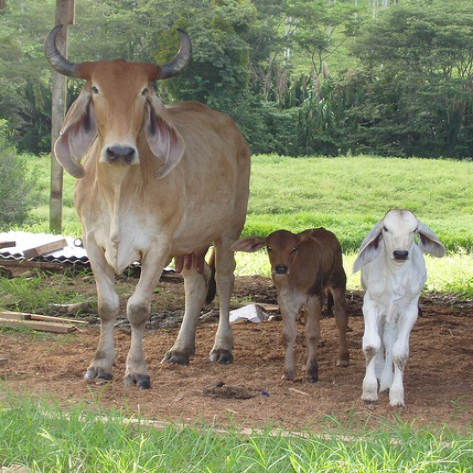}
    \end{minipage}
    \begin{minipage}   {0.12\linewidth}
        \centering
        \includegraphics [width=1\linewidth,height=0.8\linewidth]
        {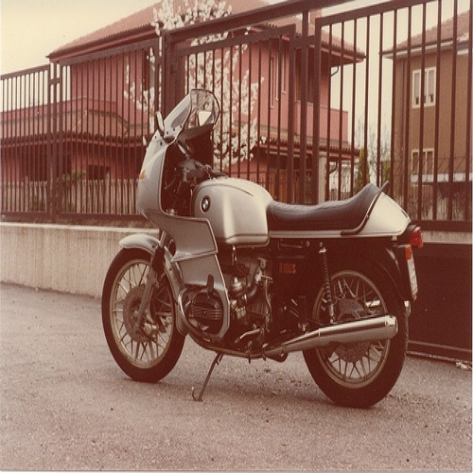}
    \end{minipage}
    \begin{minipage}   {0.12\linewidth}
        \centering
        \includegraphics [width=1\linewidth,height=0.8\linewidth]
        {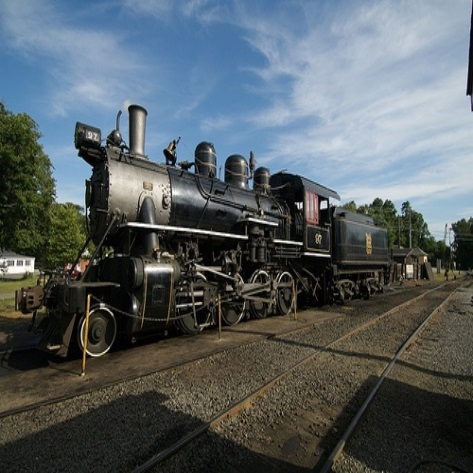}
    \end{minipage}
    \begin{minipage}   {0.12\linewidth}
        \centering
        \includegraphics [width=1\linewidth,height=0.8\linewidth]
        {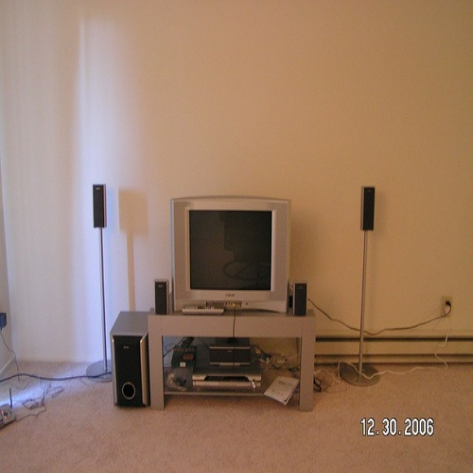}
    \end{minipage}
    \begin{minipage}   {0.12\linewidth}
        \centering
        \includegraphics [width=1\linewidth,height=0.8\linewidth]
        {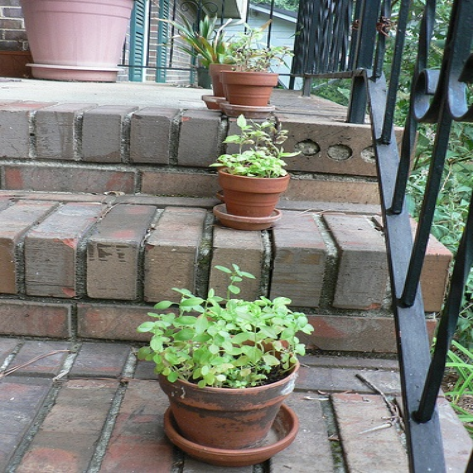}
    \end{minipage}      
    \begin{minipage}   {0.12\linewidth}
        \centering
        \includegraphics [width=1\linewidth,height=0.8\linewidth]
        {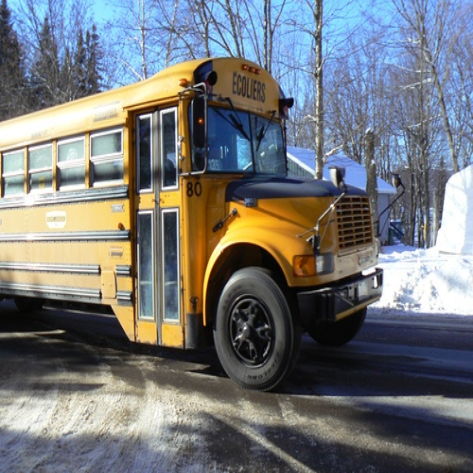}
    \end{minipage}    
    \begin{minipage}   {0.12\linewidth}
        \centering
        \includegraphics [width=1\linewidth,height=0.8\linewidth]
        {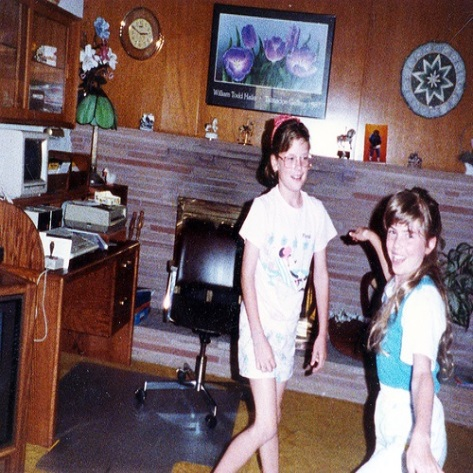}
    \end{minipage}       
    \vspace{0.075cm}
    
    \begin{minipage}   {0.12\linewidth}
        \centering
        \includegraphics [width=1\linewidth,height=0.8\linewidth] 
        {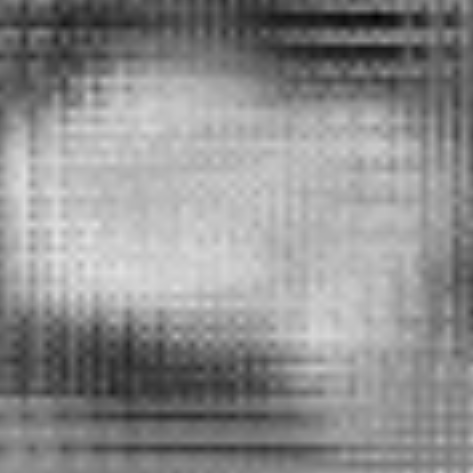}
        {Boat}
    \end{minipage}
    \begin{minipage}   {0.12\linewidth}
        \centering
        \includegraphics [width=1\linewidth,height=0.8\linewidth]
        {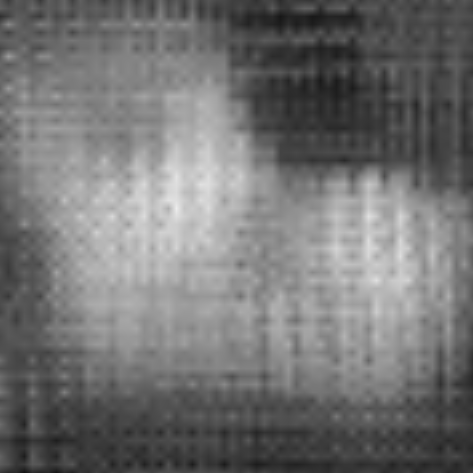}
        {Cow}
    \end{minipage}
    \begin{minipage}   {0.12\linewidth}
        \centering
        \includegraphics [width=1\linewidth,height=0.8\linewidth]
        {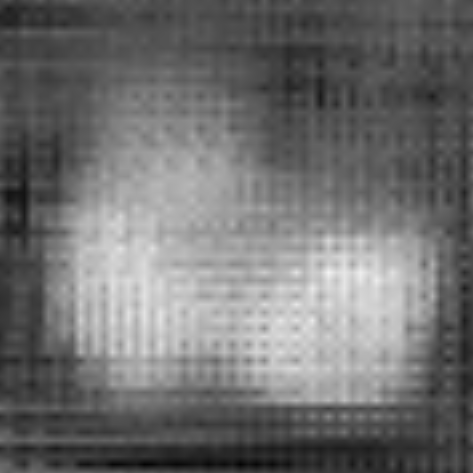}
        {Motor}
    \end{minipage}
    \begin{minipage}   {0.12\linewidth}
        \centering
        \includegraphics [width=1\linewidth,height=0.8\linewidth]
        {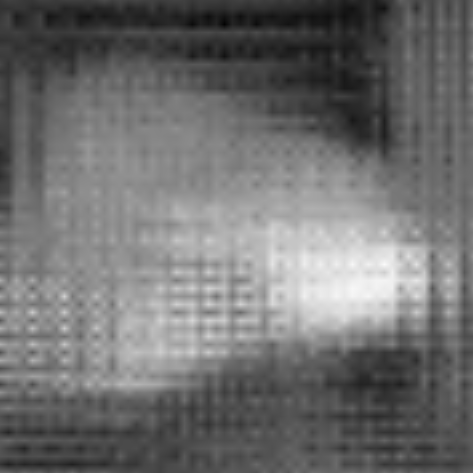}
        {Train}
    \end{minipage}
    \begin{minipage}   {0.12\linewidth}
        \centering
        \includegraphics [width=1\linewidth,height=0.8\linewidth]
        {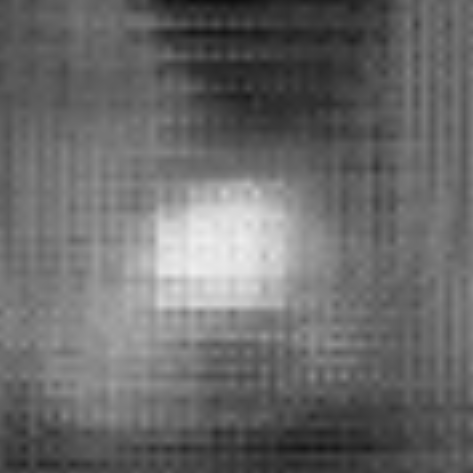}
        {Tv/Monitor}
    \end{minipage}   
    \begin{minipage}   {0.12\linewidth}
        \centering
        \includegraphics [width=1\linewidth,height=0.8\linewidth]
        {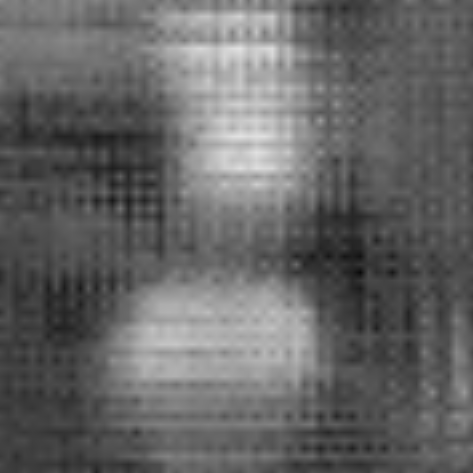}
        {Potted Plant}
    \end{minipage}      
    \begin{minipage}   {0.12\linewidth}
        \centering
        \includegraphics [width=1\linewidth,height=0.8\linewidth]
        {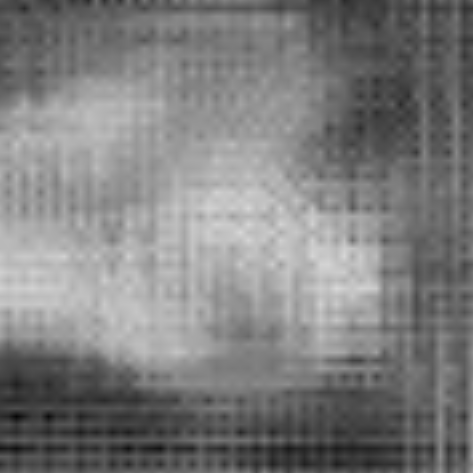}
        {Bus}
    \end{minipage}          
    \begin{minipage}   {0.12\linewidth}
        \centering
        \includegraphics [width=1\linewidth,height=0.8\linewidth]
        {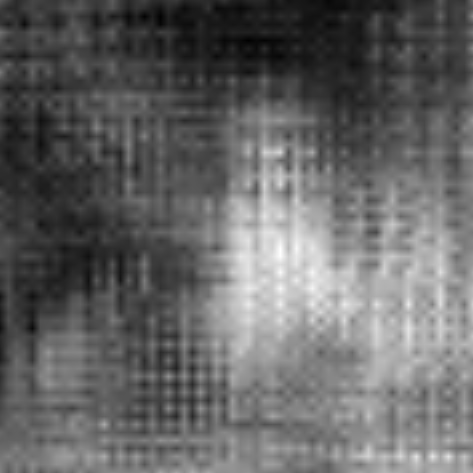}
        {Person}
    \end{minipage}    
    
    \caption{Illustration of the training-free prior generation. \textbf{Top}: support images with the masked area in the target class. \textbf{Middle}: query images. \textbf{Bottom}: prior masks of query images where the regions of interest are highlighted. }
    \label{fig:prior}
\end{figure*}

\section{Our Method}
In this section, we first briefly describe the few-shot segmentation task in Section \ref{sec:taskdescription}. Then, we present the prior generation method and the Feature Enrichment Module (FEM) in Sections \ref{sec:prior_intro} and \ref{sec:fem_intro} respectively. Finally, in Section \ref{sec:pfenet_intro}, details of our proposed Prior Guided Feature Enrichment Network (PFENet) are discussed.

\subsection{Task Description}
\label{sec:taskdescription}
A few-shot semantic segmentation system has two sets, i.e., the query set $Q$ and support set $S$. Given $K$ samples from support set $S$, the goal is to segment the area of unseen class $C_{test}$ from each query image $I_Q$ in the query set. 

Models are trained on classes $C_{train}$ (base) and tested on previously unseen classes $C_{test}$ (novel) in episodes $(C_{train} \cap C_{test} = \varnothing)$. The episode paradigm was proposed in  \cite{matchingnet} and was first applied to few-shot segmentation in \cite{shaban}. Each episode is formed by a support set $S$ and a query set $Q$ of the same class $c$. The support set $S$ consists of $K$ samples $S=\{S_1, S_2, ..., S_K\}$ of class $c$, which we call `K-shot scenario'. The $i$-th support sample $S_i$ is a pair of $\{I_{S_i}, M_{S_i}\}$ where $I_{S_i}$ and $M_{S_i}$ are the support image and label of $c$ respectively. For the query set, $Q=\{I_Q, M_Q\}$ where $I_Q$ is the input query image and $M_Q$ is the ground truth mask of class $c$. The query-support pair $\{I_Q, S\} = \{I_Q, I_{S_1}, M_{S_1}, I_{S_2}, M_{S_2}, ..., I_{S_K}, M_{S_K}\}$ forms the input data batch to the model. The ground truth $M_Q$ of the query image is invisible to the model and is used to evaluate the prediction on the query image in each episode.

\begin{figure*}[t]
	\begin{center}
		\includegraphics[width=1.0\linewidth]{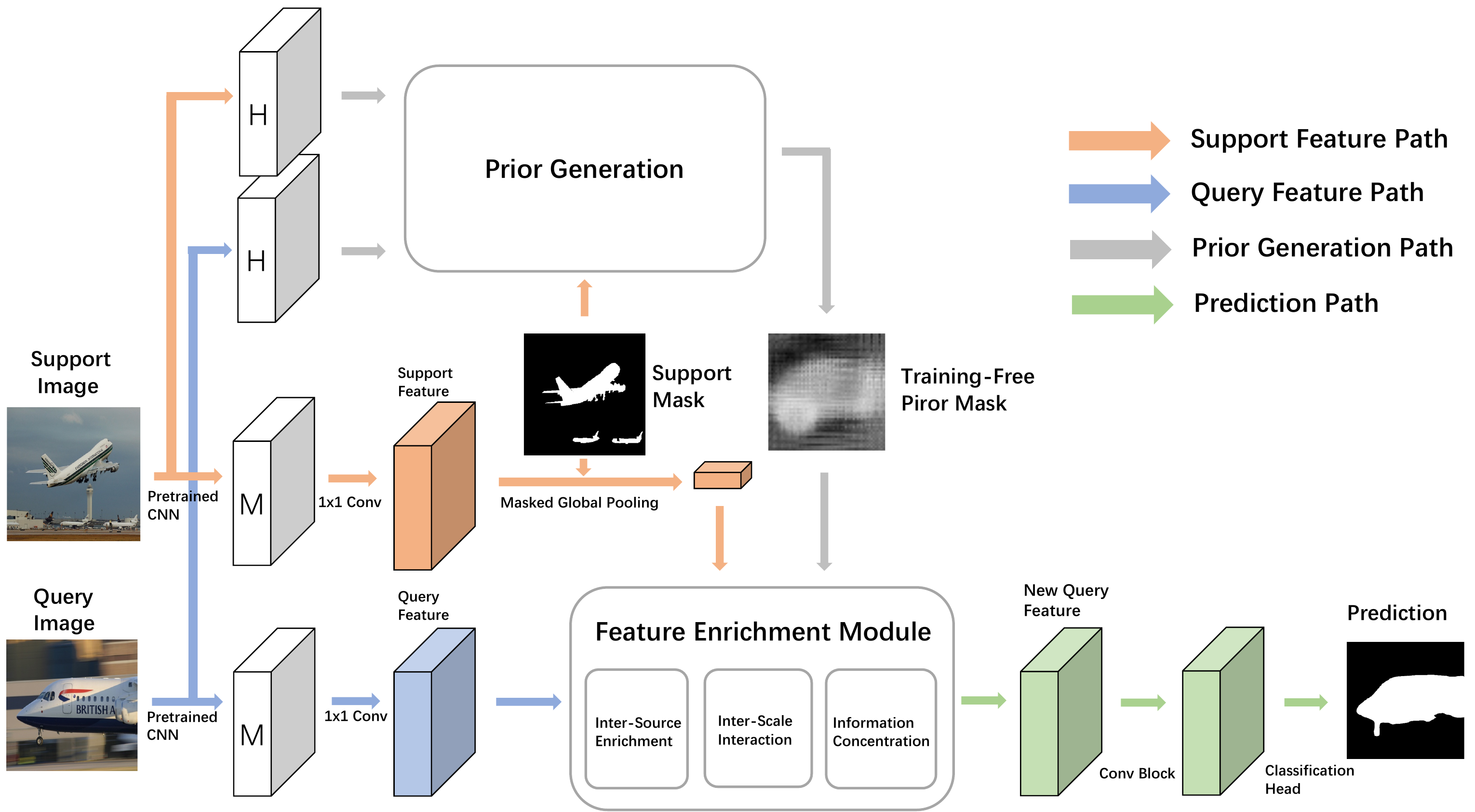}
	\end{center}
\caption{Overview of our Prior Guided Feature Enrichment Network with the prior generation and Feature Enrichment Module. White blocks marked with $H$ and $M$ represent the high- and middle-level features extracted from backbone respectively.}
\label{fig:network}
\end{figure*}

\subsection{Prior for Few-Shot Segmentation}
\label{sec:prior_intro}
\subsubsection{Important Observations}
CANet \cite{canet} outperforms previous work by a large margin on the benchmark PASCAL-5$^i$ dataset by extracting only middle-level features from the backbone (e.g., conv3\_x and conv4\_x of ResNet-50). Experiments in CANet also show that the high-level (e.g., conv5\_x of ResNet-50)  features lead to performance reduction. It is explained in \cite{canet} that the middle-level feature performs better since it constitutes object parts shared by unseen classes, but our alternative explanation is that the {\it semantic information contained in the high-level feature is more class-specific than the middle-level feature}, indicating that the former is more likely to negatively affect model's generalization power to unseen classes. In addition, higher-level feature directly provides semantic information of the training classes $C_{train}$, contributing more in identifying pixels belonging to $C_{train}$ and reducing the training loss than the middle-level information. Consequently, such behavior results in a preference for $C_{train}$. The lack of generalization and the preference for the training classes are both harmful for evaluation on unseen test classes $C_{test}$. 

It is noteworthy that contrary to the finding that high-level feature adversely affects performance in few-shot segmentation, prior segmentation frameworks \cite{icnet,unet} exploit these features to provide semantic cues for final prediction. This contradiction motivates us to find a way to make use of high-level information in a training-class-insensitive way to boost performance in few-shot segmentation.

\subsubsection{Prior Generation}
In our work, we transform the ImageNet \cite{imagenet} pre-trained high-level feature containing semantic information into a prior mask that tells the probability of pixels belonging to a target class as shown in Figure \ref{fig:prior}. During training, the backbone parameters are fixed as those in \cite{panet,canet}. Therefore, the prior generation process does not bias towards training classes $C_{train}$ and upholds class-insensitivity during the evaluation on unseen test classes $C_{test}$. Let $I_Q, I_S$ denote the input query and support images, $M_S$ denote the binary support mask, $\mathcal{F}$ denote the backbone network, and $X_Q, X_S$ denote the high-level query and support features. We have
\begin{equation}
\label{eqn:prior_1}
X_Q = \mathcal{F}(I_Q), \quad X_S = \mathcal{F}(I_S)\odot M_S,
\end{equation}
where $\odot$ is the Hadamard product -- the sizes of $X_Q$ and $X_S$ are both $[h, w, c]$. Note that the output of $\mathcal{F}$ is processed with a ReLU function. So the binary support mask $M_S$ removes the background in support feature by setting it to zero. 

Specifically, we define the prior $Y_Q$ of query feature $X_Q$ as the mask that reveals the pixel-wise correspondence between $X_Q$ and $X_S$. A pixel of query feature $X_Q$ with a high value on $Y_Q$ means that this pixel has a high correspondence with at least one pixel in support feature. Thus, it is very likely to be in the target area of the query image. By setting the background on support feature to zero, pixels of query feature yield no correspondence with the background on support feature -- they only correlate with the foreground target area. To generate $Y_Q$, we first calculate the pixel-wise cosine similarity $cos(x_q, x_s) \in \mathbb{R}$ between feature vectors of $x_q \in X_Q$ and $x_s \in X_S$ as
\begin{equation}
%\vspace{-0.05cm}
\label{eqn:cosine}
cos(x_q, x_s) = \frac{x_q^T x_s}{\left \| x_q \right \| \left \| x_s \right \|}\quad
q, s \in \{1, 2, ..., hw\}
%\vspace{-0.05cm}
\end{equation}
For each $x_q \in X_Q$, we take the maximum similarity among all support pixels as the correspondence value $c_q \in \mathbb{R}$ as
\begin{eqnarray}
\label{eqn:maxcosine}
c_q &=& \max\limits_{s\in \{1, 2, ..., hw\}}(cos(x_q, x_s)),\\
\label{eqn:prior}
C_Q &=& [c_1, c_2, ..., c_{hw}] \in \mathbb{R}^{hw\times 1}.
\end{eqnarray}
Then we produce the prior mask $Y_Q$ by reshaping $C_Q \in \mathbb{R}^{hw\times 1}$ into $Y_Q  \in \mathbb{R}^{h\times w\times 1}$. We process $Y_Q$ with a min-max normalization (Eq. \eqref{eqn:norm}) to normalize the values to between 0 and 1, as shown in Figure \ref{fig:prior}. In Eq. \eqref{eqn:norm}, $\epsilon$ is set to $1e-7$ in our experiments. 
\begin{equation}
\label{eqn:norm}
Y_Q = \frac{Y_Q - \min(Y_Q)}{\max(Y_Q) - \min(Y_Q) + \epsilon}.
\end{equation}

The key point of our proposed prior generation method lies in the use of fixed high-level features to yield the prior mask by taking the maximum value from a similarity matrix of size $hw \times hw$ as given in Eqs. \eqref{eqn:cosine} and \eqref{eqn:maxcosine}, which is rather simple and effective. Ablation study comparing other alternative methods used in \cite{panet,weighingboosting,SG-One} in Section \ref{sec:ablation_prior} demonstrates the superiority of our method. 

\subsection{Feature Enrichment Module}
\label{sec:fem_intro}
\subsubsection{Motivation}
Existing few-shot segmentation frameworks  \cite{canet,shaban,AttantionMCG,panet,weighingboosting,guide,adaptivemaskweightimprinting,prototype} use masked global average pooling for extracting class vectors from support images before further processing. However, global pooling on support images results in spatial information inconsistency since the area of query target may be much larger or smaller than support samples. Therefore, using a global pooled support feature to directly match each pixel of the query feature is not ideal. 

A natural alternative is to add PPM \cite{pspnet} or ASPP \cite{aspp} to provide multi-level spatial information to the feature. PPM and ASPP help the baseline model yield better performance (as demonstrated in our later experiments). However, these two modules are suboptimal in that: \textbf{1)} they provide spatial information to merged features without specific refinement process within each scale; \textbf{2)} the hierarchical relations across different scales are ignored. 

To alleviate these issues, we disentangle the multi-scale structure and propose the feature enrichment module (FEM) to \textbf{1)} horizontally interact the query feature with the support features and prior masks in each scale, and \textbf{2)} vertically leverage the hierarchical relations to enrich coarse feature maps with essential information extracted from the finer feature via a top-down information path. After horizontal and vertical optimization, features projected into different scales are then collected to form the new query feature. Details of FEM are as follows.

\begin{figure*}[t]
	\begin{center}
		\includegraphics[width=1.0\linewidth]{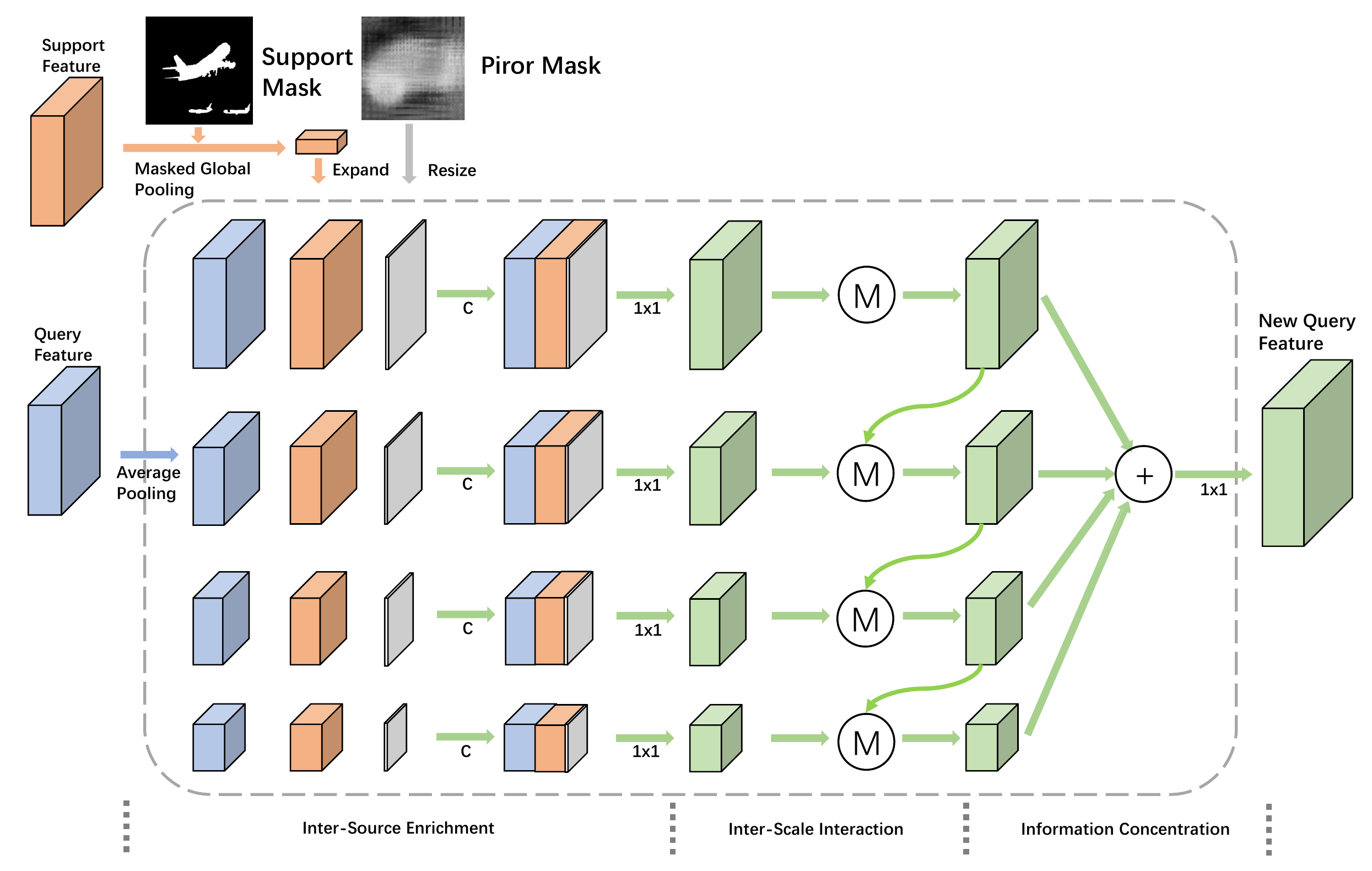}
	\end{center}
\vspace{-0.5cm}	
\caption{Visual illustration of FEM (dashed box) with four scales and a top-down path. C, 1x1 and Circled M represent concatenation, 1$\times$1 convolution and inter-scale merging module respectively. Activation functions are ReLU.}
\label{fig:FEM}
\end{figure*}

\subsubsection{Module Structure}
\label{sec:module_structure}
As shown in Figure \ref{fig:network}, the feature enrichment module (FEM) takes the query feature, prior mask and support feature as input. It outputs the refined query feature with enriched information from the support feature. The enrichment process can be divided into three sub-processes of \textbf{1)} inter-source enrichment that first projects input to different scales and then interacts the query feature with support feature and prior mask in each scale independently; \textbf{2)} inter-scale interaction that selectively passes essential information between merged query-support features across different scales; and \textbf{3)} information concentration that merges features in different scales to finally yield the refined query feature. An illustration of FEM with four scales and a top-down path for inter-scale interaction is shown in Figure \ref{fig:FEM}.

\begin{figure}[t]
	\begin{center}
		\includegraphics[width=1.0\linewidth]{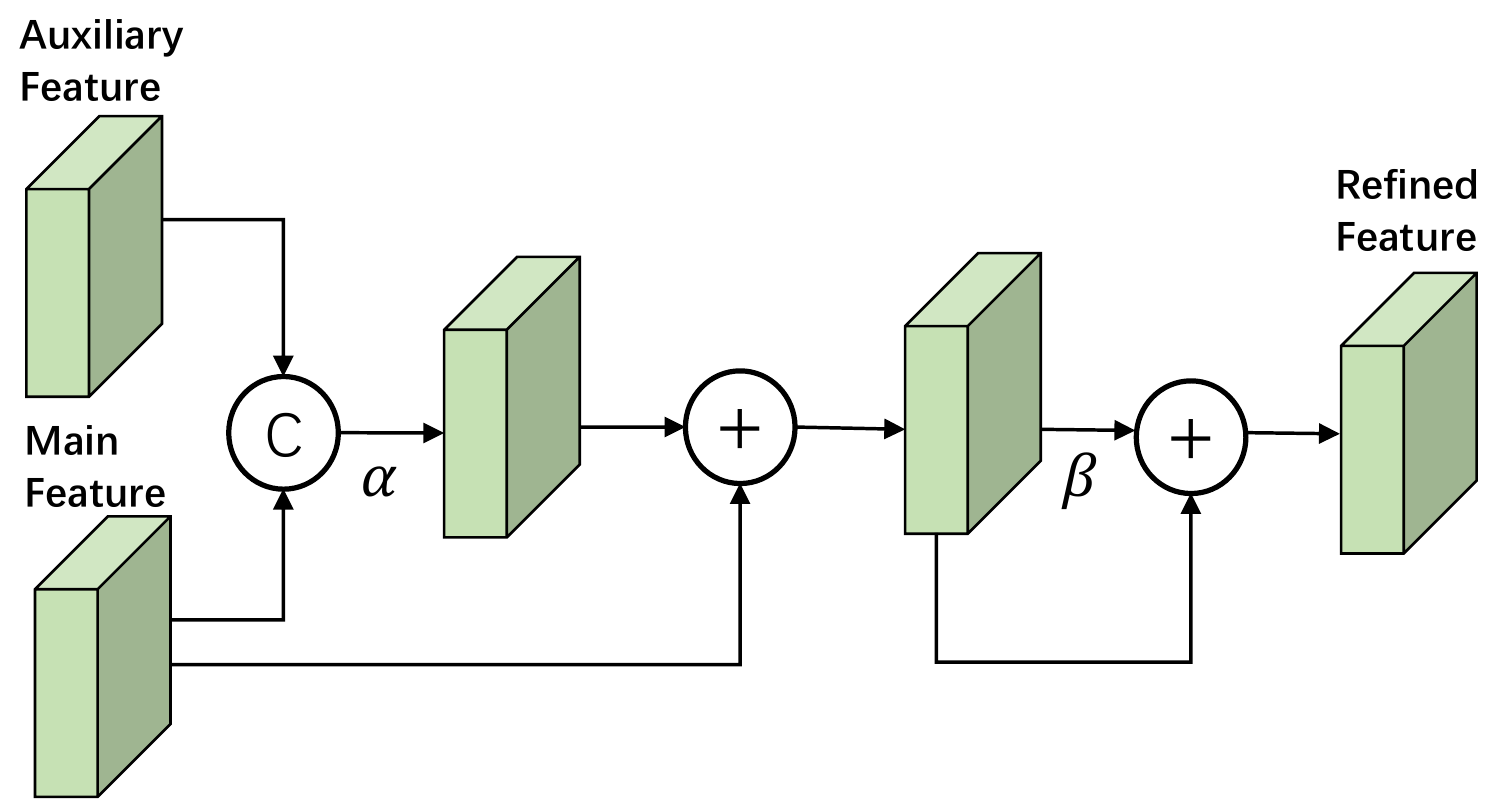}
	\end{center}
\vspace{-0.5cm}
\caption{Visual illustration of the inter-scale merging module $\mathcal{M}$. C is concatenation and $+$ is pixel-wise addition. $\alpha$ means 1$\times$1 convolution and $\beta$ represents two 3$\times$3 convolutions. Activation functions are ReLU. 
For features that do not have auxiliary features, there is no concatenation with the auxiliary feature and the refined feature is produced only by the main feature with $\alpha$ and $\beta$.}
\label{fig:merging_module}
\vspace{-0.3cm}
\end{figure}

\vspace{0.05in}
\noindent\textbf{Inter-Source Enrichment~~}
In FEM, $B = [B^1, B^2, ..., B^n] $ denotes $n$ different spatial sizes for average pooling. They are in the descending order $B^1>B^2>...>B^n$. The input query feature $X_Q \in \mathbb{R}^{h \times w \times c}$ is first processed with adaptive average pooling to generate $n$ sub-query features $X_Q^{FEM} = [X_Q^1, X_Q^2, ..., X_Q^n]$ of $n$ different spatial sizes $X_Q^{i} \in \mathbb{R}^{B^i \times B^i \times c}$. $n$ spatial sizes make the global-average pooled support feature $X_S \in \mathbb{R}^{1 \times 1 \times c}$ be expanded to different $n$ feature maps $X_S^{FEM} = [X_S^1, X_S^2, ..., X_S^n]$ ($X_S^{i} \in \mathbb{R}^{B^i \times B^i \times c}$), and the prior $Y_Q \in \mathbb{R}^{h \times w \times 1}$ is accordingly resized to $Y_Q^{FEM} = [Y_Q^1, Y_Q^2, ..., Y_Q^n]$ ($Y_Q^{i} \in \mathbb{R}^{B^i \times B^i \times 1}$).

Then, for $i \in \{1, 2, ..., n\}$, we concatenate $X_Q^i$, $X_S^i$ and $Y_Q^i$, and process each concatenated feature with convolutions to generate the merged query features $X_{Q,m}^{i} \in \mathbb{R}^{B^i \times B^i \times c}$ as
\begin{equation}
\label{eqn:fem_1}
X_{Q,m}^{i} = \mathcal{F}_{1\times1}(X_Q^i \oplus X_S^i \oplus Y_Q^i),
\end{equation}
where $\mathcal{F}_{1\times1}$ represents the 1$\times$1 convolution that yields the merged feature with $c=256$ output channels.

\vspace{0.05in}
\noindent\textbf{Inter-Scale Interaction~~}
It is worth noting that tiny objects may not exist in the down-sampled feature maps. A top-down path adaptively passing information from finer features to the coarse ones is conducive to building a hierarchical relationship within our feature enrichment module. Now the interaction is between not only the query and support features in each scale (horizontal), but also the merged features of different scales (vertical), which is beneficial to the overall performance. 

The circled $M$ in Figure \ref{fig:FEM} represents the inter-scale merging module $\mathcal{M}$ that interacts between different scales by selectively passing useful information from the auxiliary feature to the main feature to generate the refined feature $X_{Q,new}^{i}$. This process can be written as
\begin{equation}
\label{eqn:inter_scale_func}
X_{Q,new}^{i} = \mathcal{M}(X_{Q,m}^{Main, i}, X_{Q,m}^{Aux, i}),
\end{equation} 
where $X_{Q,m}^{Main, i}$ is the main feature and $X_{Q,m}^{Aux, i}$ is the auxiliary feature for the $i$-th scale $B^i$. For example, in an FEM with a top-down path for inter-scale interaction, finer feature (auxiliary) $X_{Q,m}^{i-1}$ needs to provide additional information to the coarse feature (main) $X_{Q,m}^{i}  (B^{i-1}>B^{i}, i \geqslant 2)$. In this case, $X_{Q,m}^{Aux, i} = X_{Q,m}^{i-1}$ and  $X_{Q,m}^{Main, i} = X_{Q,m}^{i}$. Other alternatives for inter-scale interaction include the bottom-up path that enriches finer features (main) with information coming from the coarse ones (auxiliary), and the bi-directional variants, i.e., a top-down path followed by a bottom-up path, and a bottom-up path followed by a top-down path. The top-down path shows its superiority in Section \ref{sec:inter_scale_ablation}.

The specific structure of the inter-scale merging module $\mathcal{M}$ is shown in Figure \ref{fig:merging_module}. We first resize the auxiliary feature to the same spatial size as the main feature. Then we use a 1$\times$1 convolution $\alpha$ to extract useful information from the auxiliary feature conditioned on the main feature. Two 3$\times$3 convolutions $\beta$ followed are used to finish the interaction and output the refined feature. The residual link within the inter-scale merging module $\mathcal{M}$ is used for keeping the integrity of the main feature in the output feature $X_{Q,new}^{i}$. For those features that do not have auxiliary features (e.g., the first merged feature $X_{Q,m}^{1}$ in the top-down path and the last merged feature $X_{Q,m}^{n}$ in the bottom-up path), we simply ignore the concatenation with the auxiliary feature in $\mathcal{M}$ -- the refined feature is produced only by the main feature.

\begin{figure}[t]
	\begin{center}
		\includegraphics[width=1.0\linewidth]{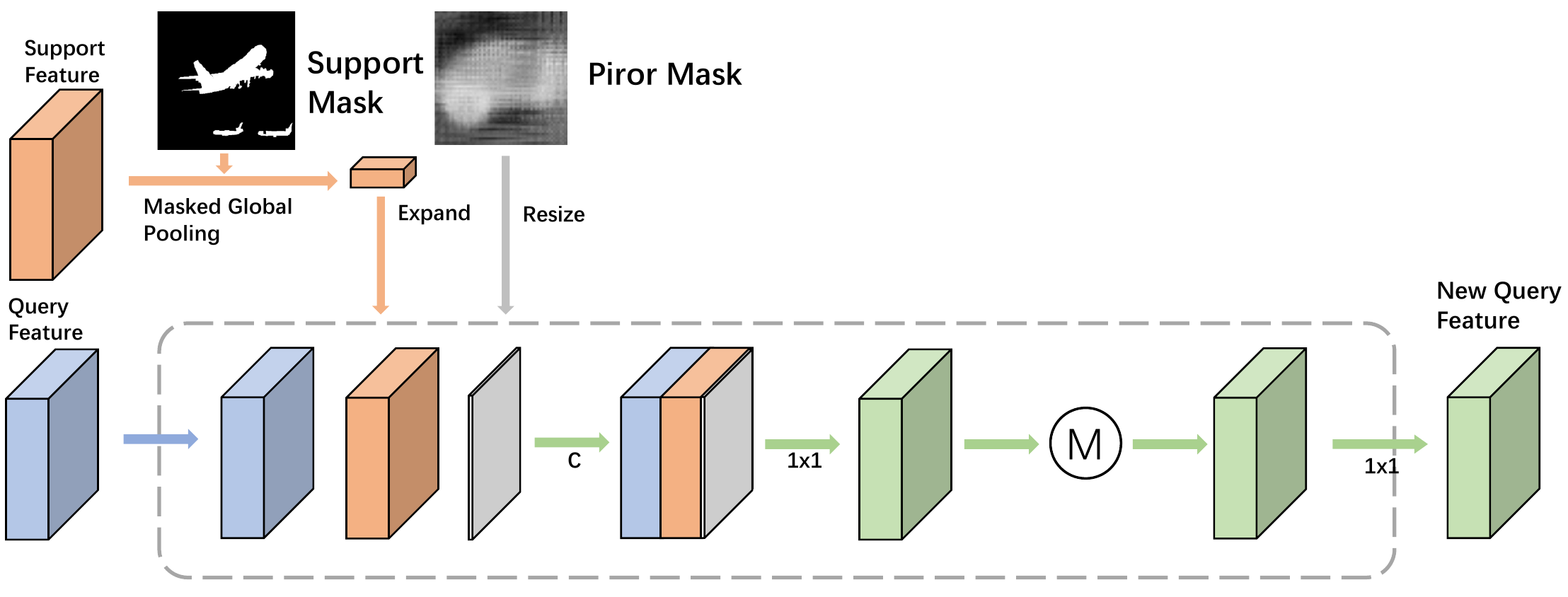}
	\end{center}
\vspace{-0.5cm}	
\caption{Visual illustration of the baseline structure that processes features in the original spatial size of the input features.}
\vspace{-0.5cm}	
\label{fig:visual_compare_fem}
\end{figure}

\vspace{0.05in}
\noindent\textbf{Information Concentration~~}
After inter-scale interaction, $n$ refined feature maps are obtained as $X_{Q,new}^{i}, i\in\{1,2,...,n\}$.
Finally, the output query feature $X_{Q,new} \in \mathbb{R}^{h \times w \times c}$ is formed by interpolation and concatenation of $n$ refined feature maps $X_{Q,new}^{i} \in \mathbb{R}^{h \times w \times c}$ followed by an 1$\times$1 convolution $\mathcal{F}_{1\times1}$ as
\begin{equation}
\label{eqn:fem_1}
X_{Q,new} = \mathcal{F}_{1\times1}(X_{Q,new}^{1} \oplus X_{Q,new}^{2} ... \oplus X_{Q,new}^{n}).
\end{equation}

The visual illustration of the baseline model without FEM ($B^1=h=w$) is shown in Figure \ref{fig:visual_compare_fem}. 
To encourage better feature enrichment, we add intermediate supervision by attaching classification head (Figure \ref{fig:refine_block}(b)) to each $X_{Q,new}^{i}$.

In summary, by incorporating the pooled support features and prior masks to query features with different spatial sizes, the model learns to adaptively enrich the query feature with information coming from the support feature at each location under the guidance of prior mask and supervision of ground-truth. Moreover, the vertical inter-scale interaction supplements the main feature with the conditioned information provided by the auxiliary feature. Therefore, FEM yields greater performance gain on baseline than other feature enhancement designs (e.g., PPM \cite{pspnet}, ASPP \cite{aspp} and GAU \cite{Zhang_2019_ICCV}). Experiments in Section \ref{sec:ablation_study} provide more details.

\subsection{Prior Guided Feature Enrichment Network}
\label{sec:pfenet_intro}
\subsubsection{Model Description}
Based on the proposed prior generation method and the feature enrichment module (FEM), we propose the Prior Guided Feature Enrichment Network (PFENet) as shown in Figure \ref{fig:network}. The ImageNet \cite{imagenet} pre-trained CNN is shared by support and query images to extract features. The extracted middle-level support and query features are processed by 1$\times$1 convolution to reduce the channel number to 256. 

\begin{figure}[t]
\centering
    \begin{minipage}   {0.49\linewidth}
        \centering
        \includegraphics [width=1\linewidth] 
        {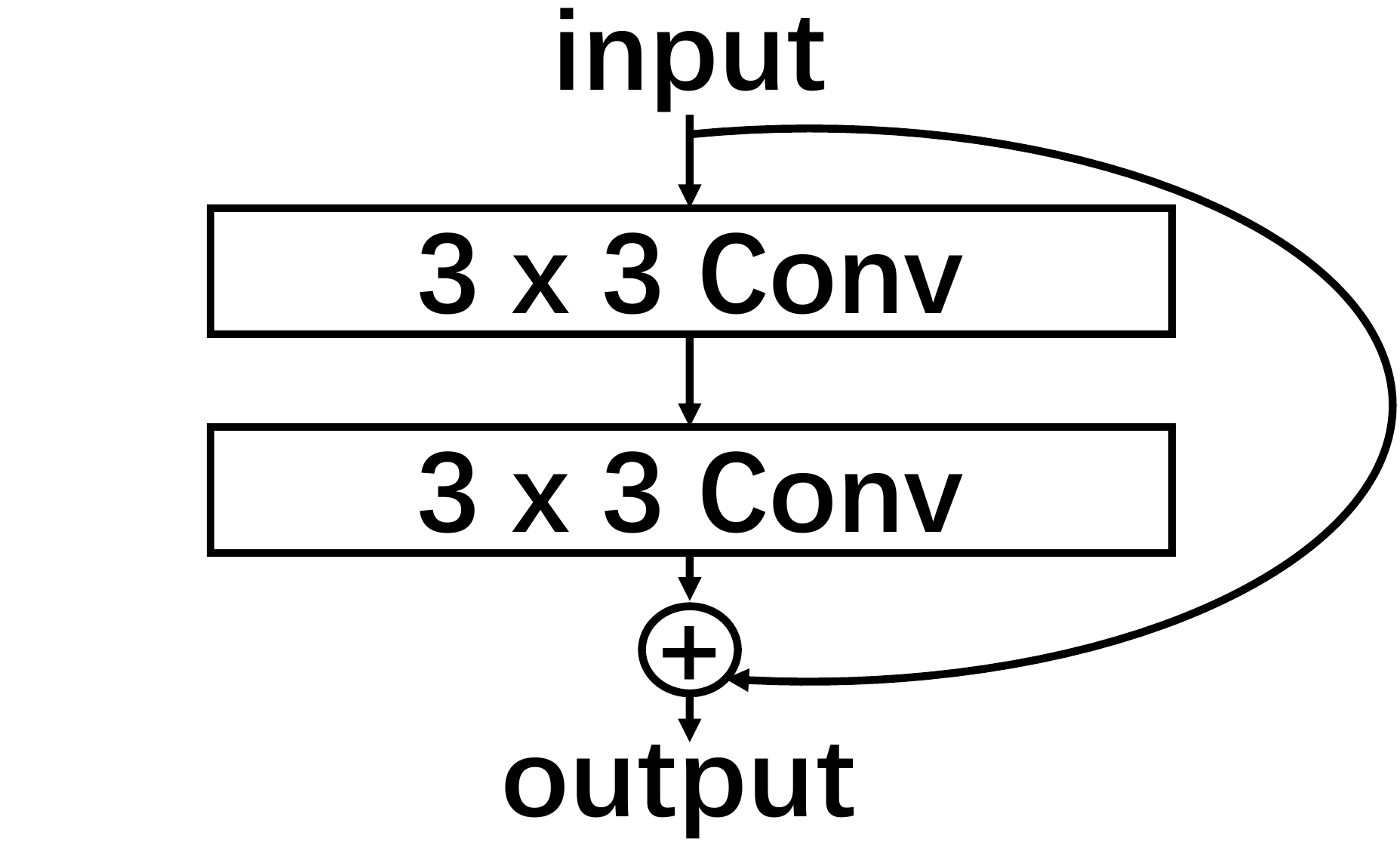}
        {(a)}
    \end{minipage}
    \begin{minipage}   {0.325\linewidth}
        \centering
        \includegraphics [width=1\linewidth] 
        {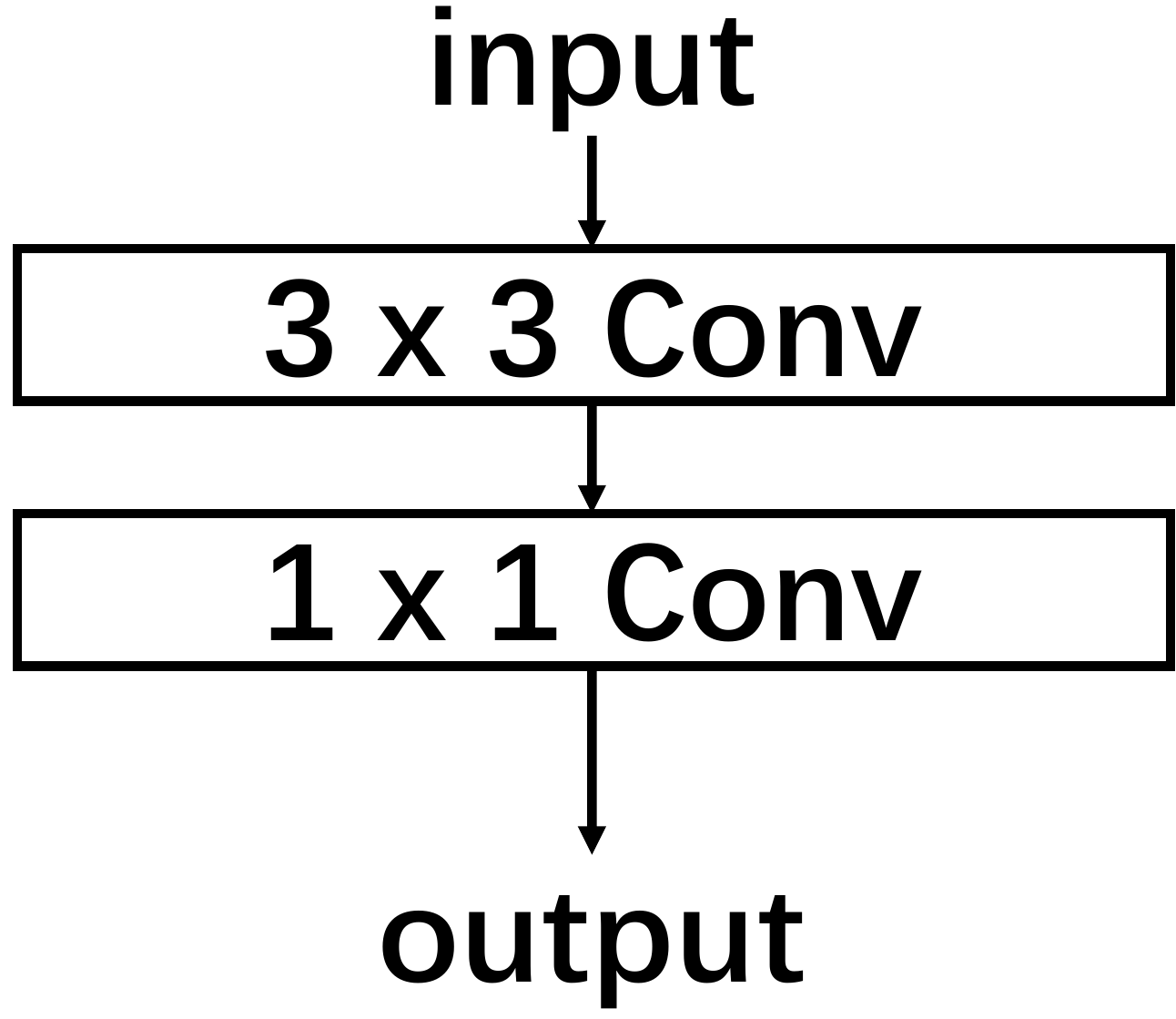}
        {(b)}        
    \end{minipage}
    \caption{Structures of (a) convolution block and (b) classification head.}
    \label{fig:refine_block}
\end{figure}

After feature extraction and channel reduction, the feature enrichment module (FEM) enriches the query feature with the support feature and prior mask. On the output feature of FEM, we apply a convolution block (Figure \ref{fig:refine_block}(a)) followed by a classification head to yield the final prediction. Classification head is composed of one 3$\times$3 convolution and 1$\times$1 convolution with Softmax function as shown in Figure \ref{fig:refine_block}(b). For all backbone networks, we use the outputs of the last layers of conv3\_x and conv4\_x as middle-level features $M$ to generate the query and support features by concatenation, and take the output of the last layer of conv5\_x as high-level features $H$ to produce the prior mask.

In the 5-shot setting, we simply take the average of 5 pooled support features as the new support feature before concatenation with the query feature. Similarly, the final prior mask before the concatenation in FEM is also obtained by averaging five prior masks produced by one query feature with different support features.

\subsubsection{Loss Function}
We select the cross entropy loss as our loss function. As shown in Section \ref{sec:module_structure} and Figure \ref{fig:network}, for a FEM with $n$ different spatial sizes, the intermediate supervision on $X_{Q,new}^{i}$ ($i \in \{1, 2, ..., n\}$) generates $n$ losses $\mathcal{L}_1^i$ ($i \in \{1, 2, ..., n\}$). The final prediction of PFENet generates the second loss $\mathcal{L}_2$. The total loss $\mathcal{L}$ is the weighted sum of $\mathcal{L}_1^i$ and $\mathcal{L}_2$ as
\begin{equation}
\label{eqn:loss}
\mathcal{L} = \frac{\sigma}{n} \sum_{i=1}^n \mathcal{L}_1^i + \mathcal{L}_2,
\end{equation}
where $\sigma$ is used to balance the effect of intermediate supervision. We empirically set $\sigma$ to 1.0 in all experiments.

\begin{table*}
    \caption{Class mIoU results on four folds of PASCAL-5$^i$.  \textbf{Params}:  number of learnable parameters.}
    \centering
    {
        \begin{tabular}{ c |  c  c  c  c | c |  c  c  c  c | c | c }
            \toprule
            \multirow{2}{*}{\textit{Methods}}  & \multicolumn{5}{c|}{1-Shot} & \multicolumn{5}{c|}{5-Shot}  & \multirow{2}{*}{Params} \\ 
            \specialrule{0em}{0pt}{1pt}
            \cline{2-11}
            \specialrule{0em}{1pt}{0pt} 
            & Fold-0 & Fold-1 & Fold-2 & Fold-3 & Mean & Fold-0 & Fold-1 & Fold-2 & Fold-3 & Mean &   \\
            
            \specialrule{0em}{0pt}{1pt}
            \hline
            \specialrule{0em}{1pt}{0pt}
            \multicolumn{12}{c}{VGG-16 Backbone} \\
            \specialrule{0em}{0pt}{1pt}
            \hline
            \specialrule{0em}{1pt}{0pt}
            
            OSLSM$_{2017}$ \cite{shaban} & 33.6 & 55.3 & 40.9 & 33.5 & 40.8  & 35.9 & 58.1 & 42.7 & 39.1 & 44.0  
            & 276.7M \\  
            co-FCN$_{2018}$ \cite{co-FCN} & 36.7 & 50.6 & 44.9 & 32.4 & 41.1  & 37.5 & 50.0 & 44.1 & 33.9 & 41.4  
            & 34.2M \\   
            SG-One$_{2018}$ \cite{SG-One} & 40.2 & 58.4 & 48.4 & 38.4 & 46.3 & 41.9 & 58.6 & 48.6 & 39.4 & 47.1 
            & 19.0M \\     
            AMP$_{2019}$ \cite{adaptivemaskweightimprinting}
            & 41.9 & 50.2 & 46.7 & 34.7 & 43.4
            & 41.8 & 55.5 & 50.3 & 39.9 & 46.9
            & 34.7M \\
            PANet$_{2019}$ \cite{panet} & 42.3 & 58.0 & 51.1 & 41.2 & 48.1 & 51.8 & 64.6 & \textbf{59.8} & 46.5 & 55.7 
            & 14.7M \\  
            FWBF$_{2019}$ \cite{weighingboosting} & 47.0 & 59.6 & 52.6 & 48.3 & 51.9  & 50.9 & 62.9 & 56.5 & 50.1 & 55.1
            & -  \\                           
            Ours
            & \textbf{56.9}  & \textbf{68.2}  & \textbf{54.4}  & \textbf{52.4}  & \textbf{58.0}  
            & \textbf{59.0}  & \textbf{69.1}  & 54.8  & \textbf{52.9 } & \textbf{59.0}  
            & \textbf{10.4M} \\

            \specialrule{0em}{0pt}{1pt}
            \hline
            \specialrule{0em}{1pt}{0pt}
            \multicolumn{12}{c}{ResNet-50 Backbone} \\
            \specialrule{0em}{0pt}{1pt}
            \hline
            \specialrule{0em}{1pt}{0pt} 
            CANet$_{2019}$ \cite{canet}  & 52.5 & 65.9 & 51.3 & 51.9 & 55.4 & 55.5 & 67.8 & 51.9 & 53.2 & 57.1  
            & 19.0M \\      
            PGNet$_{2019}$ \cite{Zhang_2019_ICCV}  & 56.0 & 66.9 & 50.6 & 50.4 & 56.0 & 54.9 & 67.4 & 51.8 & 53.0 & 56.8  
            & 17.2M \\                  
            Ours
            & \textbf{61.7}  & \textbf{69.5}  & \textbf{55.4}  & \textbf{56.3}  & \textbf{60.8}  
            & \textbf{63.1}  & \textbf{70.7}  & \textbf{55.8}  & \textbf{57.9}  & \textbf{61.9}              
            & \textbf{10.8M} \\             
            
            \specialrule{0em}{0pt}{1pt}
            \hline
            \specialrule{0em}{1pt}{0pt}
            \multicolumn{12}{c}{ResNet-101 Backbone}  \\
            \specialrule{0em}{0pt}{1pt}
            \hline
            \specialrule{0em}{1pt}{0pt} 
            FWBF$_{2019}$ \cite{weighingboosting} & 51.3 & 64.5 & \textbf{56.7} & 52.2 & 56.2 & 54.8 & 67.4 & \textbf{62.2} & 55.3 & 59.9   
            & -\\            
            Ours
            &  \textbf{60.5} & \textbf{69.4}  & 54.4  & \textbf{55.9} & \textbf{60.1}  
            & \textbf{62.8}  & \textbf{70.4}  & 54.9  & \textbf{57.6}  & \textbf{61.4}  
            & \textbf{10.8M}\\ 
            \bottomrule            
        \end{tabular}
    }
    \label{tab:compare_sota_class}
\end{table*}

\begin{table}
    \caption{FB-IoU results on PASCAL-5$^i$. Our results are single-scale ones without additional post-processing like DenseCRF \cite{densecrf}. As many other methods do not report the specific result of each fold, we present the comparison of the average FB-IoU results in this table.}
    \centering
    {
        \begin{tabular}{ c  | c  | c | c}
            \toprule
            \textit{Methods} 
            & 1-Shot 
            & 5-Shot  & Params \\ 
            \specialrule{0em}{0pt}{1pt}
            \hline
            \specialrule{0em}{1pt}{0pt}
            \multicolumn{4}{c}{VGG-16 Backbone} \\
            \specialrule{0em}{0pt}{1pt}
            \hline
            \specialrule{0em}{1pt}{0pt}
            OSLSM$_{2017}$ \cite{shaban}
            &  61.3
            &  61.5 
            & 272.6M \\      
            co-FCN$_{2018}$ \cite{co-FCN}
            &  60.1
            &  60.2  
            & 34.2M \\      
            PL$_{2018}$ \cite{prototype}
            &  61.2
            &  62.3  
            & - \\                 
            SG-One$_{2018}$ \cite{SG-One}
            & 63.9
            & 65.9  
            & 19.0M \\          
            PANet$_{2019}$ \cite{panet}
            &  66.5
            &  70.7  
            & 14.7M \\                           
            Ours
            &  \textbf{72.0}
            &  \textbf{72.3}  
            & \textbf{10.4M} \\

            \specialrule{0em}{0pt}{1pt}
            \hline
            \specialrule{0em}{1pt}{0pt}
            \multicolumn{4}{c}{ResNet-50 Backbone} \\
            \specialrule{0em}{0pt}{1pt}
            \hline
            \specialrule{0em}{1pt}{0pt} 
            CANet$_{2019}$ \cite{canet}
            &  66.2
            & 69.6  
            & 19.0M \\            
            PGNet$_{2019}$ \cite{Zhang_2019_ICCV}
            &  69.9
            & 70.5   
            & 17.2M \\                            
            Ours
            &   \textbf{73.3}
            & \textbf{73.9}  
            & \textbf{10.8M} \\
            
            \specialrule{0em}{0pt}{1pt}
            \hline
            \specialrule{0em}{1pt}{0pt}
            \multicolumn{4}{c}{ResNet-101 Backbone} \\
            \specialrule{0em}{0pt}{1pt}
            \hline
            \specialrule{0em}{1pt}{0pt} 
            A-MCG$_{2019}$ \cite{AttantionMCG}
            & 61.2
            & 62.2  
            & 86.1M \\                   
            Ours
            & \textbf{72.9}
            & \textbf{73.5}  
            & \textbf{10.8M} \\
            \bottomrule            
        \end{tabular}
    }
    \label{tab:compare_sota_fb_simple}
\end{table}

\section{Experiments}
\label{sec:experiments}
\subsection{Implementation Details}
\noindent\textbf{Datasets}~~
We use the datasets of PASCAL-5$^i$ \cite{shaban} and COCO \cite{coco} in evaluation. PASCAL-5$^i$ is composed of PASCAL VOC 2012 \cite{pascalvoc2012} and extended annotations from SDS \cite{SDS} datasets. 20 classes are evenly divided into 4 folds $i \in \{0, 1, 2, 3\}$ and each fold contains 5 classes.
Following OSLSM \cite{shaban}, we randomly sample 1,000 query-support pairs in each test.

Following \cite{weighingboosting}, we also evaluate our model on COCO by splitting four folds from 80 classes. Thus each fold has 20 classes. The set of class indexes contained in fold $i$ is written as $\{4x - 3 + i\}$ where $x \in \{1, 2, ..., 20\}, i \in \{0,1,2,3\}$. Note that the COCO validation set contains 40,137 images (80 classes), which are much more than the images in PASCAL-5$^i$. Therefore, 1,000 randomly sampled query-support pairs used in previous work are not enough for producing reliable testing results on 20 test classes. We instead randomly sample 20,000 query-support pairs during the evaluation on each fold, making the results more stable than testing on 1,000 query-support pairs used in previous work. Stability statistics are shown in Section \ref{sec:stability}.

For both PASCAL-5$^i$ and COCO, when testing the model on one fold, we use the other three folds to train the model for cross-validation. We take the average of five testing results with different random seeds for comparison as shown in Tables \ref{tab:variance} and \ref{tab:coco_variance}.

\vspace{0.05in}
\noindent\textbf{Experimental Setting}~~
Our framework is constructed on PyTorch. We select VGG-16 \cite{vgg}, ResNet-50 \cite{resnet} and ResNet-101 \cite{resnet} as our backbones for fair comparison with other methods. The ResNet we use is the dilated version used in previous work \cite{weighingboosting,canet,AttantionMCG}. The VGG we use is the original version \cite{vgg}. All backbone networks are initialized with ImageNet \cite{imagenet} pretrained weights. Other layers are initialized by the default setting of PyTorch. We use SGD as our optimizer. The momemtum and weight decay are set to 0.9 and 0.0001 respectively. We adopt the 'poly' policy  \cite{deeplab} to decay the learning rate by multiplying $(1 - \frac{current_{iter}}{max_{iter}})^{power}$ where $power$ equals to 0.9.

Our models are trained on PASCAL-5$^i$ for 200 epochs as that of \cite{canet} with learning rate 0.0025 and batch size 4. For experiments on COCO, models are trained for 50 epochs with learning rate 0.005 and batch size 8. Parameters of the backbone network are not updated. During training, samples are processed with mirror operation and random rotation from -10 to 10 degrees. Finally, we randomly crop $473\times473$ patches from the processed images as training samples. During the evaluation, each input sample is resized to the training patch size but with respect to its original aspect ratio by padding zero, then the prediction is resized back to the original label sizes. Finally, we directly output the single-scale results without fine-tuning and any additional post-processing (such as multi-scale testing and DenseCRF \cite{densecrf}). Our experiments are conducted on an NVIDIA Titan V GPU and Intel Xeon CPU E5-2620 v4 @ 2.10GHz. The code and trained models will be made publicly available. 

\vspace{0.05in}
\noindent\textbf{Evaluation Metrics~~}
Following \cite{canet,weighingboosting}, we adopt the class mean intersection over union (mIoU) as our major evaluation metric for ablation study since the class mIoU is more reasonable than the foreground-background IoU (FB-IoU) as stated in \cite{canet}. The formulation follows $mIoU = \frac{1}{C}\sum_{i=1}^{C} IoU_{i}$,
where $C$ is the number of classes in each fold (e.g., $C=20$ for COCO and $C=5$ for PASCAL-5$^i$) and $IoU_{i}$ is the intersection-over-union of class $i$. We also report the results of FB-IoU for comparison with other methods. For FB-IoU calculation on each fold, only foreground and background are considered ($C=2$). We take average of results on all folds as the final mIoU/FB-IoU.

\begin{figure*}[t]
	\centering
	\resizebox{1.0\linewidth}{!}{
	\begin{tabular}{@{\hspace{0.5mm}}c@{\hspace{1.0mm}}c@{\hspace{1.0mm}}c@{\hspace{1.0mm}}c@{\hspace{2.0mm}}c@{\hspace{1.0mm}}c@{\hspace{1.0mm}}c@{\hspace{1.0mm}}c@{\hspace{0.0mm}}}
		\rotatebox[origin=c]{90}{\scriptsize{(a) Support}}
		\includegraphics[align=c,width=0.12\linewidth]{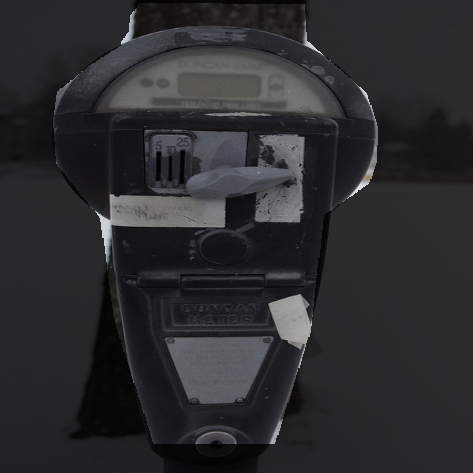}&
		\includegraphics[align=c,width=0.12\linewidth]{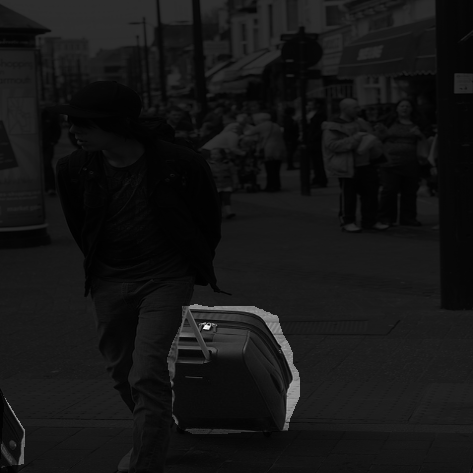}&
		\includegraphics[align=c,width=0.12\linewidth]{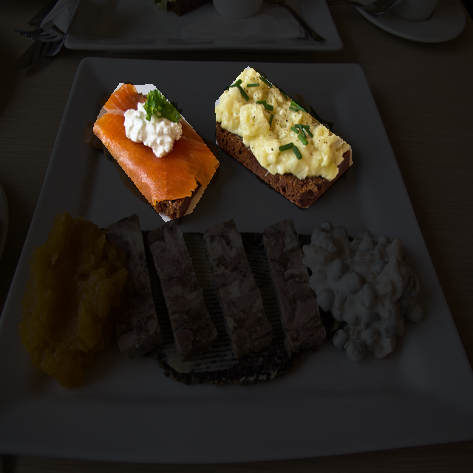}&
		\includegraphics[align=c,width=0.12\linewidth]{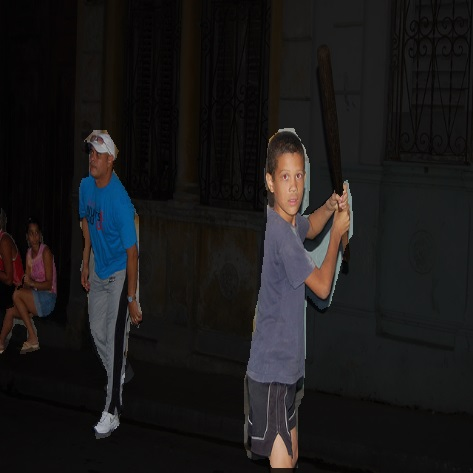}&
		\includegraphics[align=c,width=0.12\linewidth]{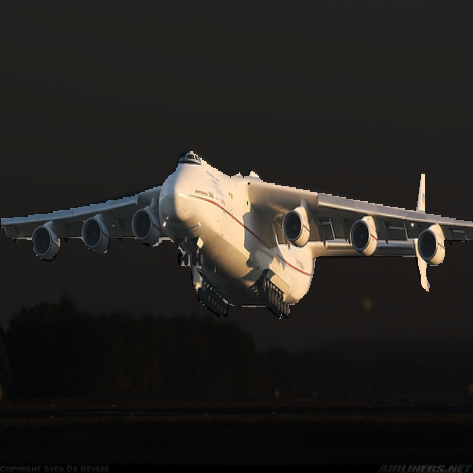}&
		\includegraphics[align=c,width=0.12\linewidth]{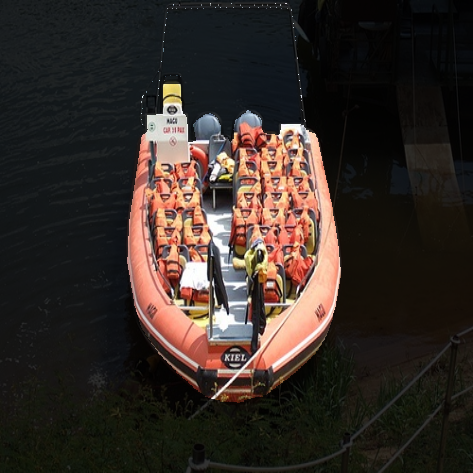}&
		\includegraphics[align=c,width=0.12\linewidth]{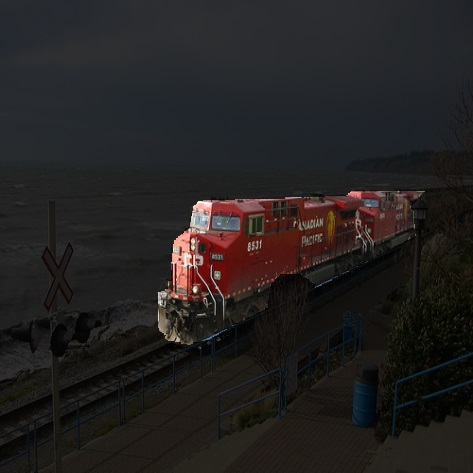}&
		\includegraphics[align=c,width=0.12\linewidth]{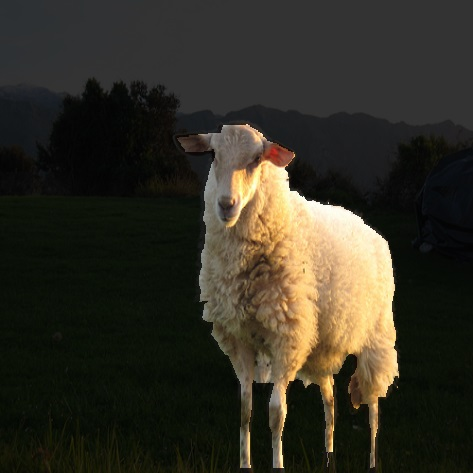}\\
		\addlinespace[3pt]
		\rotatebox[origin=c]{90}{\scriptsize{(b) Query}}
		\includegraphics[align=c,width=0.12\linewidth]{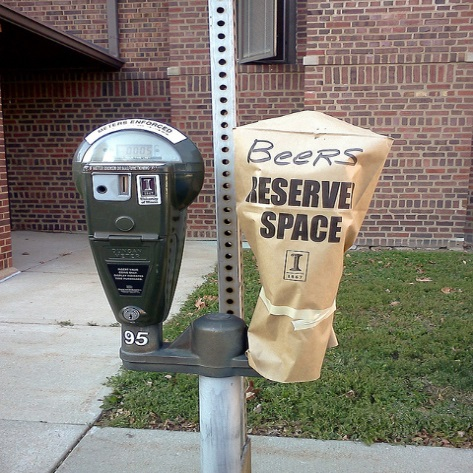}&
		\includegraphics[align=c,width=0.12\linewidth]{figure_used/compare_trans/2_input.png}&
		\includegraphics[align=c,width=0.12\linewidth]{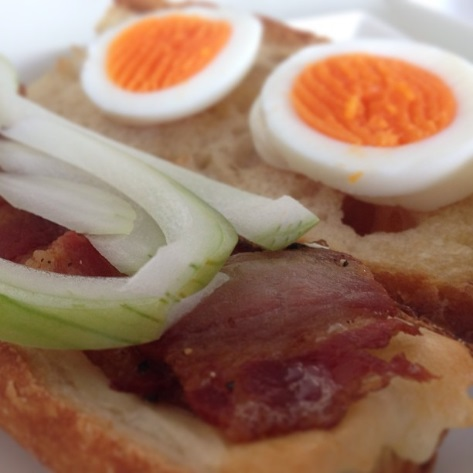}&
		\includegraphics[align=c,width=0.12\linewidth]{figure_used/compare_trans/4_input.png}&
		\includegraphics[align=c,width=0.12\linewidth]{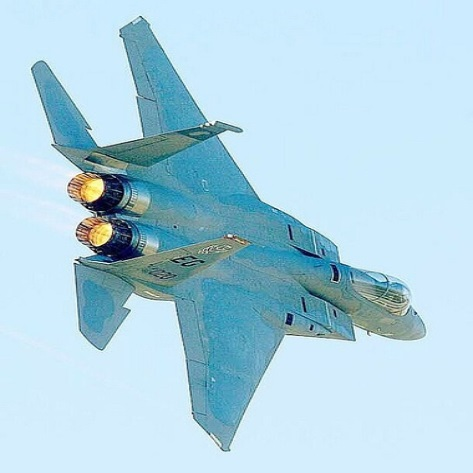}&
		\includegraphics[align=c,width=0.12\linewidth]{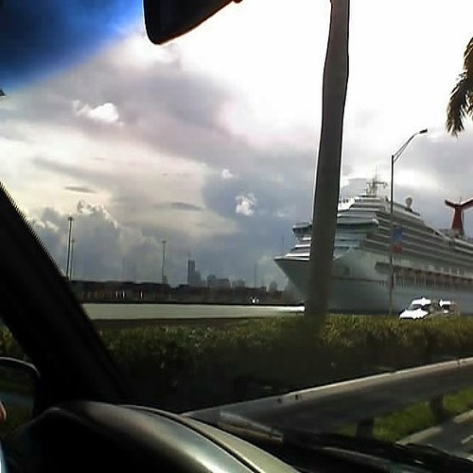}&
		\includegraphics[align=c,width=0.12\linewidth]{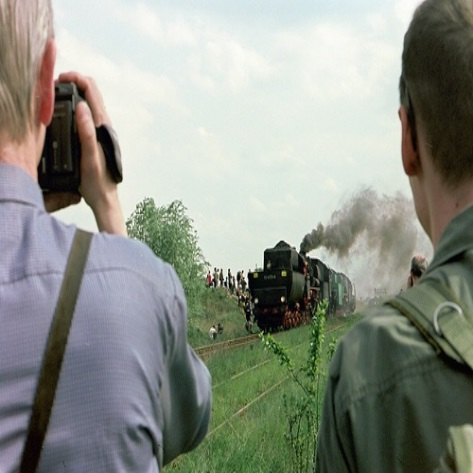}&
		\includegraphics[align=c,width=0.12\linewidth]{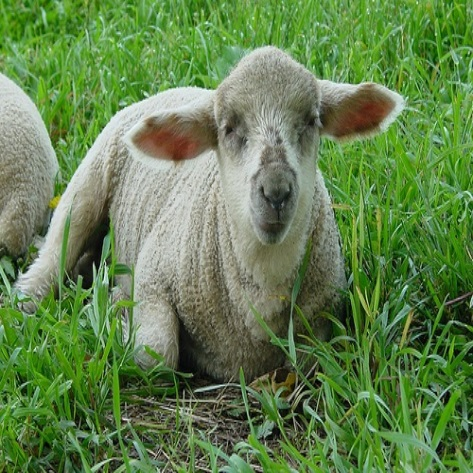}\\
		\addlinespace[3pt]
		\rotatebox[origin=c]{90}{\scriptsize{(c) GT}}
		\includegraphics[align=c,width=0.12\linewidth]{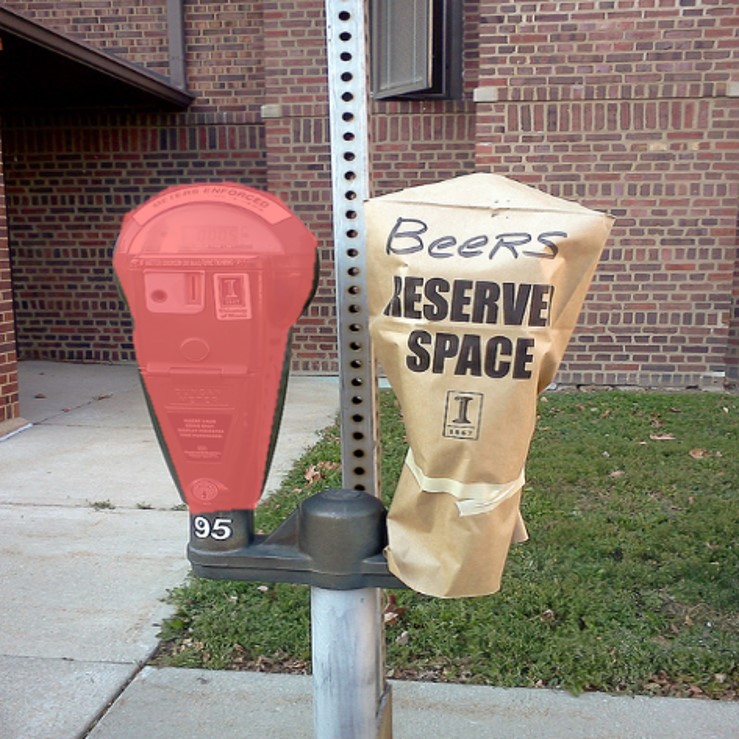}&
		\includegraphics[align=c,width=0.12\linewidth]{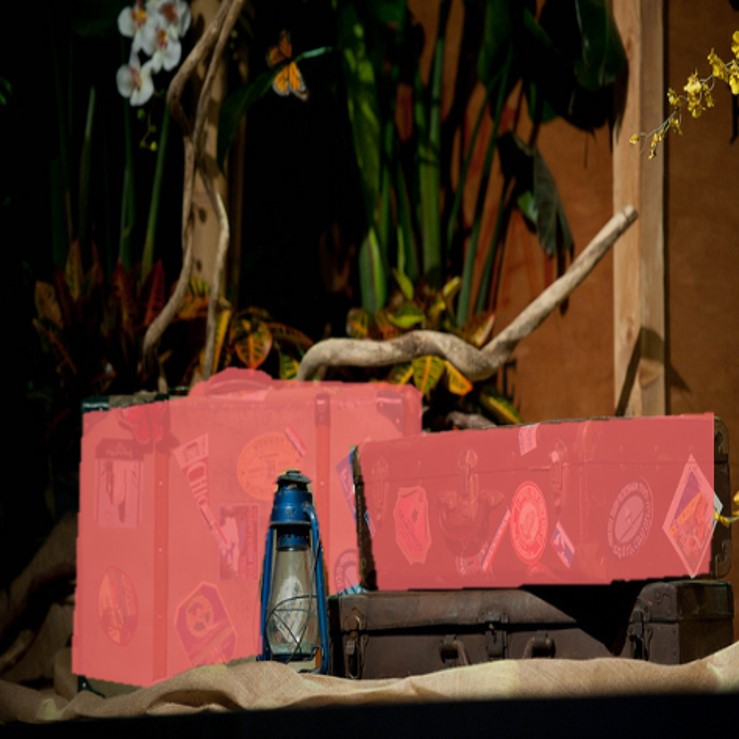}&
		\includegraphics[align=c,width=0.12\linewidth]{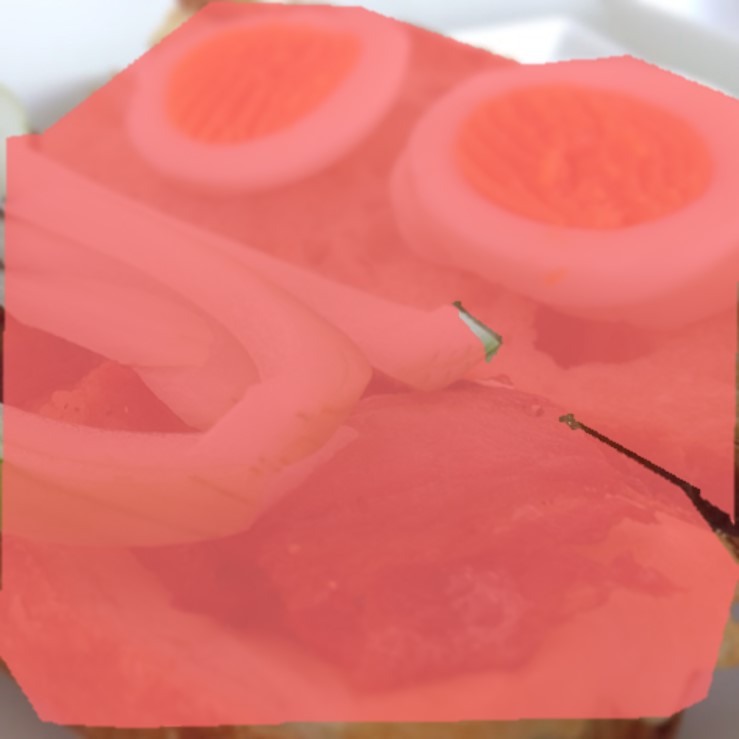}&
		\includegraphics[align=c,width=0.12\linewidth]{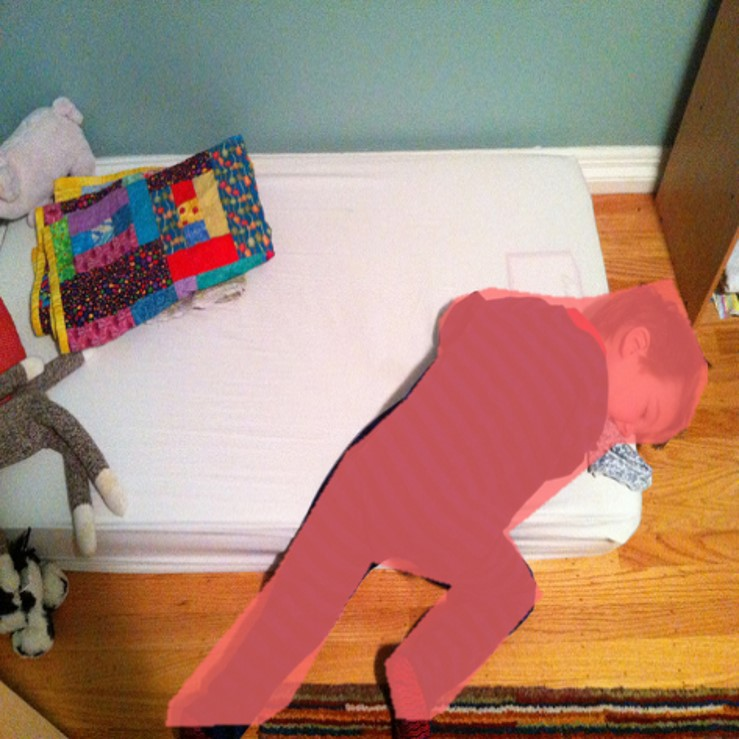}&
		\includegraphics[align=c,width=0.12\linewidth]{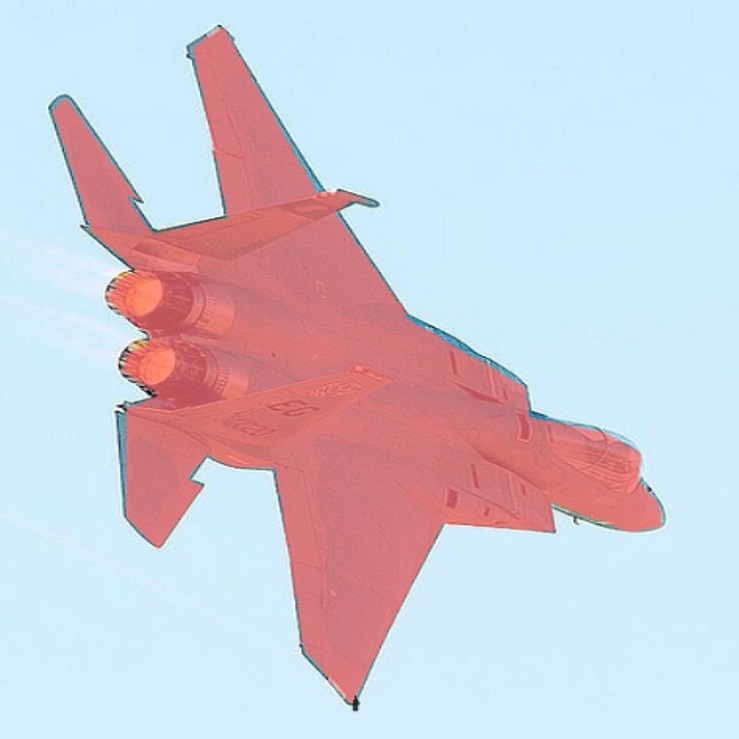}&
		\includegraphics[align=c,width=0.12\linewidth]{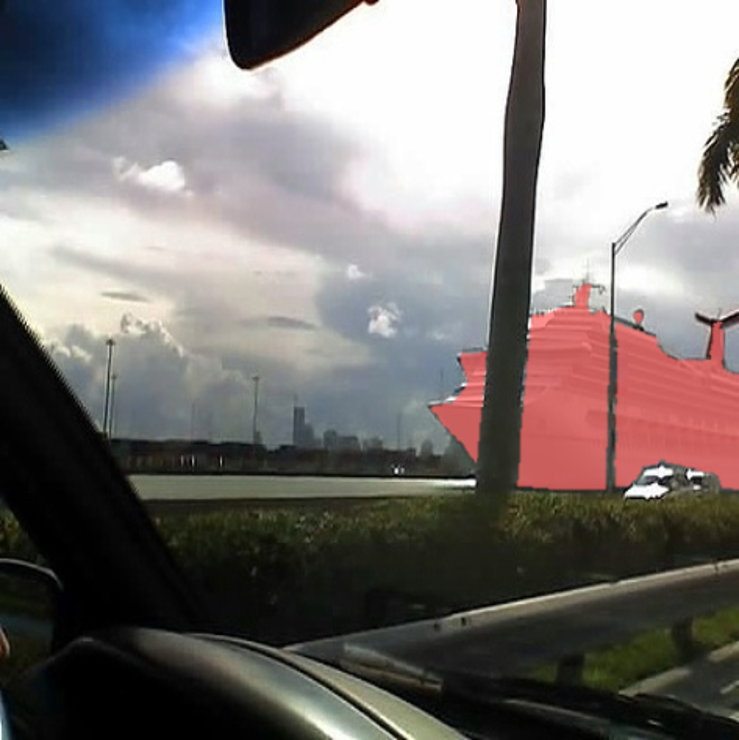}&
		\includegraphics[align=c,width=0.12\linewidth]{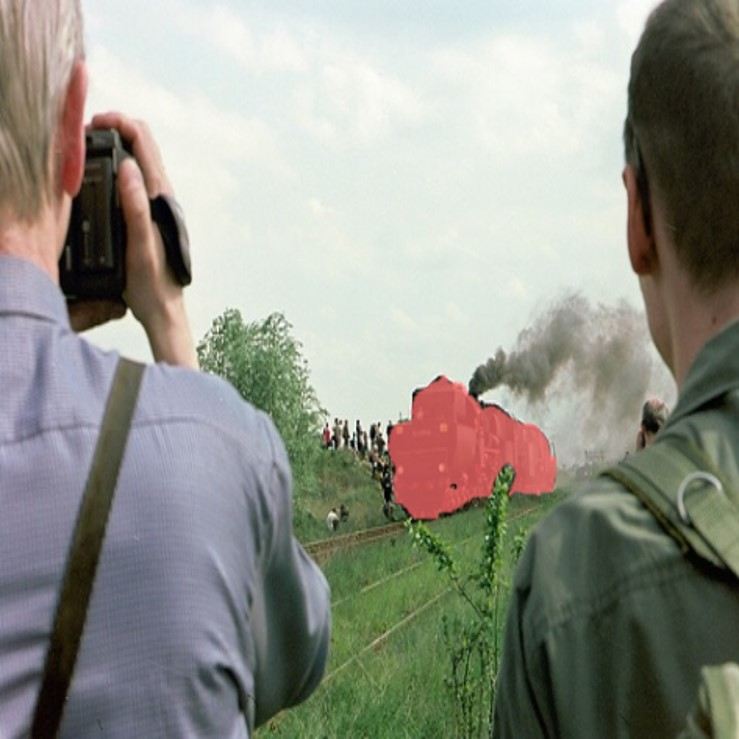}&
		\includegraphics[align=c,width=0.12\linewidth]{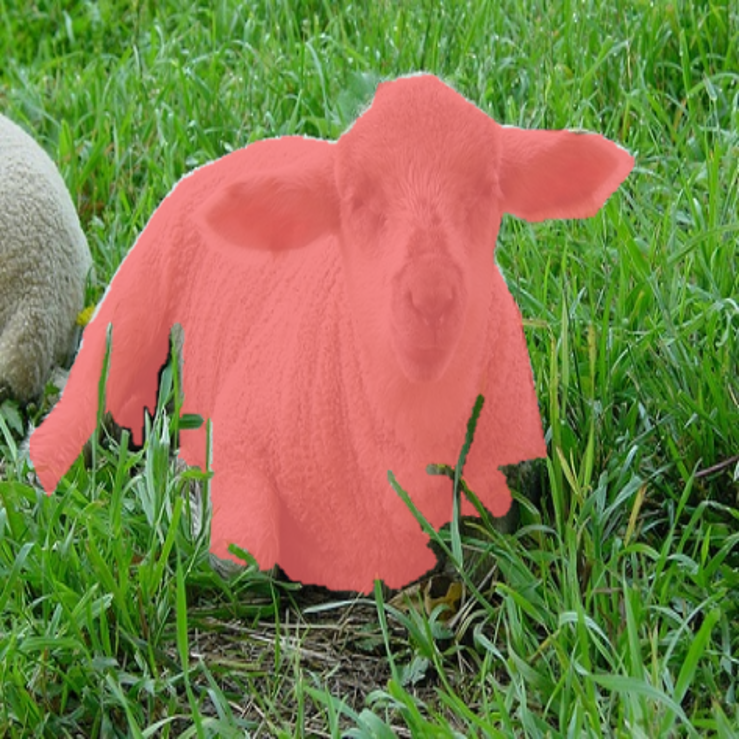}\\
		\addlinespace[3pt]
		\rotatebox[origin=c]{90}{\scriptsize{(d) Baseline}}
		\includegraphics[align=c,width=0.12\linewidth]{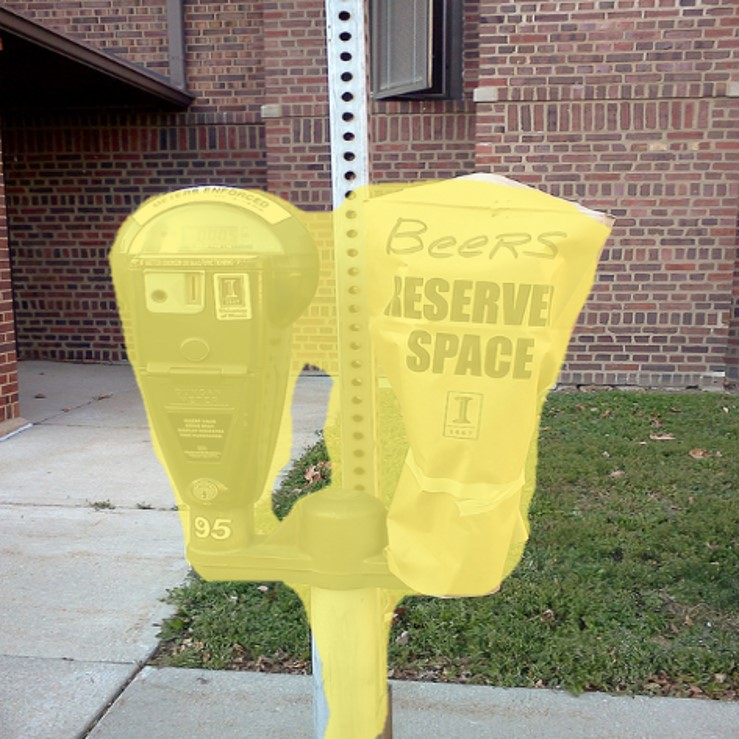}&
		\includegraphics[align=c,width=0.12\linewidth]{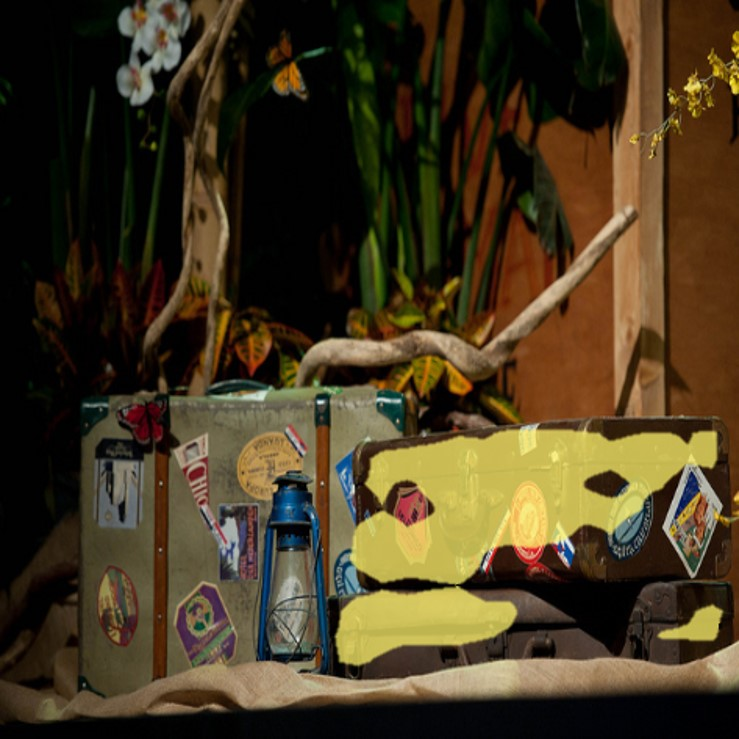}&
		\includegraphics[align=c,width=0.12\linewidth]{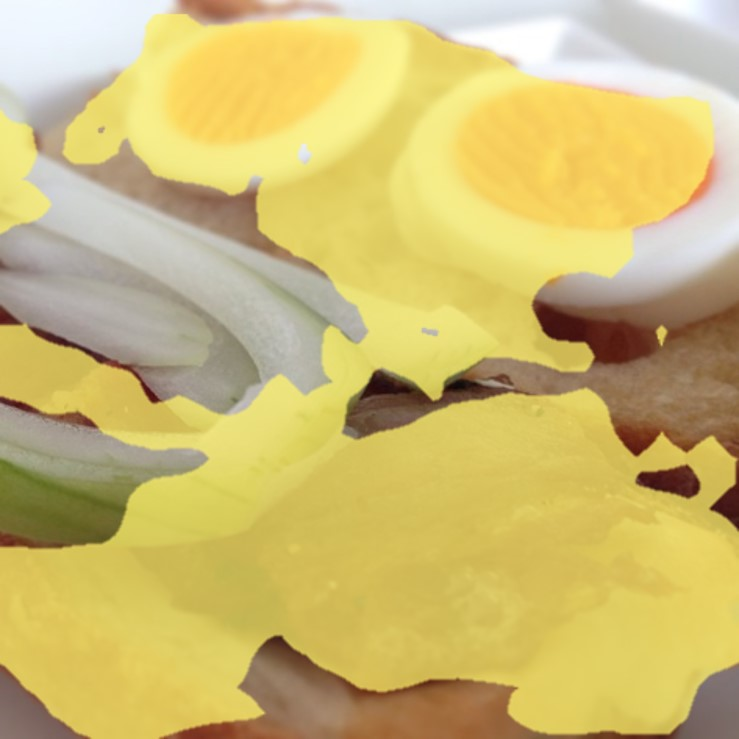}&
		\includegraphics[align=c,width=0.12\linewidth]{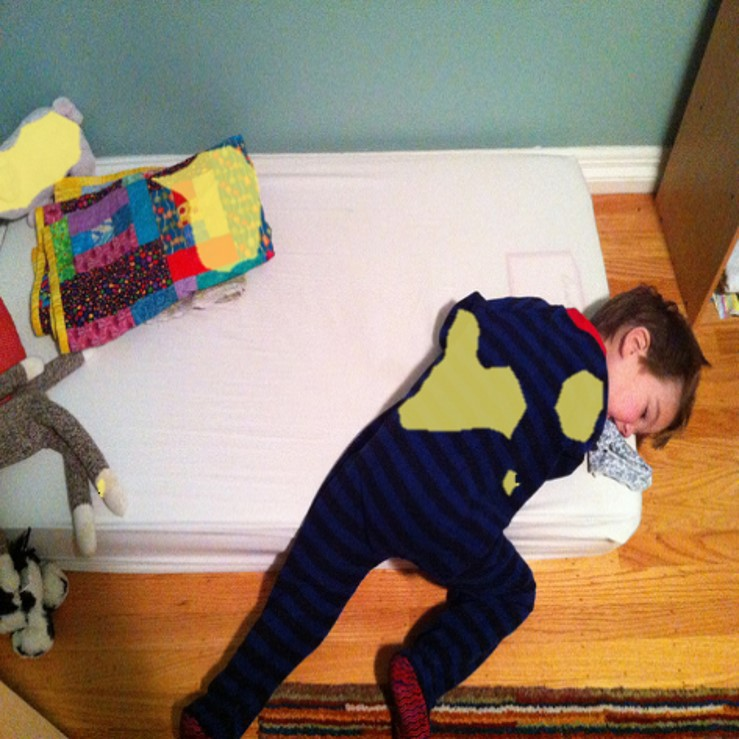}&
		\includegraphics[align=c,width=0.12\linewidth]{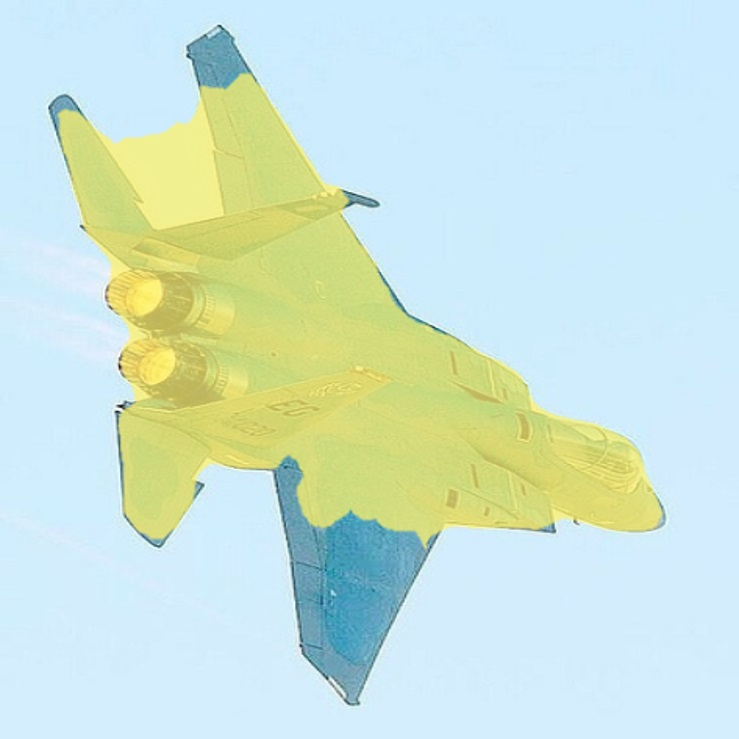}&
		\includegraphics[align=c,width=0.12\linewidth]{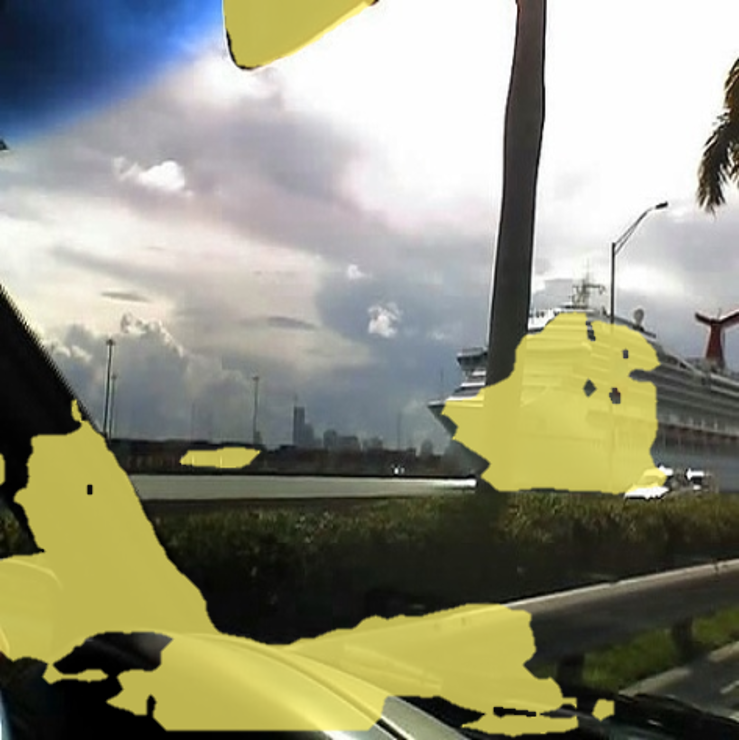}&
		\includegraphics[align=c,width=0.12\linewidth]{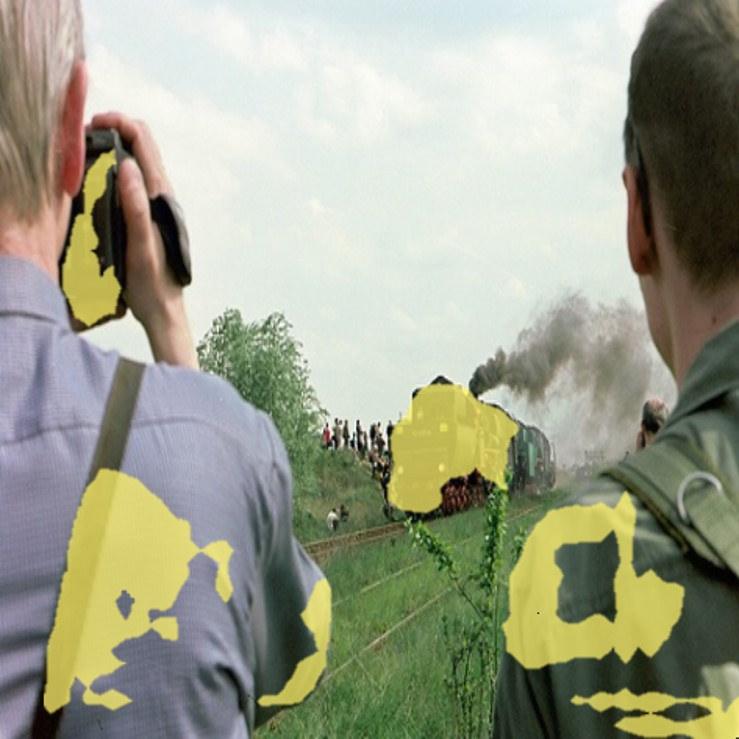}&
		\includegraphics[align=c,width=0.12\linewidth]{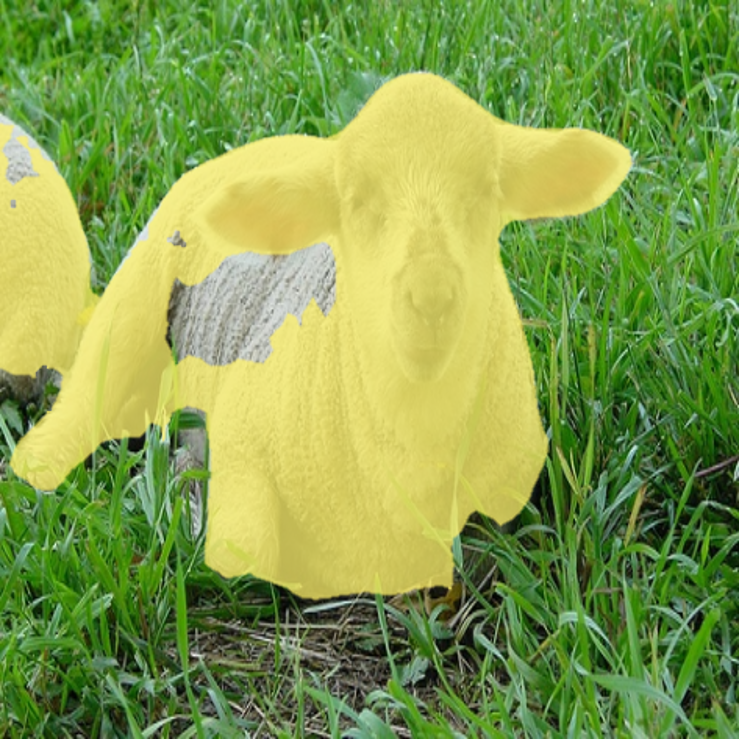}\\
		\addlinespace[3pt]
		\rotatebox[origin=c]{90}{\scriptsize{(e) PFENet}}
		\includegraphics[align=c,width=0.12\linewidth]{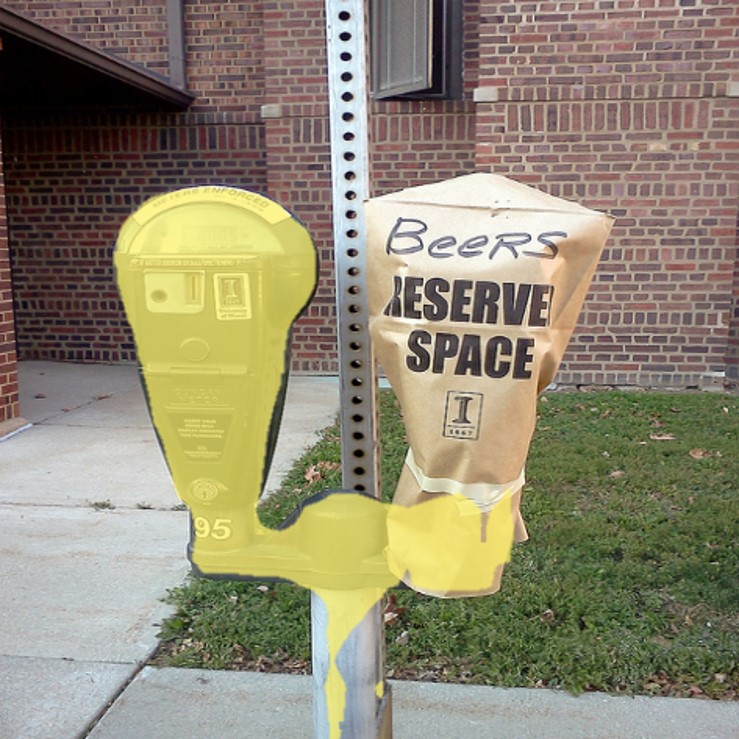}&
		\includegraphics[align=c,width=0.12\linewidth]{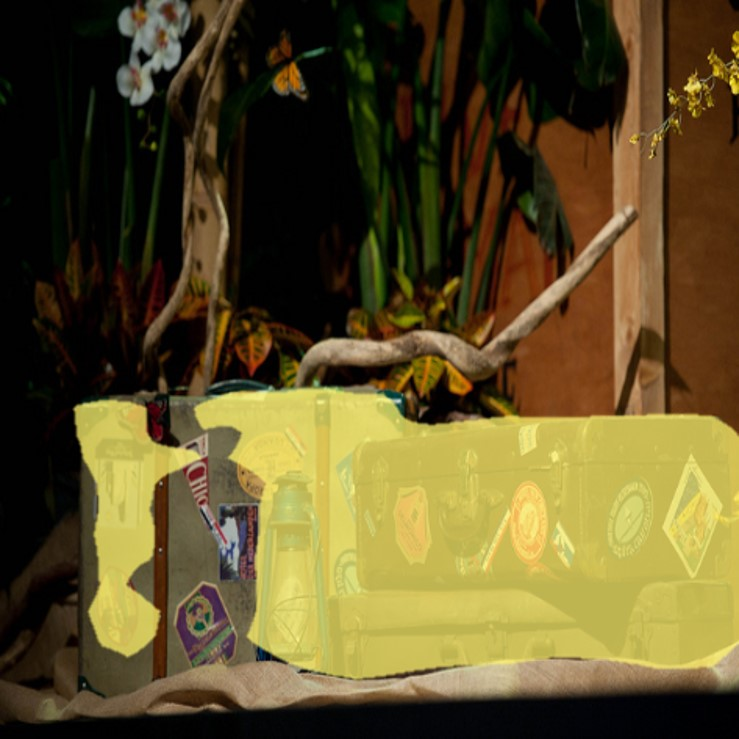}&
		\includegraphics[align=c,width=0.12\linewidth]{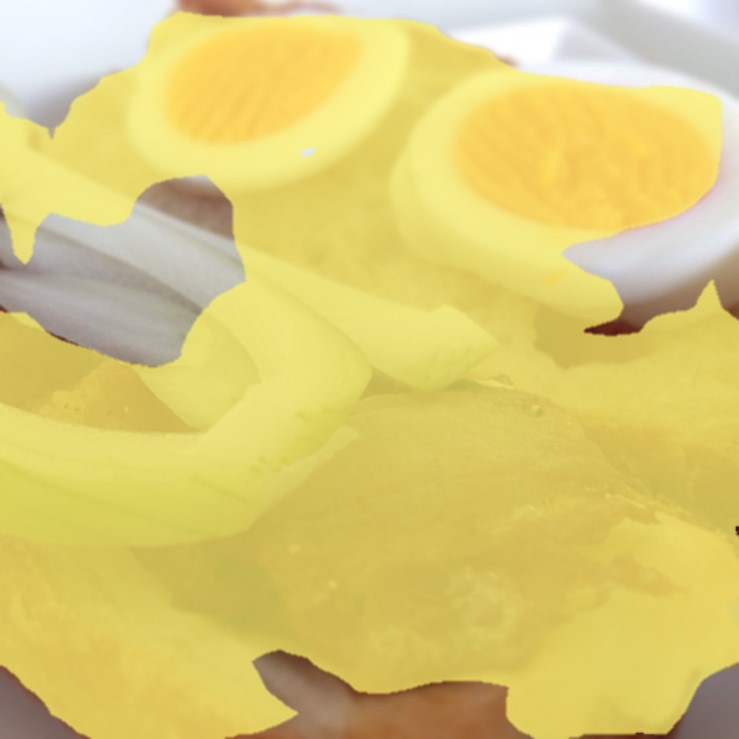}&
		\includegraphics[align=c,width=0.12\linewidth]{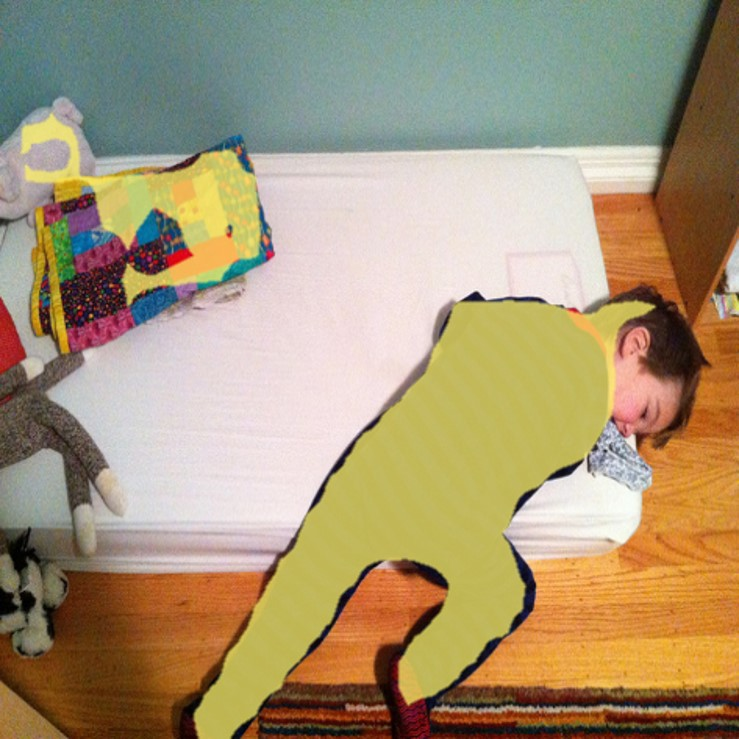}&
		\includegraphics[align=c,width=0.12\linewidth]{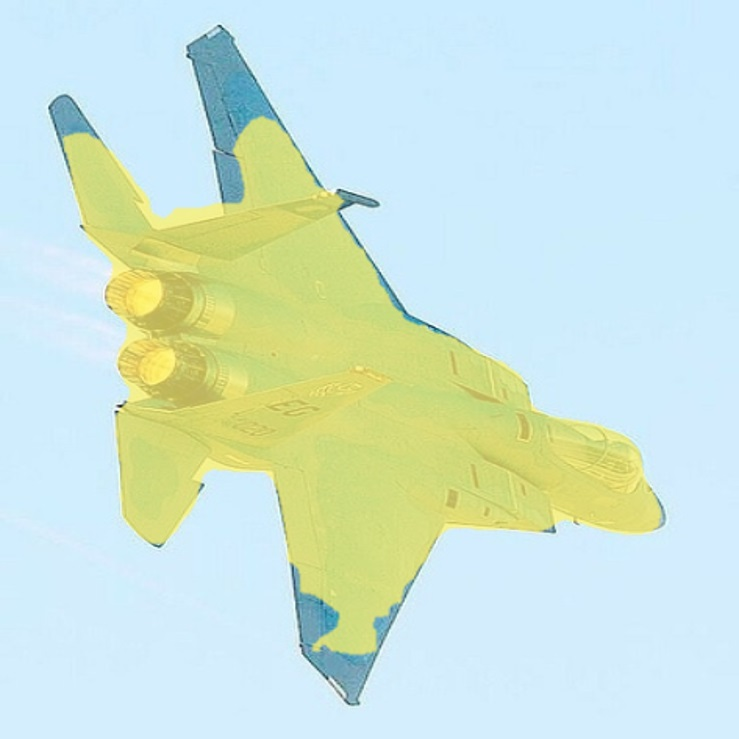}&
		\includegraphics[align=c,width=0.12\linewidth]{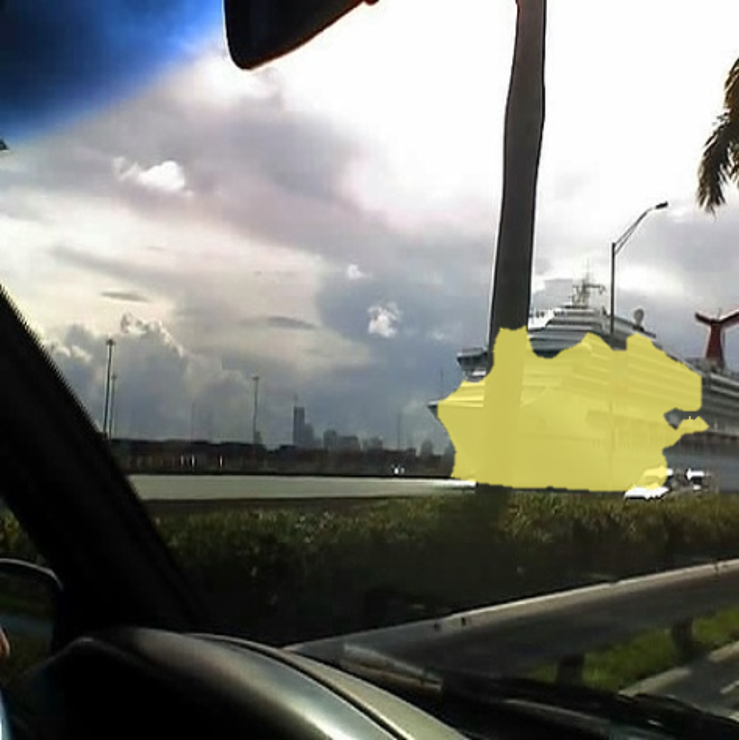}&
		\includegraphics[align=c,width=0.12\linewidth]{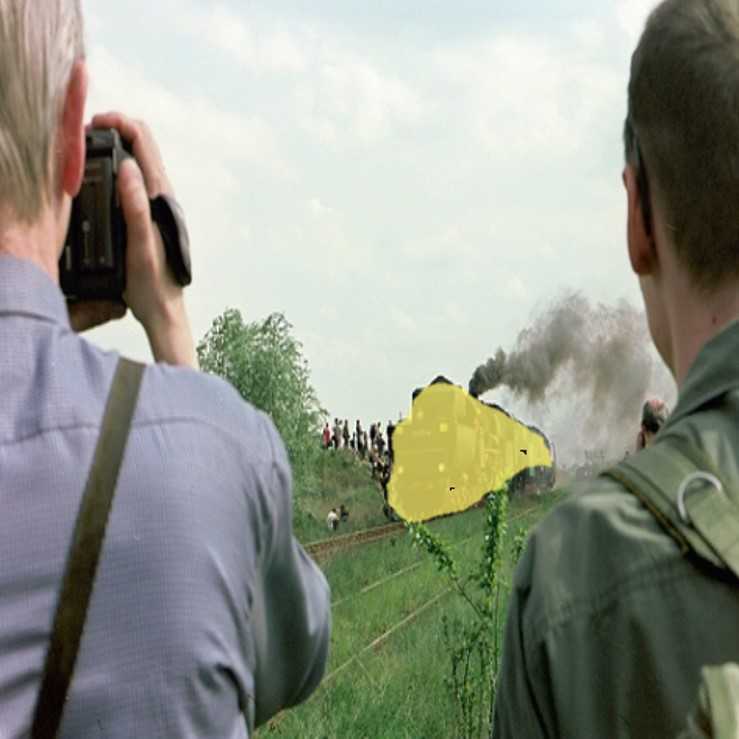}&
		\includegraphics[align=c,width=0.12\linewidth]{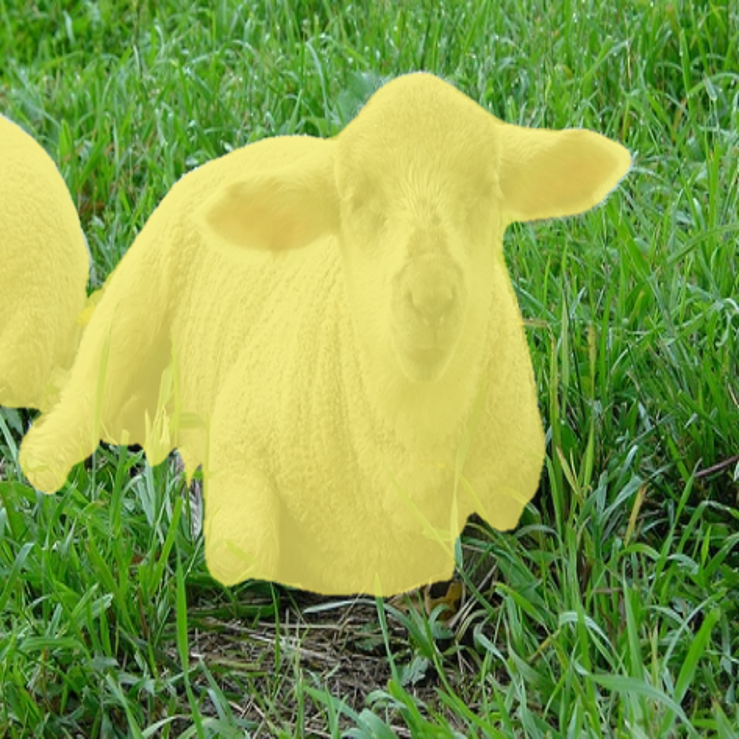}\\
	\end{tabular}}
	\caption{Qualitative results of the proposed PFENet and the baseline. The left samples are from COCO and the right ones are from PASCAL-5$^i$. From top to bottom: (a) support images, (b) query images, (c) ground truth of query images, (d) predictions of baseline, (e) predictions of PFENet.}
	\label{fig:visual_compare}
\end{figure*}

\begin{table*}
    \caption{Class mIoU / FB-IoU results on COCO. Models with $\dagger$ are evaluated on the labels resized to a fixed training crop size (473 for our models). Models without $\dagger$ are tested on labels with the original sizes.}
    \centering
    {
        \begin{tabular}{ c | c | c  c  c  c | c |  c  c  c  c | c }
            \toprule
            \multirow{2}{*}{\textit{Methods}} 
            & \multirow{2}{*}{\textit{Backbone}} 
            & \multicolumn{5}{c|}{1-Shot} 
            & \multicolumn{5}{c}{5-Shot}  \\ 
            \specialrule{0em}{0pt}{1pt}
            \cline{3-12}
            \specialrule{0em}{1pt}{0pt} 
            & & Fold-0 & Fold-1 & Fold-2 & Fold-3 & Mean
            & Fold-0 & Fold-1 & Fold-2 & Fold-3 & Mean \\
            
            \specialrule{0em}{0pt}{1pt}
            \hline
            \specialrule{0em}{1pt}{0pt}
            \multicolumn{11}{c}{Class mIoU Evaluation} \\
            \specialrule{0em}{0pt}{1pt}
            \hline
            \specialrule{0em}{1pt}{0pt}
            FWBF$_{2019}$ \cite{weighingboosting} & VGG-16
            & 18.4 & 16.7 & 19.6 & 25.4 & 20.0 
            & 20.9 & 19.2 & 21.9 & 28.4 & 22.6 \\
            Ours  & VGG-16
            & \textbf{33.4} & \textbf{36.0} & \textbf{34.1} & \textbf{32.8} & \textbf{34.1} 
            & \textbf{35.9} & \textbf{40.7} & \textbf{38.1} & \textbf{36.1} & \textbf{37.7}  \\   
            \specialrule{0em}{0pt}{1pt}
            \hline 
            \specialrule{0em}{1pt}{0pt}                           
            PANet$_{2019}\dagger$ \cite{panet} & VGG-16
            & - & - & - & - & 20.9 
            & - & - & - & - & 29.7  \\             
            Ours$\dagger$  & VGG-16
            & \textbf{35.4}  & \textbf{38.1}  & \textbf{36.8}  & \textbf{34.7}  & \textbf{36.3}
            & \textbf{38.2}  & \textbf{42.5}  & \textbf{41.8}  & \textbf{38.9}  &  \textbf{40.4} \\ 
            
            \specialrule{0em}{0pt}{1pt}
            \hline 
            \specialrule{0em}{1pt}{0pt}   
            
            FWBF$_{2019}$ \cite{weighingboosting} & ResNet-101
            & 19.9 & 18.0 & 21.0 & 28.9 & 21.2 
            & 19.1 & 21.5 & 23.9 & 30.1 & 23.7 \\
            Ours  & ResNet-101
            & \textbf{34.3} & \textbf{33.0} & \textbf{32.3} & \textbf{30.1} & \textbf{32.4}
            & \textbf{38.5} & \textbf{38.6} & \textbf{38.2} & \textbf{34.3} & \textbf{37.4} \\    
            \specialrule{0em}{0pt}{1pt}
            \hline 
            \specialrule{0em}{1pt}{0pt}                           
            Ours$\dagger$  & ResNet-101
            &  36.8 & 41.8  & 38.7  & 36.7  & 38.5
            &  40.4 & 46.8  & 43.2  & 40.5  & 42.7  \\

            \specialrule{0em}{0pt}{1pt}
            \hline
            \specialrule{0em}{1pt}{0pt}
            \multicolumn{11}{c}{FB-IoU Evaluation} \\
            \specialrule{0em}{0pt}{1pt}
            \hline
            \specialrule{0em}{1pt}{0pt}
                 
            PANet$_{2019}\dagger$ \cite{panet} & VGG-16
            & - & - & - & - & 59.2 
            & - & - & - & - & 63.5  \\             
            Ours$\dagger$  & VGG-16
            &  \textbf{53.3} & \textbf{66.1}  & \textbf{66.6}  & \textbf{67.1}  & \textbf{63.3}               
            & \textbf{53.5}  & \textbf{68.3}  & \textbf{68.2}  & \textbf{70.1}  & \textbf{65.0}\\  
            \specialrule{0em}{0pt}{1pt}
            \hline
            \specialrule{0em}{1pt}{0pt}            
            Ours & VGG-16
            & 50.0 & 63.1 & 63.5 & 63.4 & 60.0 
            & 50.3 & 65.2 & 65.2 & 65.5 & 61.6  \\                               
            \specialrule{0em}{0pt}{1pt}
            \hline 
            \specialrule{0em}{1pt}{0pt}                  
            A-MCG$_{2019}$ \cite{AttantionMCG} & ResNet-101
            & - & - & - & - & 52.0
            & - & - & - & - & 54.7  \\          
            Ours & ResNet-101
            & \textbf{52.2} & \textbf{59.5} & \textbf{61.5} & \textbf{61.4} & \textbf{58.6} 
            & \textbf{51.5} & \textbf{65.6} & \textbf{65.7} & \textbf{64.7} & \textbf{61.9}  \\     
           \specialrule{0em}{0pt}{1pt}
            \hline 
            \specialrule{0em}{1pt}{0pt}                           
            Ours$\dagger$  & ResNet-101
            & 51.6  & 65.9  & 66.6  & 66.0 & 63.0
            & 52.3  & 70.0  & 69.5  &  71.3  & 65.8 \\             
        
            \bottomrule            
        \end{tabular}
    }
    \label{tab:coco_compare_sota_class}
\end{table*}

\begin{table*}
	\caption{Class mIoU results of different ways for inter-scale interaction on PASCAL-5$^i$. All models in this table are based on ResNet-50 and are trained and tested with prior masks.
		\textbf{W/O}: FEM without the information path for inter-scale interaction.
		\textbf{TD}: FEM with top-down information path.
		\textbf{BU}: FEM with bottom-up information path.
		\textbf{TD+BU}: FEM with top-down + bottom-up information path.
		\textbf{BU+TD}: FEM with bottom-up + top-down information path.  
	}
	\centering
	{
		\begin{tabular}{ l |  c  c  c  c | c |  c  c  c  c | c }
			\toprule
			\multirow{2}{*}{\textit{Methods}} 
			& \multicolumn{5}{c|}{1-Shot} 
			& \multicolumn{5}{c}{5-Shot}  \\ 
			\specialrule{0em}{0pt}{1pt}
			\cline{2-11}
			\specialrule{0em}{1pt}{0pt} 
			& Fold-0 & Fold-1 & Fold-2 & Fold-3 & Mean
			& Fold-0 & Fold-1 & Fold-2 & Fold-3 & Mean \\
			\hline 
			W/O
			& 60.5  & 68.4  & 55.4  & 54.9  & 59.8
			& 62.8  & 68.9  & 55.6  & 56.5  & 61.0  \\  
			TD  			
			& 61.7  & 69.5  & 55.4  & 56.3  & \textbf{60.8} 
			& \textbf{63.1} & \textbf{70.7}  & 55.8  &  57.9 & 61.9  \\     
			BU
			& \textbf{62.4}  & 69.2  & 53.9  & 55.9  & 60.4
			& \textbf{63.1}  & 70.1  & 53.7  & 56.0  & 60.7  \\     
			TD+BU
			& 61.0  & \textbf{69.7}  & \textbf{55.6}  & \textbf{57.0}  & \textbf{60.8}
			& 62.4  & 70.4  & \textbf{56.4}  & \textbf{58.9}  & \textbf{62.0}  \\  
			BU+TD
			& 61.0  & 68.9  & 54.8  & 56.0  & 60.2
			& 62.4  & 69.8  & 54.5  & 56.7  & 60.8  \\   
			\bottomrule                              
		\end{tabular}
	}
	\label{tab:ablation_inter_scale}
\end{table*}

\begin{table*}
	\caption{Class mIoU of FEM with different spatial sizes and the comparison with PPM \cite{pspnet} and ASPP \cite{aspp} on PASCAL-5$^i$. The backbone is ResNet-50.  `$\{60, 30, 15, 8\}$': the input query feature is average-pooled into four scales $\{60, 30, 15, 8\}$ and concatenate with the expanded support features respectively as shown in Figure \ref{fig:FEM}.
		\textbf{WO}: without inter-scale interaction.
	}
	\centering
	{
		\begin{tabular}{ l |  c  c  c  c | c |  c  c  c  c | c }
			\toprule
			\multirow{2}{*}{\textit{Methods}} 
			& \multicolumn{5}{c|}{1-Shot} 
			& \multicolumn{5}{c}{5-Shot}  \\ 
			\specialrule{0em}{0pt}{1pt}
			\cline{2-11}
			\specialrule{0em}{1pt}{0pt} 
			& Fold-0 & Fold-1 & Fold-2 & Fold-3 & Mean
			& Fold-0 & Fold-1 & Fold-2 & Fold-3 & Mean \\
			\hline 
			\{60\} (Baseline)
			&  54.3 & 67.3 & 53.3  & 50.4 & 56.3 
			& 57.1 & 68.0 & 53.8 & 52.9 & 58.0  \\  
			\{60\} + PPM \cite{pspnet}
			& 55.4 & 68.4 & 53.2 & 51.4 & 57.1
			& 58.3 & 68.9 & 53.5 & 50.8 & 57.9  \\    
			\{60\} + ASPP \cite{aspp}
			& 57.6 & 68.4 & 52.8 & 49.0 & 56.9
			& 59.5 & 69.3 & 52.6 & 50.7 & 58.0  \\     
			
			\{60, 6, 3, 2, 1\}
			& 58.8 & 68.0 & 54.1 & 51.2 & 58.0
			& 59.8 & 68.4 & 53.8 & 52.1 & 58.5  \\  
			\{60, 30\}
			& 55.3 & 67.8  & \textbf{54.7} & 51.2 & 57.3 
			& 58.4 & 68.7 & 54.5 & 53.1 & 58.7  \\  
			\{60, 30, 15\}
			& 56.6 & 68.0 & 54.6 & 52.9 & 58.0 
			& 59.0 & 68.7 & 55.0 & 54.0 & 59.2  \\  
			\{60, 30, 15, 8\}
			& \textbf{59.4} & \textbf{68.9} & \textbf{54.7} & 53.6 & \textbf{59.2}
			& \textbf{61.5} & \textbf{69.5} & \textbf{55.4} & 55.3 & \textbf{60.4}  \\  
			\{60, 30, 15, 8, 4\}
			& 58.7 & 68.5 & 54.1 & \textbf{54.5} & 58.9
			& 60.3 & 69.3 & 54.9 & \textbf{56.4} & 60.2  \\  
			\{60, 30, 15, 8\}-WO
			& 57.9 & 67.4 & 53.7 & 53.6 & 58.2
			& 60.5 & 68.0 & 54.2 & 53.8 & 59.1  \\                               
			\bottomrule            
		\end{tabular}
	}
	\label{tab:ablation_FEM_simple}
\end{table*}

\subsection{Results}
As shown in Tables \ref{tab:compare_sota_class}, \ref{tab:compare_sota_fb_simple} and \ref{tab:coco_compare_sota_class}, we build our models on three backbones VGG-16, ResNet-50 and ResNet-101 and report the mIoU/FB-IoU results respectively. By incorporating the proposed prior mask and FEM, our model significantly outperforms previous methods, reaching new state-of-the-art on both PASCAL-5$^i$ and COCO datasets. The PFENet can even outperform other methods on COCO with more than 10 points in terms of class mIoU. Our performance advantage on FB-IoU compared to PANet is relatively smaller than class mIoU on COCO, because FB-IoU is biased towards the background and classes that cover a large part of the foreground area.  It is worth noting that our PFENet achieves the best performance with the fewest learnable parameters (10.4M for VGG based model and 10.8M for ResNet based models). Qualitative results are shown in Figure \ref{fig:visual_compare}.

\subsection{Ablation Study of FEM~~}
\label{sec:ablation_study}
The proposed feature enrichment module (FEM) adaptively enriches the query feature by merging with support features in different scales and utilizes an inter-scale path to vertically transfer useful information from the auxiliary features to the main features. To verify the effectiveness of FEM, we first compare different strategies for inter-scale interaction. It shows that the top-down information path brings a decent performance gain to the baseline without compromising the model size much. Then experiments with different designs for inter-source enrichment are presented followed by comparison with the other feature enrichment designs of HRNet \cite{hrnet_pami}, ASPP \cite{aspp} and PPM \cite{pspnet}. We also compare the Graph Attention Unit (GAU) used in the recent state-of-the-art few-shot segmentation method PGNet \cite{Zhang_2019_ICCV} to refine the query feature. In these experiments, since our input images are resized to 473 $\times$ 473, the input feature map of the module (e.g., FEM, GAU) has the spatial size 60 $\times$ 60.

\subsubsection{Inter-Scale Interaction Strategies}
\label{sec:inter_scale_ablation}
In this section, we show experimental results and analysis on different vertical inter-scale interaction strategies to manifest the rationales behind our designs of FEM.

As mentioned in Section \ref{sec:fem_intro}, there are four alternatives for the inter-scale interaction: top-down path (TD), bottom-up path (BU), top-down + bottom-up path (TD+BU), and bottom-up + top-down path (BU+TD). Our experimental results in Table \ref{tab:ablation_inter_scale} show that TD and TD+BU help the basic FEM structure without (W/O) the information path accomplish better results than both BU and BU+TD. The model with TD+BU contains more learnable parameters (16.0M) than TD (10.8M), and yet yields comparable performance. We thus choose TD for inter-scale interaction.

These experiments prove that using the finer feature (auxiliary) to provide additional information to the coarse feature (main) is more effective than using the coarse feature (auxiliary) to refine the finer feature (main). It is because the coarse features are not sufficient for targeting the query classes during the later information concentration stage if the target object disappears in small scales. 
 
Different from common semantic segmentation where contextual information is the key for good performance, the way of representation and acquisition of query information is more important in few-shot segmentation. Our motivation for designing FEM is to match the query and support features in different scales to tackle the spatial inconsistency between the query and support samples. Thus, a down-sampled coarse query feature without target information is less helpful for improving the quality of the final prediction as shown in the experiments comparing TD and BU.

\subsubsection{Comparison with Other Designs}
PPM \cite{pspnet} and ASPP \cite{aspp} are two popular feature enrichment modules for semantic segmentation by providing multi-resolution context, and HRNet \cite{hrnet_pami,hrnet,hrnet_arxiv} provides a new feature enrichment module for
the segmentation task -- it achieved SOTA results on
semantic segmentation benchmarks.
In few-shot segmentation, the Graph Attention Unit (GAU) has been used in PGNet \cite{Zhang_2019_ICCV} to refine the query feature with contextual information. We note the proposed FEM module yields even better few-shot segmentation performance.

The improvement brought by FEM stems from: \textbf{1)} the fusions of query and support features in different spatial sizes (inter-source enrichment) since it encourages the following convolution blocks to process the concatenated features independently in different spatial resolutions, which is beneficial to predicting query targets in various scales; \textbf{2)} the inter-scale interaction that selectively passes useful information from the auxiliary feature to supplement the main feature. The model without the vertical top-down information path (marked with WO) yields worse results in Table \ref{tab:ablation_FEM_simple}. 

We implement the ASPP with dilation rates $\{1, 6, 12, 18\}$ and it achieves close results to PPM. The dilated convolution is less effective than adaptive average pooling for few-shot segmentation \cite{Zhang_2019_ICCV}. In the following, we mainly make comparisons with PPM and GAU first since they both use the adaptive pooling to provide multi-scale information. Then, we make a discussion with the module proposed
by HRNet.

\vspace{0.05in}
\noindent\textbf{Pyramid Pooling Module (PPM)~~}
As shown in Table \ref{tab:ablation_FEM_simple}, the model with spatial sizes $\{60, 30, 15, 8\}$ achieves better performance than the baseline (original size with spatial size $\{60\}$) and models that replace FEM with PPM and ASPP. 
Experiments of PSPNet \cite{pspnet} show that the Pyramid Pooling Module (PPM) with spatial sizes $\{6, 3, 2, 1\}$ yields the best performance. When small spatial sizes are applied to FEM, it still outperforms PPM. But small spatial sizes are not optimal in FEM because the features pooled to spatial sizes like $\{6, 3, 2, 1\}$ are too coarse for interaction and fusion of query and support features. Similarly, with small spatial size 4, the FEM with $\{60, 30, 15, 8, 4\}$ yields inferior performance compared to using the model with spatial sizes $\{60, 30, 15, 8\}$. Hence, we select $\{60, 30, 15, 8\}$ as the feature scales for the inter-source enrichment of  FEM.

\vspace{0.05in}
\noindent\textbf{Graph Attention Unit (GAU)~~}
GAU \cite{Zhang_2019_ICCV} uses the graph attention mechanism to establish the element-to-element correspondence between the query and support features in each scale. Pixels of the support feature are weighed by the GAU and the new support feature is the weighted sum of the original support feature. Then the new support feature is concatenated with the query feature for further processing. 

We directly replace the FEM with GAU on our baseline and keep other settings for a fair comparison. GAU is implemented with the code provided by the authors. Our baseline with GAU achieves class mIoU 55.4 and 56.1 in 1- and 5-shot evaluation respectively. Noticing the original feature scales in GAU are $\{60, 8, 4\}$, we also implement it with scales $\{60, 30, 15, 8\}$ (denoted as GAU+) used in our FEM. GAU+ yields smaller mIoU than GAU (54.9 in 1-shot and 55.4 in 5-shot). Though GAU also forms a pyramid structure via adaptive pooling to capture the multi-level semantic information, it is less competitive than the proposed FEM (59.2 in 1-shot and 60.4 in 5-shot) 
because it misses the hierarchical inter-scale relationship that adaptively provides information extracted from other levels to help refine the merged feature.

\begin{figure}[t]
\centering
    \begin{minipage}   {1.0\linewidth}
        \centering
        \includegraphics 
        [width=1.0\linewidth]{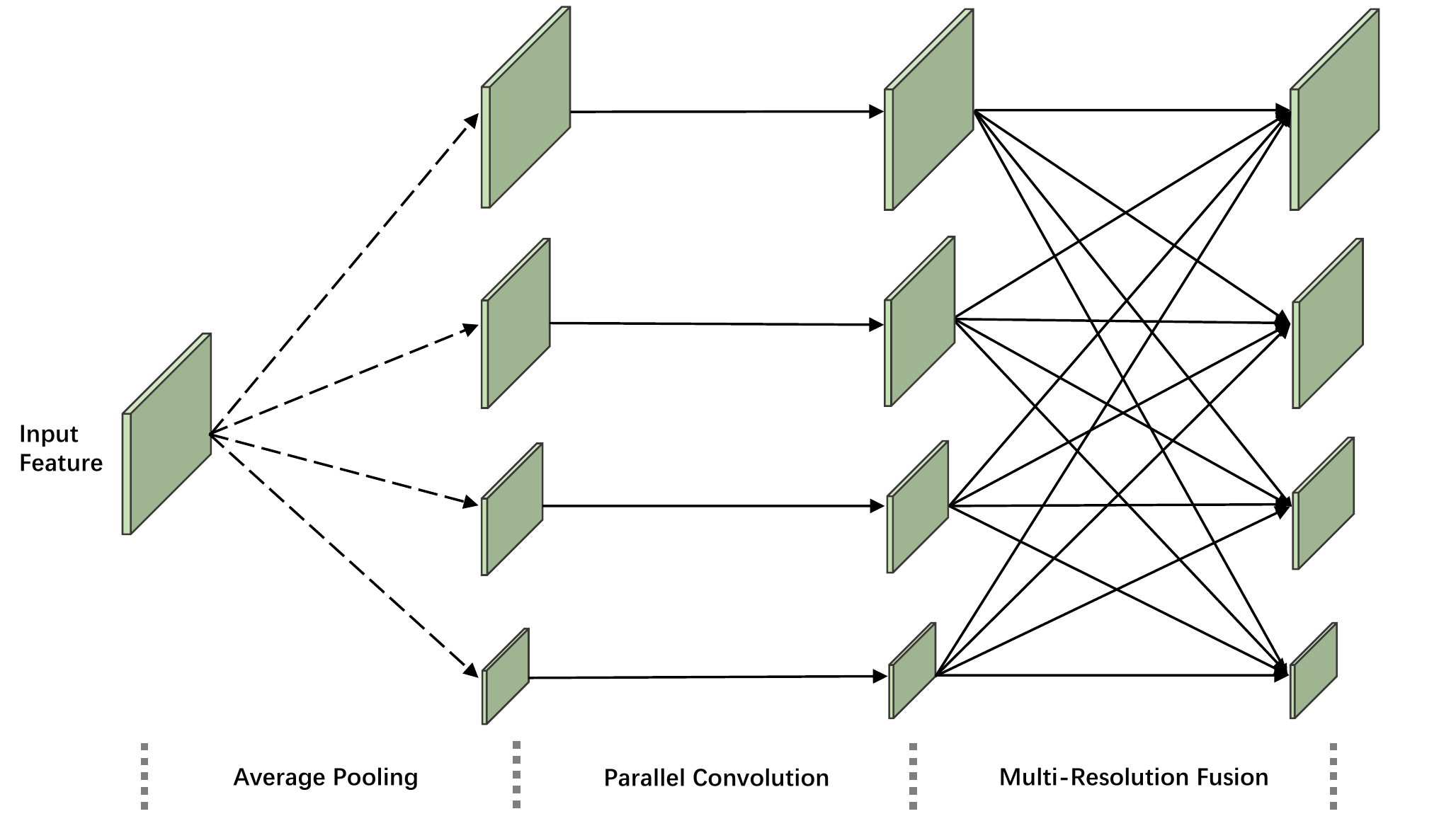}
    \end{minipage}
    \caption{Modularized block of HRNet (HRB) that applies dense multi-resolution fusions.}
    \label{fig:hrb}
\end{figure}

\vspace{0.05in}
\noindent\textbf{High-Resolution Network (HRNet)~~}
HRNet has shown its superiority on many vision tasks by maintaining a high-resolution feature through all the networks and gradually fusing multi-scale features to enrich the high-resolution features. The proposed FEM can be deemed as a variant of HRB to tackle the few-shot segmentation problem. The inter-source enrichment of FEM is analogous to the multi-resolution parallel convolution in HRB as shown in Figure \ref{fig:hrb}. But the inter-scale interaction in FEM passes conditioned information from large to small scales rather than dense interaction among all scales without selection in HRB. 

\begin{figure}[t]
\centering
    \begin{minipage}   {0.95\linewidth}
        \centering
        \includegraphics 
        [width=1.0\linewidth]{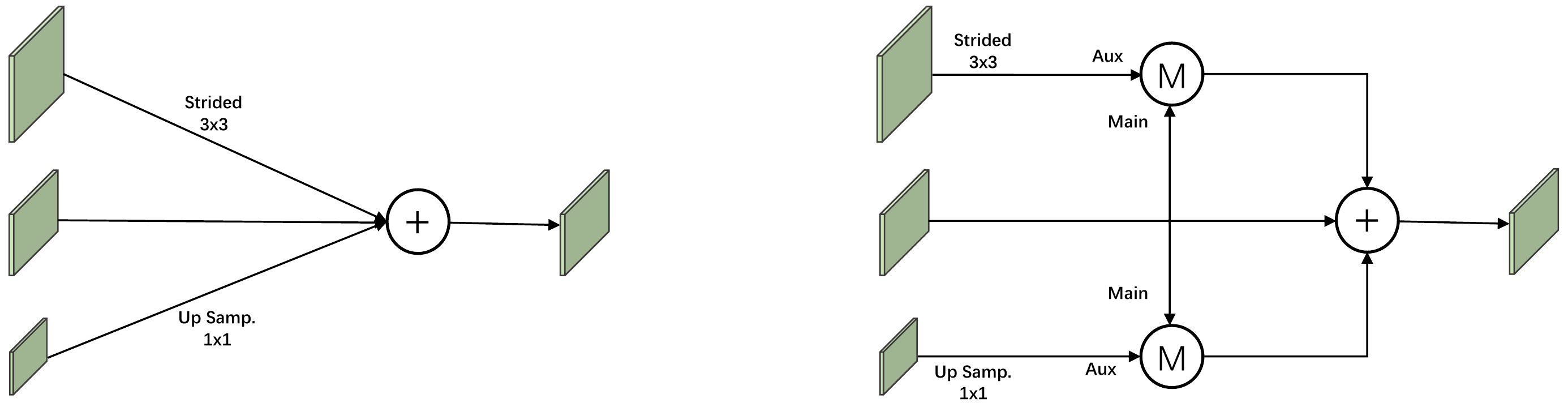}
    \end{minipage} 
    \caption{Comparison between feature fusion strategies of (left) HRB and (right) HRB-Cond. Features from different scales are directly added to the main feature in (left), while in (right), essential information is selected from auxiliary features conditioned on the main features by the inter-scale merging module $\mathcal{M}$.}
    \label{fig:hrb_merge}
\end{figure}

For comparison, we experiment with replacing the FEM in PFENet with HRB and generate feature maps in HRB with the same scales of those in FEM ($\{60,30,15,8\}$). Results are listed in Table \ref{tab:ablation_prior_simple_combine}. Directly applying HRB to the baseline (Baseline + HRB) does yield better results than PPM and ASPP. Densely passing information without selection causes redundancy to the target feature and yields suboptimal results. Our solution is, in the multi-resolution fusion stage of HRB, to apply the proposed inter-scale merging module $\mathcal{M}$ to extract essential information from the auxiliary features as shown in Figure \ref{fig:hrb_merge}. The model with conditioned feature selection (HRB-Cond) accomplishes better performance. 

As shown in Table \ref{tab:ablation_inter_scale}, passing features from coarse to fine levels (in a bottom-up order) adversely affects inter-scale interaction. We accordingly remove all bottom-up paths in HRB and only allow top-down ones (denoted as HRB-TD). It is not surprising that HRB-TD achieves better performance than HRB, and adding conditioned feature selection (HRB-TD-Cond) brings even further improvement.

The best variant of HRB (i.e., HRB-TD-Cond) yields comparable results with FEM, and yet it brings much more learnable parameters (7.5M). Therefore, for few-shot segmentation, the conditioned feature selection mechanism of the proposed inter-scale merging module $\mathcal{M}$ is essential for improving the performance of the multi-resolution structures.

\begin{table}
    \caption{Class mIoU on PASCAL-5$^i$ and efficiency of models with/without the proposed prior and FEM. Models are based on ResNet-50. 
    \textbf{Params}: The number of learnable parameters.  
    \textbf{Speed}: Average frame-per-second (FPS) of 1-shot evaluation. 
    \textbf{HRB}: Modularized block of HRNet \cite{hrnet_pami}. 
    \textbf{-TD}: Only top-down feature enrichment paths are enabled. 
    \textbf{-Cond}: The inter-scale enrichment modules are implemented to pass the conditioned information. 
    \textbf{FEM}: Feature enrichment module with $\{60, 30, 15, 8\} $. 
    \textbf{FEM$^{\ddagger}$}: FEM with spatial sizes $\{60, 30, 15, 8, 4\}$. 
    \textbf{Prior}: Prior masks got by fixed high-level features (conv5\_x).     
    \textbf{Baseline$^{\dagger}$}: Models trained with all backbone parameters.
    \textbf{Prior$^{\dagger}$}: Prior masks got by learnable high-level features.}
    \centering
    {
        \begin{tabular}{ l |  c  | c | c | c}
            \toprule
            \textit{Methods} & 1-Shot & 5-Shot & Params & Speed \\
            \specialrule{0em}{0pt}{1pt}
            \hline
            \specialrule{0em}{1pt}{0pt} 
            Baseline
            & 56.3 & 58.0 & 4.5 M & 17.7 FPS \\      
            \specialrule{0em}{0pt}{1pt}
            \hline
            \specialrule{0em}{1pt}{0pt}             
            Baseline + PPM \cite{pspnet}
            & 57.1 & 57.9 & 5.7 M & 17.6 FPS \\    
            Baseline + ASPP \cite{aspp}
            & 56.9 & 58.0 & 7.9 M & 17.5 FPS \\                    
            Baseline + HRB \cite{hrnet_pami}
            & 58.3  & 59.4 &  14.4 M & 15.7 FPS \\  
            \specialrule{0em}{0pt}{1pt}
            \hline
            \specialrule{0em}{1pt}{0pt}                   
            Baseline + HRB-Cond 
            & 59.2  & 60.0 &  23.0 M & 14.5 FPS \\                 
            Baseline + HRB-TD
            & 58.9  & 60.0 &  14.0 M & 16.1 FPS \\                 
            Baseline + HRB-TD-Cond
            & 59.3  & 60.4 &  18.3 M & 15.6 FPS \\                                              
            \specialrule{0em}{0pt}{1pt}
            \hline
            \specialrule{0em}{1pt}{0pt}         
            Baseline + FEM
            & 59.2 & 60.4 & 10.8 M & 17.3 FPS \\    
            Baseline + FEM$^{\ddagger}$
            & 58.9 & 60.2 & 12.9 M & 16.1 FPS \\   
            Baseline + Prior
            & 58.2 & 59.6 & 4.5 M & 16.5 FPS\\    
            Baseline + FEM + Prior
            & \textbf{60.8} & \textbf{61.9} & 10.8 M & 15.9 FPS \\ 
                     
            \specialrule{0em}{0pt}{1pt}
            \hline 
            \specialrule{0em}{1pt}{0pt}             
            Baseline$^{\dagger}$ 
            & 48.8 & 50.1 & 28.2 M & 17.7 FPS \\  
            Baseline$^{\dagger}$ + FEM 
            & 50.2 & 52.3 & 34.5 M & 16.1 FPS  \\               
            Baseline$^{\dagger}$ + Prior$^{\dagger}$
            & 49.7 & 53.1 & 28.2 M & 16.5 FPS \\    
            Baseline$^{\dagger}$ + FEM + Prior$^{\dagger}$ 
            & 51.9 & 55.3 & 34.5 M & 15.9 FPS \\                 
            \bottomrule            
        \end{tabular}
    }
    \label{tab:ablation_prior_simple_combine}
\end{table}

\begin{figure}
	\centering
	\begin{tabular}{@{\hspace{0.0mm}}c@{\hspace{1.0mm}}c@{\hspace{1.0mm}}c@{\hspace{1.0mm}}c@{\hspace{1.0mm}}c@{\hspace{1.0mm}}c@{\hspace{1.0mm}}}
		\includegraphics[width=0.16\linewidth]{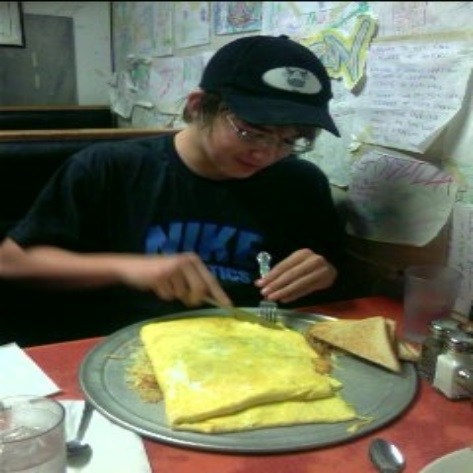}&
		\includegraphics[width=0.16\linewidth]{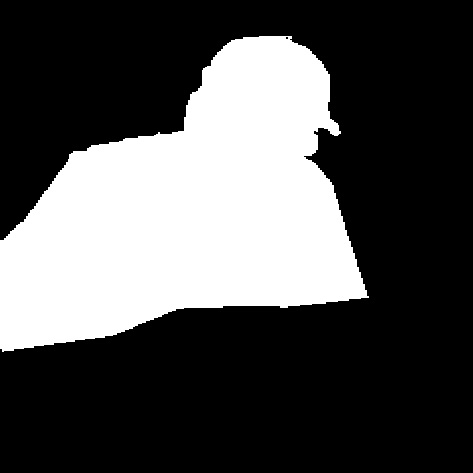}&
		\includegraphics[width=0.16\linewidth]{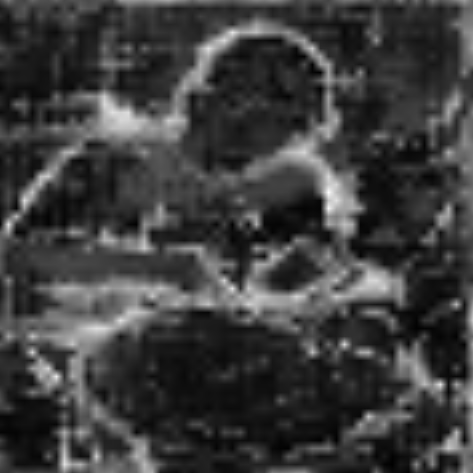}&
		\includegraphics[width=0.16\linewidth]{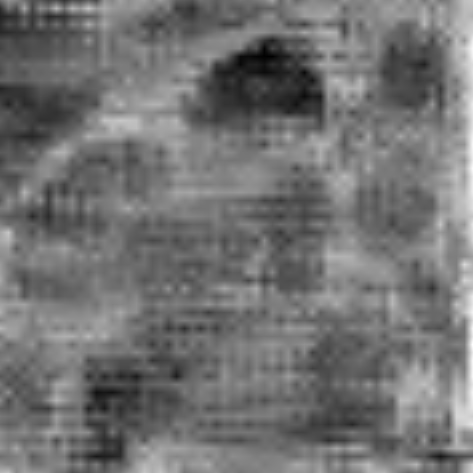}&
		\includegraphics[width=0.16\linewidth]{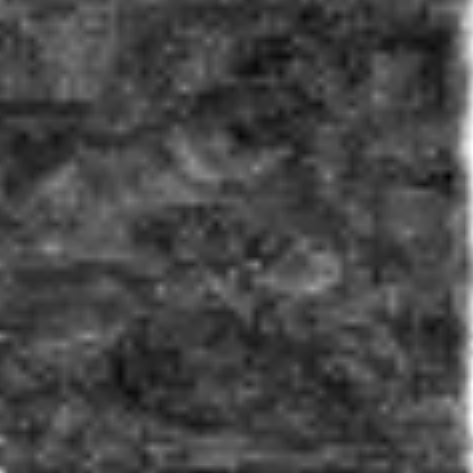}&
		\includegraphics[width=0.16\linewidth]{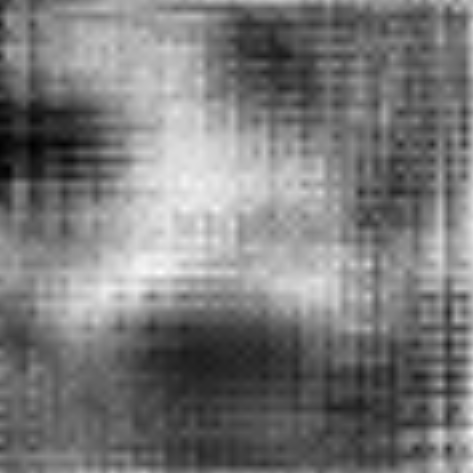}\\
		\includegraphics[width=0.16\linewidth]{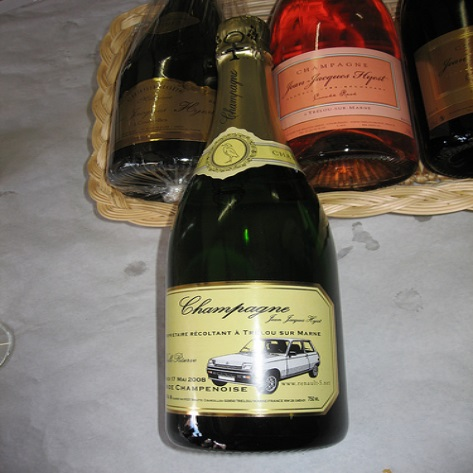}&
		\includegraphics[width=0.16\linewidth]{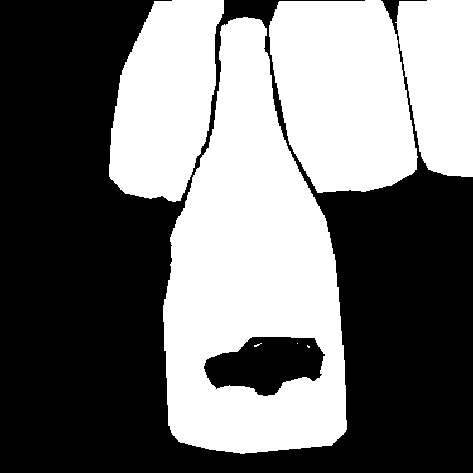}&
		\includegraphics[width=0.16\linewidth]{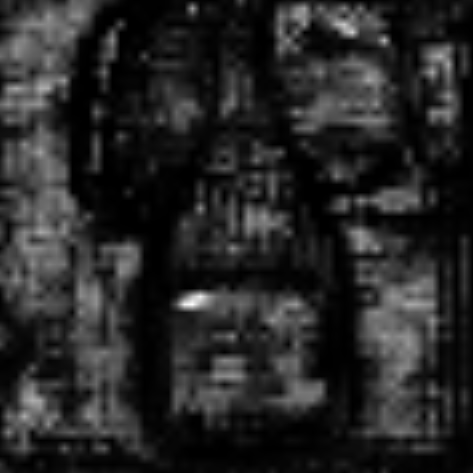}&
		\includegraphics[width=0.16\linewidth]{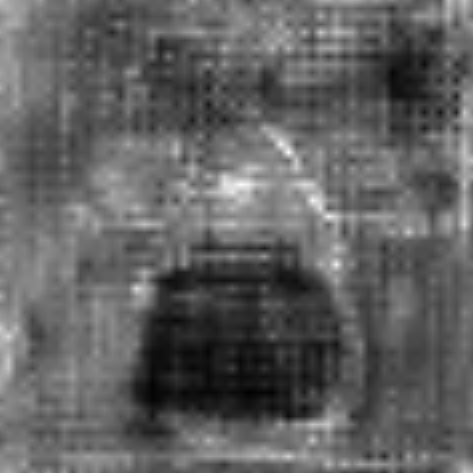}&
		\includegraphics[width=0.16\linewidth]{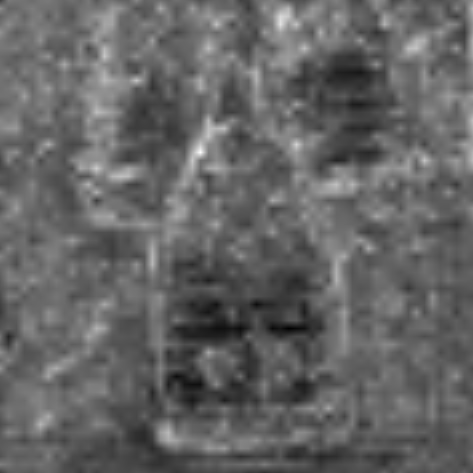}&
		\includegraphics[width=0.16\linewidth]{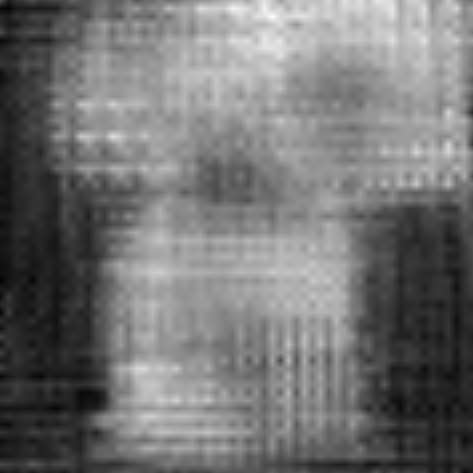}\\
		\includegraphics[width=0.16\linewidth]{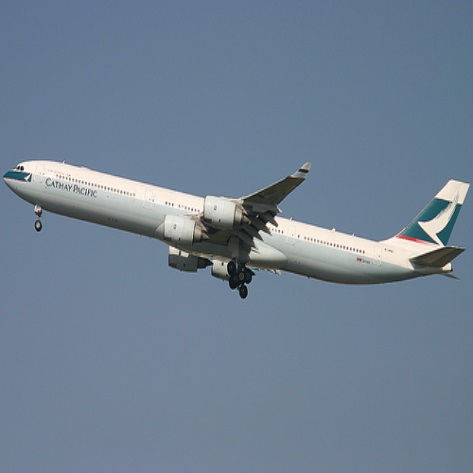}&
		\includegraphics[width=0.16\linewidth]{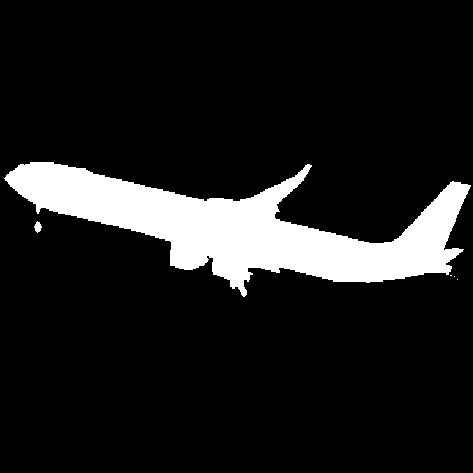}&
		\includegraphics[width=0.16\linewidth]{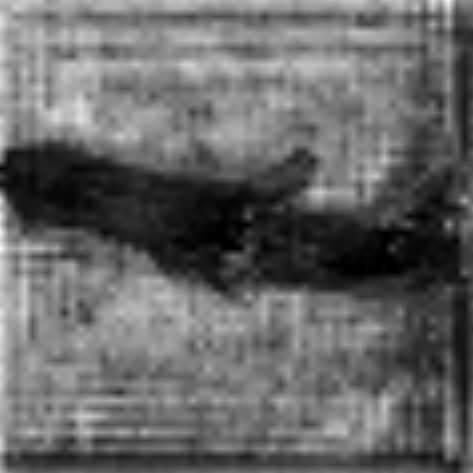}&
		\includegraphics[width=0.16\linewidth]{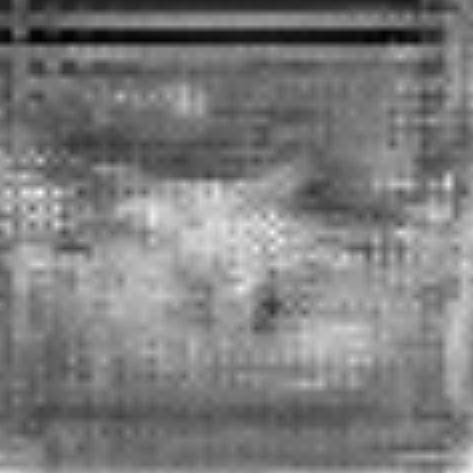}&
		\includegraphics[width=0.16\linewidth]{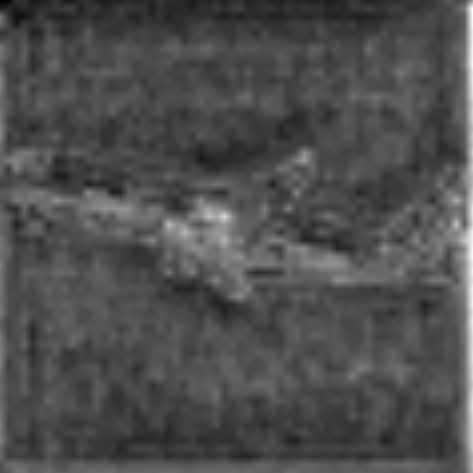}&
		\includegraphics[width=0.16\linewidth]{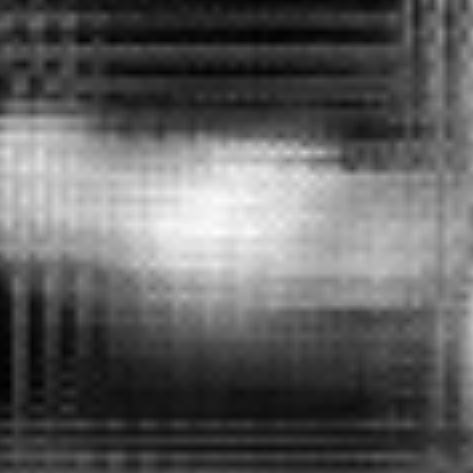}\\
		\scriptsize{(a) Input} & \scriptsize{(b) GT} & \scriptsize{(c) L-M} & \scriptsize{(d) L-H} & \scriptsize{(e) F-M} & \scriptsize{(f) F-H}
	\end{tabular}
	\caption{Visual comparison between priors generated by different sources. Prior values are normalized to 0-1, which implies the probability of being the target region. \textbf{GT}: Ground truth. \textbf{L-M}: Learnable middle-level features. \textbf{L-H}: Learnable high-level features. \textbf{F-M}: Fixed middle-level features. \textbf{F-H}: Fixed high-level features.}
    \label{fig:visual_compare_prior}
\end{figure}

\begin{figure*}
	\centering
	\begin{tabular}{@{\hspace{0.0mm}}c@{\hspace{1.0mm}}c@{\hspace{1.0mm}}c@{\hspace{1.0mm}}c@{\hspace{1.0mm}}}
		\includegraphics[width=0.24\linewidth]{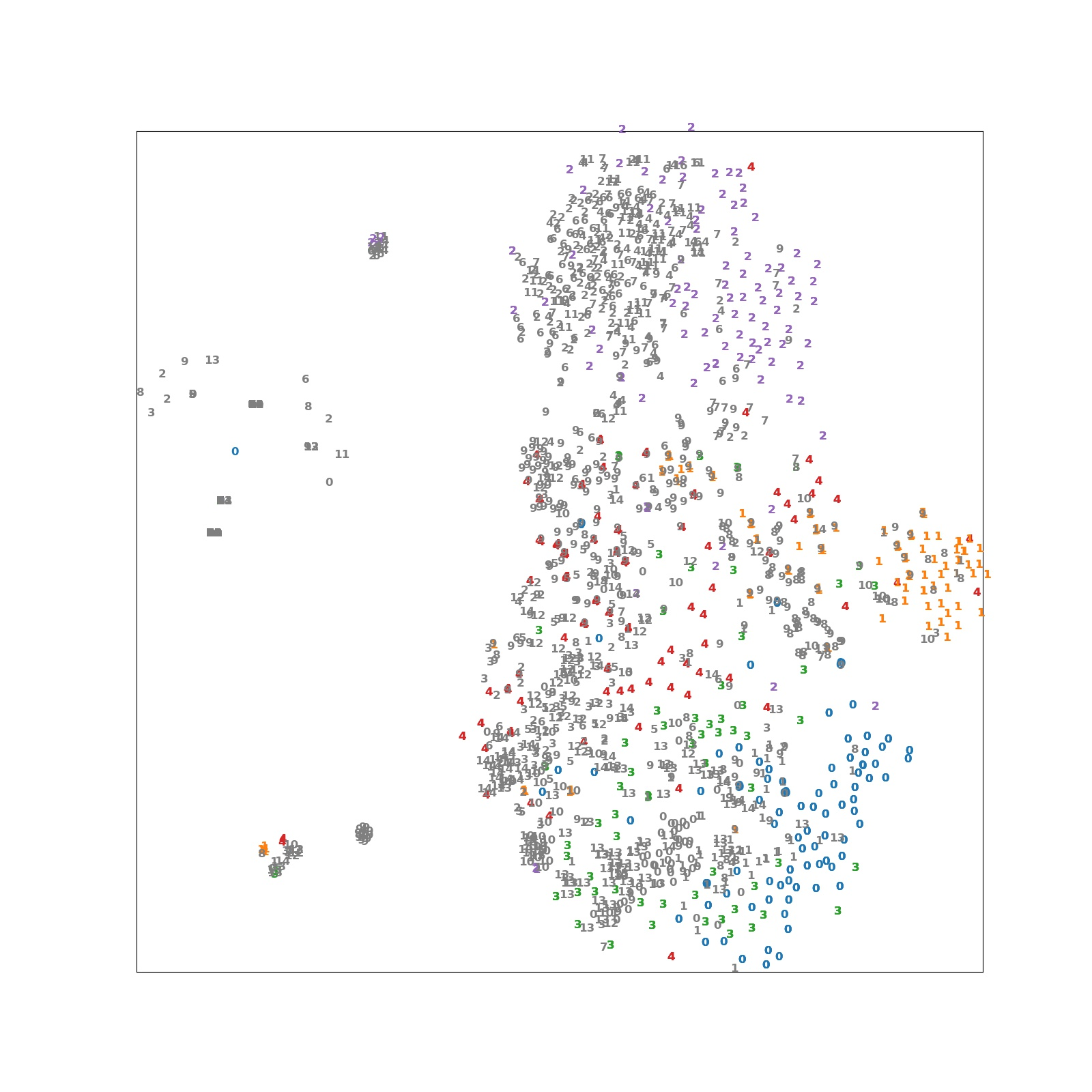}&
		\includegraphics[width=0.24\linewidth]{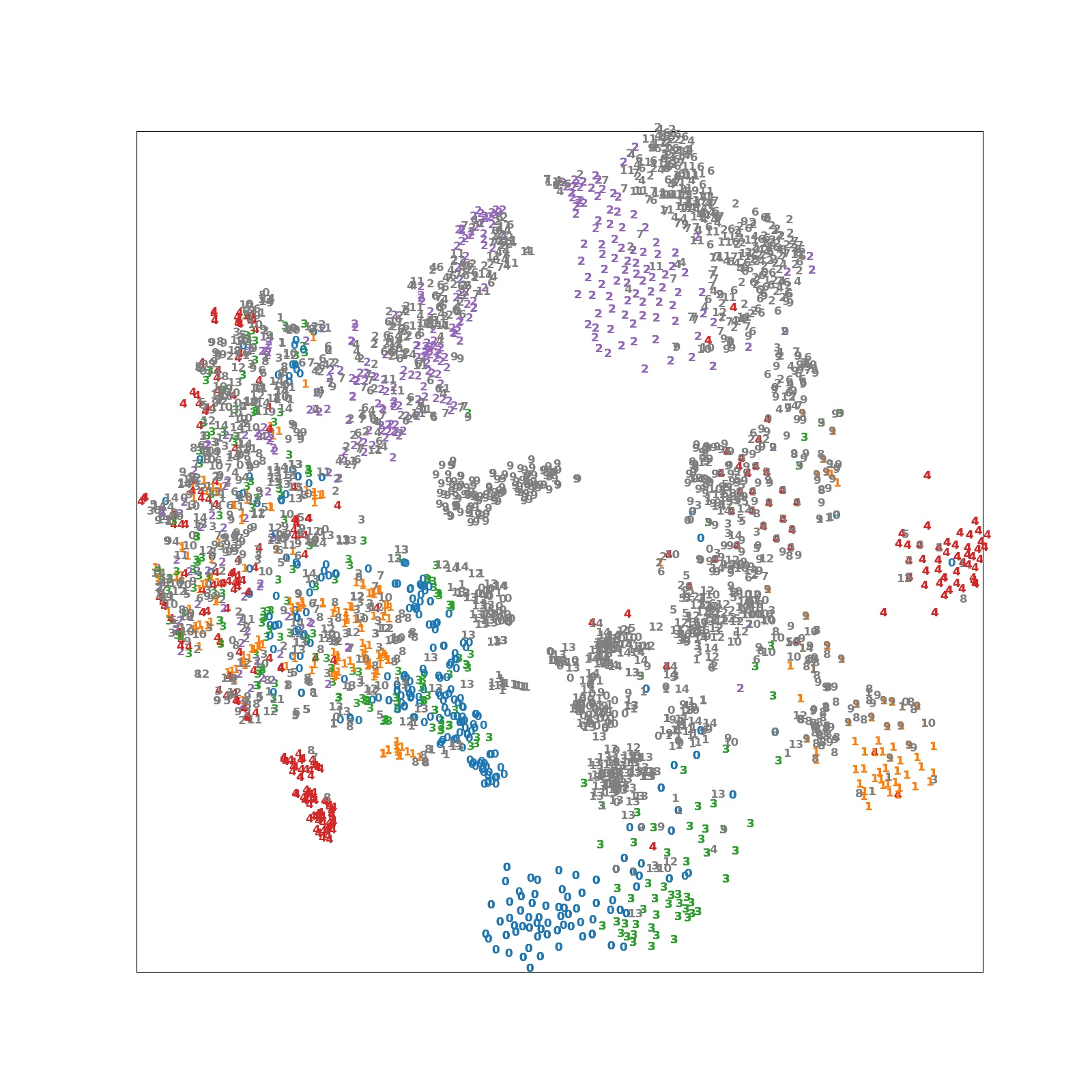}&
		\includegraphics[width=0.24\linewidth]{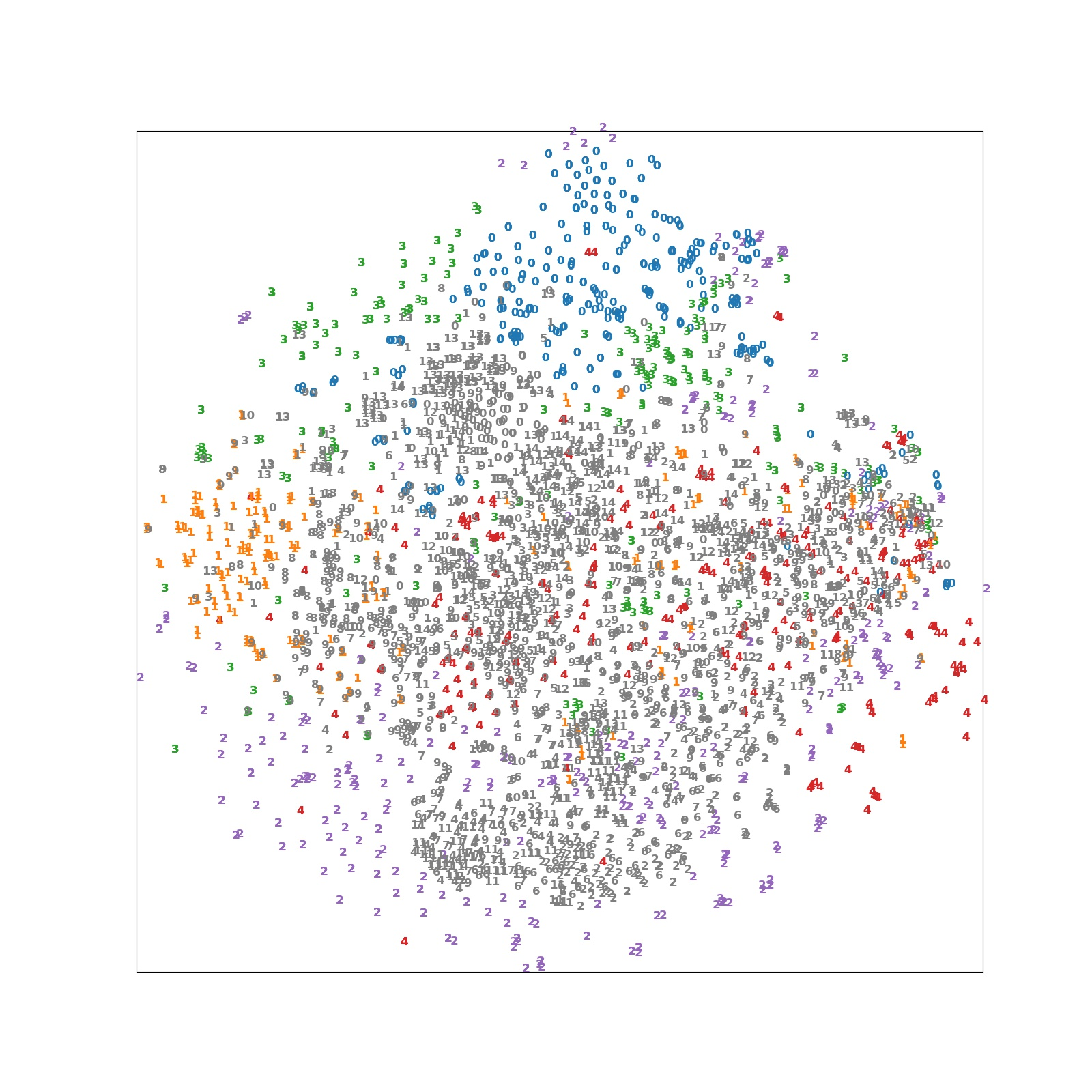}&
		\includegraphics[width=0.24\linewidth]{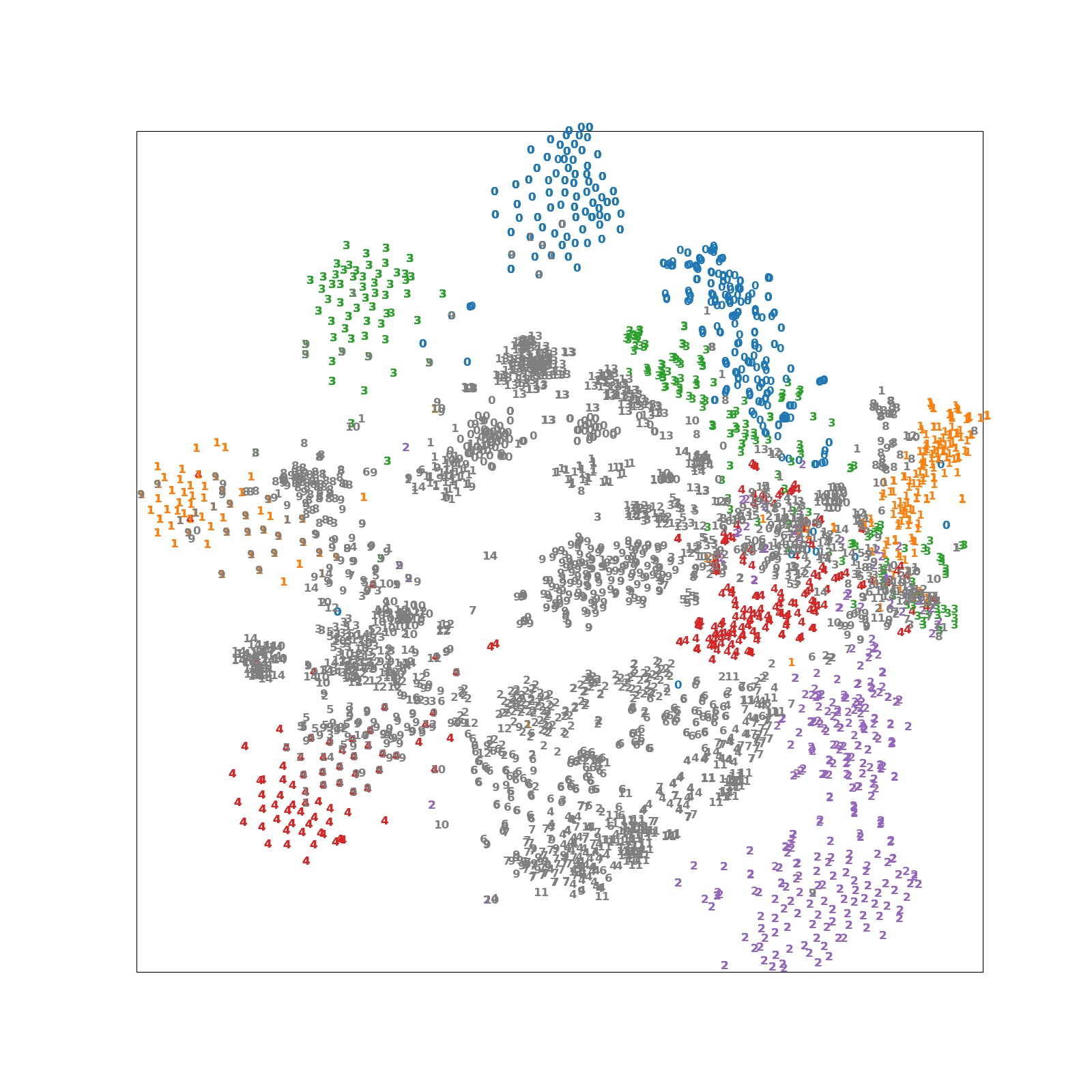} \\
	 \scriptsize{(a) L-M} & \scriptsize{(b) L-H} & \scriptsize{(c) F-M} & \scriptsize{(d) F-H}
	\end{tabular}
	\vspace{-0.2cm}
	\caption{Visual comparison between t-SNE results of different feature sources. 1,000 features in gray color are from base classes and 1,000 features in other colors are from novel classes.  \textbf{L-M}: Learnable middle-level features. \textbf{L-H}: Learnable high-level features. \textbf{F-M}: Fixed middle-level features. \textbf{F-H}: Fixed high-level features.}
	\vspace{-0.3cm}
    \label{fig:visual_compare_tsne}
\end{figure*}

\begin{table*}
    \caption{Class mIoU results of different prior masks on PASCAL-5$^i$. All models in this table are based on VGG-16.
    \textbf{LM}: Learnable middle-level features.
    \textbf{LH}: Learnable high-level features.
    \textbf{FM}: Fixed middle-level features.
    \textbf{FH}: Fixed high-level features.
    \textbf{Prior}: Prior mask got by taking the maximum similarity value.     
    \textbf{Prior-A}: Prior mask got by the average similarity value. 
    \textbf{Prior-P}: Prior mask generated with the mask-pooled support feature.
    \textbf{Prior-FW}: Prior mask got by the feature weighting mechanism proposed in \cite{weighingboosting}.
    }         
    \centering
    {
        \begin{tabular}{ l |  c  c  c  c | c |  c  c  c  c | c }
            \toprule
            \multirow{2}{*}{\textit{Methods}} 
            & \multicolumn{5}{c|}{1-Shot} 
            & \multicolumn{5}{c}{5-Shot}  \\ 
            \specialrule{0em}{0pt}{1pt}
            \cline{2-11}
            \specialrule{0em}{1pt}{0pt} 
            & Fold-0 & Fold-1 & Fold-2 & Fold-3 & Mean
            & Fold-0 & Fold-1 & Fold-2 & Fold-3 & Mean \\
            \hline
            Baseline
            & 49.4  & 64.6  & 53.3  & 46.0  & 53.3 
            & 51.5  & 65.5  & 52.5  & 47.0  & 54.1   \\    
                          
            Baseline + Prior$_{LM}$
            & 50.3  & 54.5  & 53.0  & 46.2  & 53.5  
            & 51.9 &  65.7 & 52.9  & 47.2  & 54.4  \\       
            
            Baseline + Prior$_{LH}$
            & 37.8 & 60.8 & 53.5 & 43.4 & 48.9
            & 42.5 & 64.2 & 57.8 & 47.6 & 53.0  \\         
             
            Baseline + Prior$_{FM}$
            &  51.2 & 64.4  & 53.9  & 45.7  & 53.8  
            & 52.8  & 65.1  & 53.2  & 47.5  & 54.7  \\      
            
            Baseline + Prior$_{FH}$
            & \textbf{53.5}  & \textbf{65.6}  & 53.6  & \textbf{48.8}  & \textbf{55.4}  
            & \textbf{55.7}  & \textbf{66.4}  & 53.8  & 49.8  & \textbf{56.4}  \\

            Baseline + Prior-A$_{FH}$
            & 52.2  & 65.4  & 54.5  & 48.5  & 55.1   
            & 54.8  & 66.0  & 54.3  & 50.2  & 56.3  \\           
            
            Baseline + Prior-P$_{FH}$
            & 52.4  & 65.8  & 53.1  & 47.6  & 54.7   
            & 54.9  & 67.0  & 53.5  & 48.8  & 56.1  \\   

            Baseline + Prior-FW$_{LM}$
            & 50.6  & 64.9  & 52.4  & 42.9  & 52.7  
            & 53.4  & 65.5  & 51.7  & 43.2  &  53.5 \\      

            Baseline + Prior-FW$_{LH}$
             & 37.5 & 60.3 & \textbf{54.8} & 43.9 & 49.1
             & 44.2 & 62.8 & \textbf{58.5} & 47.0 & 53.1 \\   

            Baseline + Prior-FW$_{FM}$
            & 50.6 & 64.7 & 54.4 & 47.0 & 54.2
            & 52.5 & 65.4 & 53.7 & 47.8 & 54.9  \\   
            
            Baseline + Prior-FW$_{FH}$
            & 51.0 &  65.1 &  53.9 & \textbf{48.8} & 54.7  
            &  52.7 &  66.1 & 53.8  & \textbf{50.4}  & 55.8  \\                                             
            \bottomrule            
        \end{tabular}
    }
    \label{tab:ablation_prior_source}
    \vspace{-0.3cm}
\end{table*}

\subsection{Ablation Study of the Prior Generation~~}
\label{sec:ablation_prior}
Experimental results in Table \ref{tab:ablation_prior_simple_combine} show that the prior improves models w/ and wo/ FEM. 
The cosine-similarity is widely used for tackling few-shot segmentation. PANet \cite{panet} uses the cosine-similarity to yield the intermediate and the final prediction masks; SG-One \cite{SG-One} and \cite{weighingboosting} both utilize the cosine-similarity mask from the mask pooled support feature to provide additional guidance. However, these methods overlooked two factors. First, the mask generation process contains trainable components and the generated mask is thus biased towards the base classes during training.

Second, the discrimination loss is led by the masked average pooling on support features, since the most relevant information in the support feature may be overwhelmed by the irrelevant ones during the pooling operation. For example, the discriminative regions for ``cat \& dog'' are mainly around their heads. The main bodies share similar characteristics (e.g., tailed quadrupeds), making representation produced by masked global average pooling lose the discriminative information contained in the support samples.

In the following, we first show the rationale behind our prior generation using the fixed high-level feature and taking the maximum pixel-wise correspondence value from the similarity matrix. Then we make a comparison with other methods to demonstrate the superiority of our strategy. We also include the analysis of the generalization ability on the unseen objects out of the ImageNet \cite{imagenet} dataset to further manifest the robustness of our method.

\subsubsection{Feature Selection~~}
In our design, we select the fixed high-level feature for the prior generation because it can provide sufficient semantic information for accurate segmentation without sacrificing the generalization ability. The proposed prior generation is independent of the training process. So it does not lead to loss of generalization power. The prior masks provide the bias-free prior information from high-level features for both seen and unseen data during the evaluation, while masks produced by learnable feature maps (e.g., \cite{panet, SG-One, weighingboosting}) are affected by parameter learning during training. As a result, the preference for the training classes is inevitable for these later masks during the inference. To show the superiority of our choice, we conduct experiments on different sources of features for generating prior masks. 

\vspace{0.05in}
\noindent\textbf{Quantitative Analysis~~}
Table \ref{tab:ablation_prior_source} shows that the mask generated by either learnable or fixed middle-level features (Prior$_{LM}$ or Prior$_{FM}$) is less improved than our Prior$_{FH}$ since the middle-level feature is less effective to reveal the semantic correspondence between the query and support features. However, the results of mask got by learnable high-level feature (Prior$_{LH}$) are even significantly worse than that of our baseline due to the fact that the learnable high-level feature severely overfits to the base classes: the model relies on the accurate prior masks produced by the learnable high-level feature for locating the target region of base classes during training and therefore it hardly generalizes to the previously unseen classes during inference.

\vspace{0.05in}
\noindent\textbf{Qualitative Analysis~~}
Generated prior masks are shown in Figure \ref{fig:visual_compare_prior}. Masks of unseen classes generated by learnable high-level feature maps (L-H) cannot reveal the potential region-of-interest clearly while using the fixed high-level feature maps (F-H) keeps the general integrity of the target region. Compared to high-level features, prior masks produced by middle-level ones (L-M and F-M) are more biased towards the background region.  

To help explain the quantitative results and those in Figure \ref{fig:visual_compare_prior}, embedding visualization is shown in Figure \ref{fig:visual_compare_tsne} where 1,000 samples of base classes (gray) and 1,000 samples of novel classes (colored in green, red, purple, blue and orange) are processed by the backbone followed by t-SNE \cite{tsne}. Based on the overlapping area between the clusters of the base and novel classes, we draw two conclusions. First, the middle-level features in Figures \ref{fig:visual_compare_tsne} (a) \& (c) are less discriminative than the high-level features as shown in Figure \ref{fig:visual_compare_tsne}(b) \& (d). Second, learnable features lose discrimination ability as shown in (a) \& (b) because embeddings of novel classes bias towards that of the base classes, which is detrimental to the generalization on unseen classes.

\subsubsection{Discrimination Ability~~}
In our model, the prior mask acts as a pixel-wise indicator for each query image. As given in Eq. \eqref{eqn:maxcosine}, taking the maximum correspondence value from the pixel-wise similarity between the query and support features indicates that there exists at least one pixel/area in the support image that has close semantic relation to the query pixel with a high prior value. It is beneficial to reveal most of the potential targets on query images. 
Other alternatives include using mask pooled support feature to generate the similarity mask as \cite{panet,SG-One,weighingboosting}, and taking the average value rather than the maximum value from the pixel-wise similarity.

To verify the effectiveness of our design, we train two additional models in Table \ref{tab:ablation_prior_source}: one with prior masks generated by averaging similarities (Prior-A$_{FH}$), and another whose prior masks are obtained by the mask-pooled support feature (Prior-P$_{FH}$). They both perform less satisfyingly than the proposed strategy (Prior$_{FH}$). 

We note the following fact. 
Our prior generation method takes the maximum value from a similarity matrix of size $hw \times hw$ to generate the prior mask of size $h \times w$ (Eq. \eqref{eqn:maxcosine}), in contrast to Prior-P forming the mask from the similarity matrix of size $hw \times 1$, the difference of speed is rather small because computational complexities of the two mask generation methods are much smaller than that of the rest of network. The FPS values of Prior$_{FH}$, Prior-A$_{FH}$, Prior-P$_{FH}$ and Prior-FW$_{FH}$ based on VGG-16 baseline are both around 23.1 FPS because the output features only contain 512 channels. The FPS values of Prior$_{FH}$, Prior-A$_{FH}$, Prior-P$_{FH}$ and Prior-FW$_{FH}$ based on ResNet-50 baseline whose output features have 2048 channels are 16.5, 16.5, 17.4 and 17.0 respectively.

\subsubsection{Comparison with Other Designs~~}
Some other methods also use the similarity mask as an intermediate guidance for improving performance (e.g. \cite{panet}, \cite{SG-One}, \cite{weighingboosting}). Their masks are obtained by the learnable mask-pooled support and learnable query feature that is then used for further processing the making final prediction. The strategy of this type of method is similar to Prior-P$_{LM}$. 

In \cite{weighingboosting}, the good discrimination ability of features makes activation high on the foreground and low elsewhere. 
We follow Eqs. (3)-(6) in \cite{weighingboosting} to implement the feature weighting mechanism on both the query and support features used for prior mask generation.
In \cite{weighingboosting}, the weighting mechanism is directly applied to learnable features, and we offer two choices in our model: the learnable middle- and high-level features. However, it does not perform better for Prior-FW$_{LM}$ and Prior-FW$_{LH}$. Results of Prior-FW$_{FH}$ demonstrates the effectiveness of our feature selection strategy (with fixed high-level features) for prior generation. Our feature selection strategy is complementary to the weighting mechanism of \cite{weighingboosting}.

\subsubsection{Generalization on Totally Unseen Objects~~}
Many objects of PASCAL-$5^i$ and COCO have been included in ImageNet \cite{imagenet} for backbone pre-training. For those previously unseen objects, the backbone still provides strong semantic cues to help identify the target area in query images with the information provided by the support images. 
The class `Person' in PASCAL-5$^i$ is not contained in ImageNet, and the baseline with the prior mask achieves 15.81 IoU, better than that without the prior mask (14.38). However, the class `Person' is not rare in ImageNet samples even if their labels are not `Person'.

To further demonstrate our generalization ability to totally unseen objects, we conduct experiments on the recently proposed FSS-1000 \cite{FSS-1000} dataset where the foreground IoU is used as the evaluation metric. FSS-1000 is composed of 1,000 classes, among which 486 classes are not included in any other existing datasets \cite{FSS-1000} \footnote{In practice, 420 unseen classes are filtered out. The author of FSS-1000 has clarified in email that they "have made incremental changes to the dataset to improve class balance and label quality so the number may have changed. Please do experiments according to the current version."}. 
We train our models with ResNet-50 backbone on the seen classes for 100 epochs with batch size 16 and initial learning rate 0.01, and then test them on the unseen classes. The number of query-support pairs sampled for testing is equal to five times the number of unseen samples.

\begin{figure*}
	\centering
	\begin{tabular}{@{\hspace{0.0mm}}c@{\hspace{1.0mm}}c@{\hspace{1.0mm}}c@{\hspace{1.0mm}}c@{\hspace{1.0mm}}c@{\hspace{1.0mm}}c@{\hspace{1.0mm}}c@{\hspace{1.0mm}}c@{\hspace{1.0mm}}}
		\includegraphics[width=0.11\linewidth]{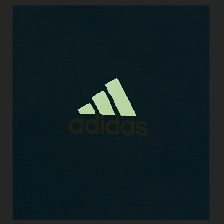}&
		\includegraphics[width=0.11\linewidth]{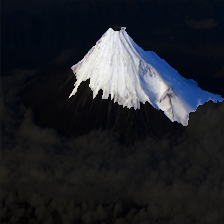}&
		\includegraphics[width=0.11\linewidth]{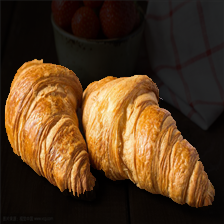}&
		\includegraphics[width=0.11\linewidth]{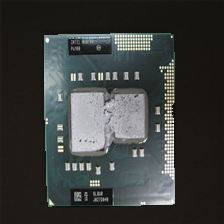}&
		\includegraphics[width=0.11\linewidth]{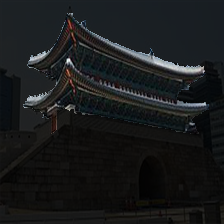}&
		\includegraphics[width=0.11\linewidth]{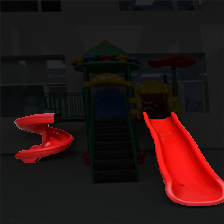}&
		\includegraphics[width=0.11\linewidth]{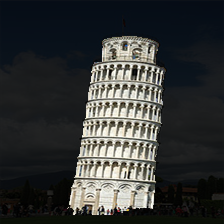}&
		\includegraphics[width=0.11\linewidth]{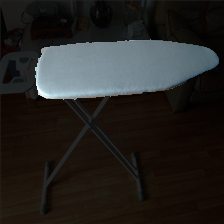}\\
			
		\includegraphics[width=0.11\linewidth]{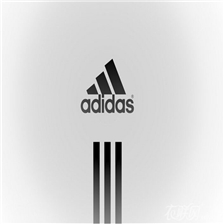}&
		\includegraphics[width=0.11\linewidth]{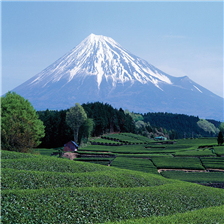}&
		\includegraphics[width=0.11\linewidth]{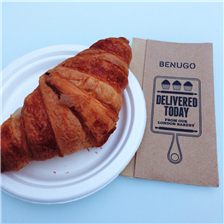}&
		\includegraphics[width=0.11\linewidth]{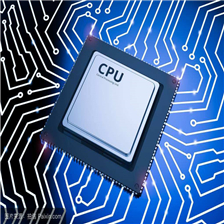}&
		\includegraphics[width=0.11\linewidth]{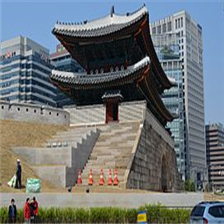}&
		\includegraphics[width=0.11\linewidth]{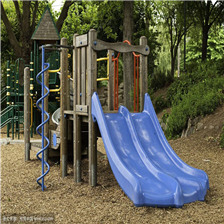}&
		\includegraphics[width=0.11\linewidth]{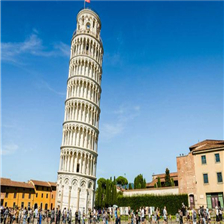}&
		\includegraphics[width=0.11\linewidth]{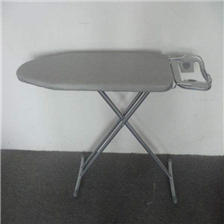}\\
		
		\includegraphics[width=0.11\linewidth]{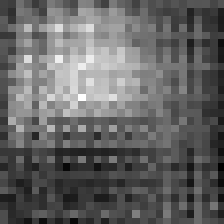}&
		\includegraphics[width=0.11\linewidth]{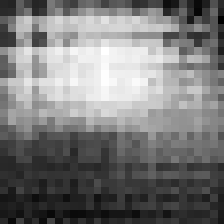}&
		\includegraphics[width=0.11\linewidth]{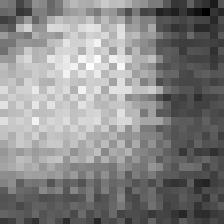}&
		\includegraphics[width=0.11\linewidth]{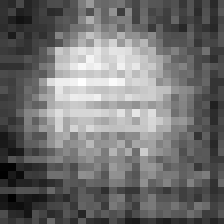}&
		\includegraphics[width=0.11\linewidth]{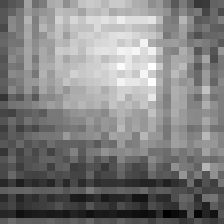}&
		\includegraphics[width=0.11\linewidth]{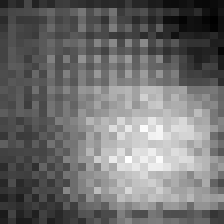}&
		\includegraphics[width=0.11\linewidth]{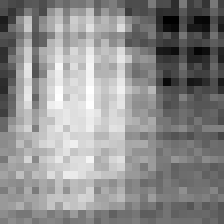}&
		\includegraphics[width=0.11\linewidth]{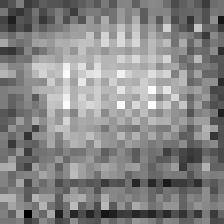}\\
	\end{tabular}
	\vspace{-0.1cm}
	\caption{Visual illustrations of prior masks for totally unseen objects in FSS-1000 dataset. \textbf{Top}: support images with the masked area in the target class. \textbf{Middle}: query images. \textbf{Bottom}: prior masks of query images where the regions of interest are highlighted. }
    \label{fig:unseen_prior}
\end{figure*}

As shown in Table \ref{tab:ablation_unseen_object}, the baseline with the prior mask achieves 80.8 and 81.4 foreground IoU in 1- and 5-shot evaluations respectively that outperform the vanilla baseline (79.7 and 80.1) by more than 1.0 foreground IoU in both settings. 
The visual illustration is given in Figure \ref{fig:unseen_prior} where the target regions can still be highlighted in the prior masks even if these objects were not witnessed by the ImageNet pre-trained backbone. 

\begin{table}
    \caption{Foreground IoU results on totally unseen classes of FSS-1000 \cite{FSS-1000}.}         
    \centering
     \vspace{-0.2cm}
    {
        \begin{tabular}{ l |  c  c  }
            \toprule
            \textit{Methods} & 1-Shot & 5-Shot  \\
            \hline
            
            Baseline
            & 79.7  & 80.1   \\    
		   Baseline + Prior
            & \textbf{80.8}  & \textbf{81.4}  \\        
            \bottomrule      
     
        \end{tabular}
    }
    \label{tab:ablation_unseen_object}
    \vspace{-0.4cm}
\end{table}

\subsection{Backbone Training~~}
In OSLSM \cite{shaban}, two backbone networks are trained to achieve few-shot segmentation. However, backbone parameters in recent work \cite{canet,panet} are kept to prevent overfitting. There is no experiment to show what effect the backbone training has. To reach a better understanding of how the backbone affects our method, the results of four models trained with all parameters in the backbone are shown in the last four rows of Table \ref{tab:ablation_prior_simple_combine}. 

The additional trainable backbone parameters cause significant performance reduction due to the overfitting of training classes. Moreover, the backbone training nearly doubles the training time of each batch because an additional parameter update is required. It does not, however, affect the inference speed. As shown in the results, the improvement that FEM and prior mask bring to models with trainable backbones is less significant than on those with fixed backbones. 
We note that the prior masks in this section are produced by learnable high-level features because the whole backbone is trainable. The learnable high-level features bring worse performance to the fixed backbone as shown in Table \ref{tab:ablation_prior_source}, but they are beneficial to the trainable backbone.
On 5-shot evaluation, the prior yields higher performance gain compared to FEM, because the prior is averaged over five support samples, providing a more accurate prior mask than 1-shot for query images to combat overfitting. Finally, the model with both FEM and the prior still outperforms the baseline model, which demonstrates the robustness of our proposed design even with all learnable parameters.

\subsection{Model Efficiency}
\noindent\textbf{Parameters~~}
The parameters of our backbone network are fixed as those in \cite{panet,canet,Zhang_2019_ICCV}. Four parts in the baseline model are learnable: two 1$\times$1 convolutions for reducing dimension number of the query and support features, FEM, one convolution block and one classification head. As shown in Table \ref{tab:ablation_prior_simple_combine}, our best model (Baseline + FEM + Prior) only has 10.8M trainable parameters that are much fewer than other methods shown in Table \ref{tab:compare_sota_class}. The prior generation does not bring additional parameters to the model, and FEM with spatial sizes $\{60, 30, 15, 8\}$ only brings 6.3M additional learnable parameters to the baseline (4.5M $\rightarrow$ 10.8M). To prove that the improvement brought by FEM is not due to more learnable parameters, we show results of the model with FEM$^{\ddagger}$ that has more parameters (12.9M) but it yields even worse results than FEM (10.8M). 

\vspace{0.05in}
\noindent\textbf{Speed~~}
PFENet based on ResNet-50 yields the best performance with 15.9 and 5.1 FPS in 1- and 5-shot setting respectively on an NVIDIA Titan V GPU. During evaluation, test images are resized to 473 $\times$ 473. As shown in Table \ref{tab:ablation_prior_simple_combine}, FEM does not affect the inference speed much (from 17.7 to 17.3 FPS). Though the proposed prior generation process slows down the baseline from 17.7 to 16.5 FPS, the final model is still efficient with 15+ FPS. Note that we include the processing time of the last block of ResNet in these experiments for a fair comparison.

\begin{table}
    \caption{Mean and Std. of five test results (class mIoU) on PASCAL-5$^i$. `Fm - Sn' means the n-shot results of Fold-m. Each row shows five test results with the values of mean and standard deviation (Std.).    \vspace{-0.175cm}}
    \renewcommand\arraystretch{1} 
    \centering
    \resizebox{1.0\linewidth}{!}{
    {
        \begin{tabular}{ c |  c   c   c   c   c | c   c }
            \toprule
            \textbf{Fold - Shot} 
            & \textbf{1} & \textbf{2} & \textbf{3} & \textbf{4} & \textbf{5} & \textbf{Mean} & \textbf{Std.} \\
            \hline
            F0 - S1
            & 61.1 & 61.9 & 62.2 & 61.6 & 61.7 & 61.7 & 0.406  \\            
            F0 - S5
            & 63.1 & 63.2 & 63.3 & 63.1 & 63.3 & 63.1 & 0.148  \\  
            \bottomrule
            F1 - S1
            & 69.5 & 69.7 & 69.1 & 69.5 & 69.7 & 69.5 & 0.245  \\           
            F1 - S5
            & 70.7 & 70.8 & 70.9 & 70.6 & 70.5 & 70.7 & 0.158  \\  
            \bottomrule            
            F2 - S1
            & 55.3 & 55.2 & 55.6 & 55.4 & 55.1 & 55.4 & 0.230  \\             
            F2 - S5
            & 55.2 & 56.3 & 55.5 & 55.9 & 56.0 & 55.8 & 0.432  \\  
            \bottomrule
            F3 - S1
            & 56.0 & 56.2 & 56.2 & 56.7 & 56.3 & 56.3 & 0.259  \\      
            F3 - S5
            & 57.9 & 58.1 & 57.9 & 58.0 & 57.6 & 57.9 & 0.187  \\  
            \bottomrule            
        \end{tabular}
    }}
    \label{tab:variance}
    \vspace{-0.3cm}
\end{table}

\begin{table*}[th]
    \caption{Analysis on values of mean and std. of five test results (class mIoU) on COCO with different numbers of test query-support pairs (1,000 and 20,000). The model is based on VGG-16 \cite{vgg}. 20,000 query-support pairs yield more stable results with a lower standard deviation than 1,000 query-support pairs. .  \vspace{-0.175cm}}
    \centering
    \resizebox{0.98\linewidth}{!}{
    {
        \begin{tabular}{ c | c | c  c  c  c  c  | c  c |  c  c  c  c  c  | c  c }
            \toprule
            \multirow{2}{*}{\textit{Folds}} 
            & \multirow{2}{*}{\textit{Pairs}} 
            & \multicolumn{7}{c|}{1-Shot} 
            & \multicolumn{7}{c}{5-Shot}  \\ 
            \specialrule{0em}{0pt}{1pt}
            \cline{3-16}
            \specialrule{0em}{1pt}{0pt} 
            & & Test-1 & Test-2 & Test-3 & Test-4 & Test-5 & Mean & Std
            & Test-1 & Test-2 & Test-3 & Test-4 & Test-5 & Mean & Std \\
            
            \specialrule{0em}{0pt}{1pt}
            \hline
            \specialrule{0em}{1pt}{0pt}
            
            Fold-0 & 1,000
            & 35.0 & 36.8 & 35.4 & 37.8 & 34.6 & 35.9 & 1.339
            & 38.6 & 35.5 & 37.8 & 38.4 & 38.9 & 37.8 & 1.369 \\
            Fold-0 & 20,000
            & 35.5  & 35.6 & 35.5 & 35.3 & 35.2 & 35.4 & 0.164 
            & 38.2 & 37.7 & 38.4 & 38.5 & 38.2 & 38.2 & 0.308 \\
            \specialrule{0em}{0pt}{1pt}
            \hline
            \specialrule{0em}{1pt}{0pt}
            Fold-1 & 1,000
            & 36.4 & 36.9 & 38.9 & 34.7 & 36.1 & 36.6 & 1.523 
            & 41.8 & 41.8 & 39.2 & 42.8 & 40.1 & 41.4 & 1.455 \\
            Fold-1 & 20,000
            & 38.3 & 37.8 & 38.2 & 38.2 & 38.2 & 38.1 & 0.195
            & 42.2 & 42.6 & 42.3 & 42.6 & 42.8 & 42.5 & 0.245 \\            
            \specialrule{0em}{0pt}{1pt}
            \hline
            \specialrule{0em}{1pt}{0pt}
            Fold-2 & 1,000
            & 36.9 & 35.1 & 37.0 & 34.3 & 32.0 & 35.1 & 2.067
            & 39.8 & 40.8 & 41.7 & 40.4 & 38.3 & 40.2 & 1.267 \\
            Fold-2 & 20,000
            & 37.0 & 36.4 & 36.9 & 36.4 & 37.2 & 36.8 & 0.363
            & 42.1 & 41.5 & 41.9 & 41.6 & 41.8 & 41.8 & 0.239 \\
            \specialrule{0em}{0pt}{1pt}
            \hline
            \specialrule{0em}{1pt}{0pt}
            Fold-3 & 1,000
            & 34.9 & 35.1 & 36.5 & 34.7 & 35.9 & 35.4 & 0.756
             & 38.2 & 38.4 & 39.5 & 36.9 & 38.7 & 38.3 & 0.945\\
            Fold-3 & 20,000
            & 34.6 & 34.8 & 34.5 & 34.6 & 34.9 & 34.7 & 0.164
            & 38.8 & 38.8 & 39.2 & 38.9 & 38.7 & 38.9 & 0.192 \\
            \bottomrule            
        \end{tabular}
    }}
    \vspace{-0.175cm}
    \label{tab:coco_variance}
\end{table*}

\subsection{Analysis on Result Stability}
\label{sec:stability}
As mentioned in the implementation details, evaluating 1,000 query-support pairs on PASCAL-5$^i$ and COCO may cause instability on results. In this section, we show the analysis of result stability by conducting multiple experiments with different support samples.

\vspace{0.05in}
\noindent\textbf{PASCAL-5$^i$}~~
Results in Table \ref{tab:variance} show that the values of standard deviation are lower than 0.5 in both 1-shot and 5-shot setting, which shows the stability of our results on PASCAL-5$^i$ with 1,000 pairs for evaluation. 

\vspace{0.05in}
\noindent\textbf{COCO}~~
However, 1,000 pairs are not sufficient to provide reliable results for comparison as shown in Table \ref{tab:coco_variance}, since the COCO validation set contains 40,137 images and 1,000 pairs could not even cover the entire 20 test classes. Based on this observation, we instead randomly sample 20,000 query-support pairs to evaluate our models on four folds, and the results in Table \ref{tab:coco_variance} show that 20,000 pairs bring much more stable results than 1,000 pairs.

\subsection{Extension to Zero-Shot Segmentation}
Zero-shot learning aims at learning a model that is robust even when no labeled data is given. It is an extreme case of few-shot learning. To further demonstrate the robustness of our proposed PFENet in the extreme case, we modify our model by replacing the pooled support features with class label embeddings. Note that our proposed prior generation method requires support features. Therefore the prior is not applicable and we only verify FEM on the baseline with VGG-16 backbone in the zero-shot setting.

\begin{table}
    \caption{Experimental results in the zero-shot setting. Models shown in this table are based on VGG-16.      \vspace{-0.175cm}
    }
    \centering
    \resizebox{0.98\linewidth}{!}{
    {
        \begin{tabular}{ l |  c  | c   c   c   c  | c   }
            \toprule
            \textit{Methods} & Shot & Fold-0 & Fold-1 & Fold-2 & Fold-3 & Mean  \\
			\specialrule{0em}{0pt}{1pt}
			\hline
			\specialrule{0em}{1pt}{0pt}
            OSLSM$_{2017}$ \cite{shaban} 
            & 5 & 35.9 & 58.1 & 42.7 & 39.1 & 44.0 
             \\  
            co-FCN$_{2018}$ \cite{co-FCN} 
            & 5 & 37.5 & 50.0 & 44.1 & 33.9 & 41.4 
             \\   
            SG-One$_{2018}$ \cite{SG-One} 
            & 5 & 41.9 & 58.6 & 48.6 & 39.4 & 47.1 
             \\     
            AMP$_{2019}$ \cite{adaptivemaskweightimprinting} 
            & 5 & 41.8 & 55.5 & 50.3 & 39.9 & 46.9 
             \\ 
			\specialrule{0em}{0pt}{1pt}
			\hline
			\specialrule{0em}{1pt}{0pt} 
			Kato \etal$_{2019}$ \cite{zeroshot_Kato_2019_ICCV_Workshops}
            & 0 & 39.6 & 52.6 & 41.0 & 35.6 & 42.2
             \\			
            Baseline
            & 0 & 49.4 & 67.1 & 50.3 & 46.0 & 53.2
             \\
            Baseline + FEM 
            & 0 & 50.0 & 68.5 & 51.7 & 46.6 & 54.2
             \\       
            \bottomrule            
        \end{tabular}
    }}
    \label{tab:zeroshot}
\end{table}

\vspace{0.05in}
\noindent\textbf{Structural Change~~}
Embeddings of Word2Vec \cite{w2c} and FastText \cite{fasttext} are trained on Google News \cite{googlenews} and Common Crawl \cite{crawl} respectively. The concatenated feature of Word2Vec and FastText embeddings directly replaces the pooled support feature in the original model without normalization. Therefore the structural change on the model structure is the first learnable 1$\times$1 convolution for reducing the support feature channel. Its input channel number 768 (512 + 256) in the original few-shot model (VGG-16 backbone) is updated to 600 (300 + 300) in the zero-shot model. 

\vspace{0.05in}
\noindent\textbf{Results~~}
As shown in Table \ref{tab:zeroshot}, our base structure achieves 53.2 class mIoU on unseen classes without support samples, which even outperforms some models with five support samples on PASCAL-5$^i$ in the few-shot setting of OSLSM \cite{shaban}. Also, the proposed FEM tackles the spatial inconsistency in the zero-shot setting and brings 1.0 points mIoU improvement (from 53.2 to 54.2) to the baseline.

\section{Conclusion}
We have presented the prior guided feature enrichment network (PFENet) with the proposed prior generation method and the feature enrichment module (FEM). The prior generation method boosts the performance by leveraging the cosine-similarity calculation on pre-trained high-level features. The prior mask encourages the model to localize the query target better without losing generalization power. FEM helps solve the spatial inconsistency by adaptively merging the query and support features at multiple scales with intermediate supervision and  conditioned feature selection.
With these modules, PFENet achieves new state-of-the-art results on both PASCAL-5$^i$ and COCO datasets without much model size increase and notable efficiency loss. Experiments in the zero-shot scenario further demonstrate the robustness of our work. Possible future work includes extending these two designs to few-shot object detection and few-shot instance segmentation.

\ifCLASSOPTIONcaptionsoff
  \newpage
\fi

{\small
    \bibliographystyle{ieee}
    \bibliography{fewshot}
}

\end{document}